\documentclass[twoside,11pt]{article}

\usepackage{jmlr2e}
\usepackage{url}            
\usepackage{mathtools}
\usepackage{algorithm}
\usepackage{algorithmic}
\usepackage{subfigure}
\usepackage{array}
\usepackage{floatflt}
\usepackage{multirow}
\usepackage{wrapfig}
\usepackage{microtype}  
\usepackage{color}
\usepackage{longtable,slashbox}

\usepackage{lastpage}
\jmlrheading{20}{2019}{1-\pageref{LastPage}}{12/17; Revised 1/19}{7/19}{17-723}{Bin Hong, Weizhong Zhang, Wei Liu, Jieping Ye, Deng Cai, Xiaofei He and Jie Wang}
\ShortHeadings{Scaling Up Sparse SVMs by Simultaneous Feature and Sample Reduction}{Hong, Zhang, Liu, Ye, Cai, He and Wang}
\firstpageno{1}

\def \a {\mathbf{a}}
\def \c {\mathbf{c}}
\def \e {\mathbf{e}}

\def \r {\mathbf{r}}

\def \u {\mathbf{u}} 
\def \v {\mathbf{v}}
\def \w {\mathbf{w}}
\def \x {\mathbf{x}}
\def \y {\mathbf{y}}
\def \z {\mathbf{z}}

\def \wt {\tilde{\mathbf{w}}}

\def \xb {\bar{{\mathbf{x}}}}
\def \thetab {\mathbf{\theta}}
\def \Xb {\bar{{\mathbf{X}}}}

\def \G {\mathbf{G}}
\def \I {\mathbf{I}}

\def \X {\mathbf{X}}

\def \R {\mathbb{R}}

\def \calD {\mathcal{D}}
\def \calE {\mathcal{E}}
\def \calF {\mathcal{F}}
\def \calG {\mathcal{G}}

\def \calJ {\mathcal{J}}

\def \calL {\mathcal{L}}
\def \calR {\mathcal{R}}
\def \calS {\mathcal{S}}
\def \calW {\mathcal{W}}

\def \hcalD {\hat{\mathcal{D}}}

\def \hcalF {\hat{\mathcal{F}}}
\def \hcalL {\hat{\mathcal{L}}}
\def \hcalR {\hat{\mathcal{R}}}

\renewcommand{\eqref}[1]{Eq.~(\ref{#1})}
\newcommand{\figref}[1]{Fig.~\ref{#1}}

\begin{document}

\title{Scaling Up Sparse Support Vector Machines by Simultaneous Feature and Sample Reduction}

\author{\name Bin Hong\thanks{The first two authors contribute equally.~~$\dag.$ Corresponding author.} \email hongbinzju@gmail.com\\
	\addr State Key Lab of CAD$\&$CG, College of Computer Science, Zhejiang University\\
	Hangzhou, 310058, China
	\AND
	Weizhong Zhang\footnotemark[1] \email zhangweizhongzju@gmail.com \\
	\addr Tencent AI Lab, Shenzhen, 518000, China\\
	\addr State Key Lab of CAD$\&$CG, College of Computer Science, Zhejiang University\\
	Hangzhou, 310058, China
	\AND
	\name Wei Liu \email wl2223@columbia.edu\\
	\addr Tencent AI Lab, Shenzhen, 518000, China
	\AND
	\name Jieping Ye \email jpye@umich.edu\\
	\addr Department of Computational Medicine and Bioinformatics, University of Michigan\\
	Ann Arbor, MI, 48104-2218, USA.
	\AND 
	\name Deng Cai$^\dag$ \email dengcai@cad.zju.edu.cn\\
	\name Xiaofei He \email xiaofeihe@cad.zju.edu.cn\\
	\addr State Key Lab of CAD$\&$CG, College of Computer Science, Zhejiang University\\
Hangzhou, 310058, China
\AND
	\name Jie Wang \email jiewangustc@gmail.com\\
	\addr Department of Electronic Engineering and Information Science\\
	University of Science and Technology of China, Hefei, 230026, China}
\editor{Sanjiv Kumar}
\maketitle

\begin{abstract}
	Sparse support vector machine (SVM) is a popular classification technique that can simultaneously learn a small set of the most interpretable features and identify the support vectors. It has achieved great successes in many real-world applications. However, for large-scale problems involving a huge number of samples and ultra-high dimensional features, solving sparse SVMs remains challenging. By noting that sparse SVMs induce sparsities in both feature and sample spaces, we propose a novel approach, which is based on accurate estimations of the primal and dual optima of sparse SVMs, to simultaneously identify the inactive features and samples that are guaranteed to be irrelevant to the outputs. Thus, we can remove the identified inactive samples and features from the training phase, leading to substantial savings in the computational cost without sacrificing the accuracy. Moreover, we show that our method can be extended to multi-class sparse support vector machines. To the best of our knowledge, the proposed method is the \emph{first} \emph{static} feature and sample reduction method for sparse SVMs and multi-class sparse SVMs. Experiments on both synthetic and real data sets demonstrate that our approach significantly outperforms state-of-the-art methods and the speedup gained by our approach can be orders of magnitude.
\end{abstract}


\section{Introduction}\label{sec:introduction}
Sparse support vector machine (SVM) \citep{bi2003dimensionality,wang2006doubly}  is a powerful technique that can simultaneously perform classification by margin maximization and variable selection via $\ell_1$-norm penalty. The last few years have witnessed many successful applications of sparse SVMs, such as text mining \citep{joachims1998text, yoshikawa2014latent}, bioinformatics \citep{narasimhan2013svm}, and image processing \citep{mohr2004topographic, kotsia2007facial}. Many algorithms \citep{hastie2004entire, fan2008liblinear,catanzaro2008fast,hsieh2008dual, shalev2011pegasos} have been proposed to efficiently solve sparse SVM problems. However, the applications of sparse SVMs to large-scale learning problems, which involve a huge number of samples and extremely high-dimensional features, still remain challenging. 

An emerging technique, called \emph{screening} \citep{EVR:10}, has been shown to be promising in accelerating the training processes of large-scale sparse learning models. The essential idea of screening is to quickly identify the zero coefficients in the sparse solutions without solving any optimization problems such that the corresponding features or samples---that are called \emph{inactive} features or samples---can be removed from the training phase. Then, we only need to perform optimization on the reduced data sets instead of the full data sets, leading to a substantial saving in the computational cost. Here, we need to emphasize that screening differs greatly from feature selection, although they look similar at the first glance. To be precise, screening is devoted to accelerating the training processes of many sparse models including Lasso, sparse SVMs, etc., while feature selection is the goal of these models. In the past few years, many screening methods are proposed for a large set of sparse learning techniques, such as Lasso \citep{tibshirani2012strong, xiang2012fast, wang2015lasso}, group Lasso \citep{NIPS2016_6405}, $\ell_1$-regularized logistic regression \citep{wang2014safe}, and SVMs \citep{ogawa2014safe}. Most of the existing methods are in the same framework, i.e., estimating the optimum of the dual (resp. primal) problem and then developing the screening rules based on the estimations to infer which components of the primal (resp. dual) optimum are zero from the KKT conditions. The main differences among them are the techniques they use to develop their optima estimations and rules and the different sparse models they focus on. For example, SAFE \citep{EVR:10} estimates the dual optimum by calculating the optimal value's lower bound of the dual problem. The Lasso screening method \citep{wang2015lasso} estimates the optimum based on the non-expensiveness \citep{bauschke2011convex} of the projection operator by noting that its dual problem boils down to finding the projection of a given point on a convex set. Moreover, a screening method is called \emph{static}  if it triggers its screening rules before training, and \emph{dynamic} if during the training process. Empirical studies indicate that screening methods can lead to orders of magnitude of speedup in computation time.

However, most existing screening methods study either feature screening or sample screening individually and their applications have very different scenarios. Specifically, to achieve better performance (say, in terms of speedup), we favor feature screening methods when the number of features $p$ is much larger than the sample size $n$, while sample screening methods are preferable when $n\gg p$. Note that there is another class of sparse learning techniques, like sparse SVMs, which induce sparsities in both feature and sample spaces. All these screening methods tend to be helpless in accelerating the training process of these models with large $n$ and $p$. We cannot address this problem by simply combining the existing feature and sample screening methods either. The reason is that they could mistakenly discard relevant data as they are specifically designed for different sparse models. Recently, \citet{shibagaki2016simultaneous} considered this issue and proposed a method to simultaneously identify the inactive features and samples in a dynamic manner, and they trigger their testing rule when there is a sufficient decrease in the duality gap during the training process. Thus, the method in \citealt{shibagaki2016simultaneous} can discard more and more inactive features and samples as the training proceeds and one has small-scale problems to solve in the late stage of the training process. Nevertheless, the overall speedup is limited as the problems' size can be large in the early stage of the training process. To be specific, the method in \citealt{shibagaki2016simultaneous} depends heavily on the duality gap during the training process. The duality gap in the early stage can always be large, which makes the dual and primal estimations inaccurate and finally results in ineffective screening rules. Hence, it is essentially solving a large problem in the early stage.  This will be verified in the experimental comparisons in Section \ref{sec:experiments} and similar results can also be found in the recent work \citep{massias2018celer}, which merely focuses on Lasso.

In this paper, to address the limitations in the dynamic screening method, we propose a novel screening method that can \textbf{S}imultaneously identify \textbf{I}nactive \textbf{F}eatures and \textbf{S}amples (SIFS) for sparse SVMs in a static manner, and we only need to perform SIFS once \emph{before} (instead of during) training. Thus, we only need to run the training algorithm on small-scale problems. The major technical challenge in developing SIFS is that we need to accurately estimate the primal and dual optima. The more accurate the estimations are, the more effective SIFS is in detecting inactive features and samples. Thus, our major technical contribution is a novel framework, which is based on the strong convexity of the primal and dual problems of sparse SVMs for deriving accurate estimations of the primal and dual optima (see Section \ref{sec:estimation}). Another appealing feature of SIFS is the so-called \emph{synergistic effect} \citep{shibagaki2016simultaneous}. Specifically, the proposed SIFS consists of two parts, i.e., \textbf{I}nactive \textbf{F}eature \textbf{S}creening (IFS) and \textbf{I}nactive \textbf{S}ample \textbf{S}creening (ISS). We show that discarding inactive features (resp. samples) identified by IFS (resp. ISS) leads to a more accurate estimation of the primal (resp. dual) optimum, which in turn dramatically enhances the capability of ISS (resp. IFS) in detecting inactive samples (resp. features). Thus, SIFS applies IFS and ISS in an alternating manner until no more inactive features and samples can be identified, leading to much better performance in scaling up large-scale problems than the application of ISS or IFS individually. Moreover, SIFS (see Section \ref{sec:SIFS}) is safe in the sense that the detected features and samples are guaranteed to be absent from the sparse representations. To the best of our knowledge, SIFS is the first static screening rule for sparse SVMs, which is able to simultaneously detect inactive features and samples. Experiments (see Section \ref{sec:experiments}) on both synthetic and real data sets demonstrate that SIFS significantly outperforms the state-of-the-art \citep{shibagaki2016simultaneous} in improving the training efficiency of sparse SVMs and the speedup can be orders of magnitude.

To demonstrate the flexibility of our proposed method SIFS, we extend its idea to multi-class sparse support vector machine (see Section \ref{sec:SIFS-MSSVM}), which also induces sparsities in both feature and sample spaces. Although multi-class sparse SVM has a very different structure and is more complex than sparse SVM, we will see that its dual problem  has some similar properties (especially the strong convexity) with those of sparse SVM. Recall that SIFS we developed for sparse SVMs is mainly based on  the strong convexity of the primal and dual problems. Therefore, the idea of SIFS is also applicable for multi-class sparse SVMs. Experimental results show that the speedup gained by SIFS in multi-class sparse SVMs can also be orders of magnitude.


For the convenience of presentation, we postpone the detailed proofs of all the theorems in this paper to the appendix.
At last, we should point out that this journal paper is an extension of our own previous work \citep{weizhonz2017} published at the International Conference on Machine Learning (ICML) 2017. 

\textbf{Notations:} Let $\|\cdot\|_1$, $\|\cdot\|$, and $\|\cdot\|_{\infty}$ be $\ell_1$, $\ell_2$, and $\ell_{\infty}$ norms, respectively. We denote the inner product between vectors $\x$ and $\y$ by $\langle \x, \y \rangle$, and the $i$-th component of $\x$ by $[\x]_i$. Let $[p]=\{1,2...,p\}$ for a positive integer $p$. Given a subset $\calJ:=\{j_1,...,j_k\}$ of $[p]$, let $|\calJ|$ be the cardinality of $\calJ$. For a vector $\x$, let $[\x]_\calJ = ([\x]_{j_1},...,[\x]_{j_k})^\top$. For a  matrix $\X$, let $[\X]_{\calJ}=(\x_{j_1},...,\x_{j_k})$ and ${}_{\calJ}[\X] = ((\x^{j_1})^\top,...,(\x^{j_k})^\top)^\top$, where $\x^i$ and $\x_j$ are the $i^{th}$ row and $j^{th}$ column of $\X$, respectively. For a scalar $t$, we denote $\max\{0,t\}$ by $[t]_{+}$. Let $\e_k\in \R^{K}$ be the index vector, that is, $[\e_k]_i = 1$ if $i=k$, otherwise $[\e_k]_i=0$. At last, we denote the set of nonnegative real numbers as $\mathbb{R}_+$.

\section{Basics and Motivations}\label{sec:basics}

In this section, we briefly review some basics of sparse SVMs and then motivate SIFS via the KKT conditions. Specifically, we focus on an $\ell_1$-regularized SVM with a smoothed hinge loss, which takes the form of 
\begin{align}
\min_{\w\in \R^p} P(\w;\alpha, \beta) = \frac{1}{n}\sum\limits_{i=1}^{n}\ell(1-\langle \xb_i,\w\rangle )+ \frac{\alpha}{2}\|\w\|^2+ \beta ||\w||_1, \tag{P$^*$}\label{eqn:primal}
\end{align}
where $\w\in \R^p$ is the parameter vector to be estimated, $\{\x_i,y_i\}_{i=1}^n$ is the training set, $\x_i\in\mathbb{R}^p$, $y_i\in \{-1,+1\}$, $\xb_i = y_i \x_i$, $\alpha$ and $\beta$ are two positive parameters, and $\ell(\cdot):\mathbb{R}\rightarrow\mathbb{R}_+$ is the smoothed hinge loss, i.e.,  
\begin{align}
\ell(t) = 
\begin{cases}
0, \hspace{7mm}\mbox{ if } t<0,\\
\frac{t^2}{2\gamma},\hspace{5mm}\mbox{      if } 0 \leq t \leq \gamma,\\
t-\frac{\gamma}{2}, \hspace{1mm}\mbox{ if } t > \gamma,
\end{cases}\label{eqn:hinge-loss}
\end{align}
where $\gamma \in (0,1)$. 

\begin{remark} We use the smoothed hinge loss instead of the vanilla one in order to make the objective of the Lagrangian dual problem of (\ref{eqn:primal}) strongly convex, which is needed in developing our accurate optima estimations. We should point out that the smoothed hinge loss is a pretty good approximation to the vanilla one with strong theoretical guarantees. See Section 5 in \citealt{shalev2016accelerated} for the details.
\end{remark}

We present the Lagrangian dual problem of problem (\ref{eqn:primal}) and the KKT conditions in the following theorem, which play a fundamentally important role in developing our screening rules. We will not provide its proof in the appendix since it follows from Fenchel Duality \citep[see Corollary 31.2.1 in][]{rockafellar2015convex} and a similar result can be found in \citealt{shibagaki2016simultaneous}.

\begin{theorem} \textup{\citep{rockafellar2015convex}}\label{thm:dual-kkt}
	Let $\Xb = (\xb_1,\xb_2,..., \xb_n)$ and $\calS_\beta(\cdot)$ be the soft-thresholding operator \citep{hastie2015statistical}, i.e., $[\calS_\beta(\u)]_{i} = \mbox{\rm sign}([\u]_i)(|[\u]_i|-\beta)_{+}$. Then, for problem \textup{(\ref{eqn:primal})}, the followings hold:\\
	$\rm{(i)}$ the dual problem of \textup{(\ref{eqn:primal})} is
	\begin{align}
	\min_{\thetab \in [0,1]^n} D(\theta;\alpha,\beta) =\frac{1}{2\alpha}\left\|\calS_{\beta}\left(\frac{1}{n}\Xb \thetab\right)\right\|^2 +\frac{\gamma}{2n}\|\theta\|^2- \frac{1}{n}\langle \mathbf{1},{\theta} \rangle \tag{D$^*$},\label{eqn:dual}
	\end{align}
	where $\mathbf{1}\in\mathbb{R}^n$ is a vector with all components equal to $1$;\\
	$\rm{(ii)}$ denote the optima of \textup{(\ref{eqn:primal})} and \textup{(\ref{eqn:dual})} by $\w^*(\alpha,\beta)$ and $\theta^*(\alpha,\beta)$, respectively, then,
	\begin{align}
	& \w^*(\alpha,\beta) = \frac{1}{\alpha}\calS_{\beta}\left(\frac{1}{n}\Xb \theta^*(\alpha, \beta)\right), \tag{KKT-1} \label{eqn:KKT1}\\
	&[\theta^*(\alpha,\beta)]_i = 
	\begin{cases}
	0,\hspace{13mm} \mbox{ if } 1 - \langle \xb_i, \w^*(\alpha,\beta)\rangle<0;\\
	1, \hspace{13mm} \mbox{ if } 1 - \langle \xb_i, \w^*(\alpha,\beta)\rangle > \gamma;\\
	\frac{1}{\gamma } (1 - \langle \xb_i, \w^*(\alpha,\beta)\rangle ),\hspace{5mm} \mbox{ otherwise.}
	\end{cases} \tag{KKT-2} \label{eqn:KKT2}
	\end{align}	
\end{theorem} 
According to \ref{eqn:KKT1} and \ref{eqn:KKT2}, we define 4 index sets:
\begin{align*}
\calF &= \left\{j\in [p]:\frac{1}{n}|[\Xb \thetab^*(\alpha,\beta)]_j| \leq \beta \right\},\\
\calR &=\{i\in [n]: 1-\langle \w^*(\alpha,\beta), \xb_i  \rangle<0 \},\\
\calE &=\{i\in [n]: 1-\langle \w^*(\alpha,\beta), \xb_i  \rangle \in [0,\gamma] \},\\
\calL &=\{i\in [n]: 1-\langle \w^*(\alpha,\beta), \xb_i  \rangle>\gamma \},
\end{align*}
which imply that
\begin{align}
\mbox{(i) } i \in \calF \Rightarrow [\w^*(\alpha, \beta)]_i = 0, \mbox{(ii) } \left\{
\begin{array}{cc}
i\in \calR \Rightarrow& [\theta^*(\alpha, \beta)]_i = 0, \\
i\in \calL \Rightarrow& [\theta^*(\alpha, \beta)]_i = 1. 
\end{array}
\right.&\tag{R} \label{screening-rule-origin}
\end{align}
Thus, we call the $j$-th feature \emph{inactive} if $j\in\calF$. The samples in $\calE$ are the so-called support vectors and we call the samples in $\calR$ and $\calL$ \emph{inactive} samples. 

Suppose that we are given subsets of $\calF$, $\calR$, and $\calL$. Then by the rules in (\ref{screening-rule-origin}), we can see that many coefficients of $\w^*(\alpha, \beta)$ and $\thetab^*(\alpha, \beta)$ are known. Thus, we may have much fewer unknowns to solve and the problem size can be dramatically reduced. We formalize this idea in Lemma \ref{lemma:dual-scaled}.

\begin{lemma}\label{lemma:dual-scaled}
	Given index sets $\hat{\calF}\subseteq \calF, \hat{\calR} \subseteq \calR$, and $ \hat{\calL}\subseteq \calL$, the followings hold:\\
	$\rm{(i)}$ $[\w^*(\alpha, \beta)]_{\hat{\calF}}=0$, $[\theta^*(\alpha, \beta)]_{\hat{\calR}} = 0$, $[\theta^*(\alpha, \beta)]_{\hat{\calL}} = 1$;\\
	$\rm{(ii)}$ let $\hat{\calD} = \hat{\calR}\cup \hat{\calL}$, $\hat{\G}_{1} ={}_{\hat{\calF}^c} [\Xb]_{\hat{\calD}^c}$, and $\hat{\G}_2 = {}_{\hat{\calF}^c} [\Xb]_{\hat{\calL}}$, where $\hat{\calF}^c=[p]\setminus\hat{\calF}$, $\hat{\calD}^c=[n]\setminus\hat{\calD}$, and $\hat{\calL}^c=[n]\setminus\hat{\calL}$. Then, $[\theta^*(\alpha, \beta)]_{\hat{\calD}^c}$ solves the following scaled dual problem:
	\begin{align}
	\min_{\hat{\thetab} \in [0,1]^{|\hat{\calD}^c|}}\Big\{ \frac{1}{2\alpha}\left\|\calS_{\beta}\left(\frac{1}{n}\hat{\G}_{1} \hat{\thetab}+\frac{1}{n} \hat{\G}_{2} \mathbf{1}\right)\right\|^2+ \frac{\gamma}{2n}\|\hat{\theta}\|^2-\frac{1}{n}\langle\mathbf{1}, \hat{\thetab}\rangle\Big\}; \tag{scaled-D$^*$}\label{eqn:dual-scaled}
	\end{align}
    $\rm{(iii)}$ suppose that $\theta^*(\alpha,\beta)$ is known, then,  
	\begin{align}
	[\w^*(\alpha,\beta)]_{\hcalF^c} = \frac{1}{\alpha}\calS_{\beta}\left(\frac{1}{n}{}_{\hcalF^c}[\Xb]\theta^*(\alpha,\beta)\right).\nonumber
	\end{align} 
\end{lemma}

Lemma \ref{lemma:dual-scaled} indicates that, if we can identify index sets $\hat{\calF}$ and $\hat{\calD}$ and the cardinalities of  $\hat{\calF}^c$ and $\hat{\calD}^c$ are much smaller than the feature dimension $p$ and the sample size $n$, we only need to solve problem (\ref{eqn:dual-scaled}) that may be much \emph{smaller} than problem (\ref{eqn:dual}) to exactly recover the optima $\w^*(\alpha, \beta)$ and $\theta^*(\alpha, \beta)$ \emph{without sacrificing any accuracy}. 

However, we cannot directly apply Rules in (\ref{screening-rule-origin}) to identify subsets of $\calF$, $\calR$, and $\calL$, as they require the knowledge of $\mathbf{w}^*(\alpha,\beta)$ and $\theta^*(\alpha,\beta)$ that are usually unavailable. Inspired by the idea in \citealt{EVR:10}, we can first estimate regions $\calW$ and $\Theta$ that contain $\w^*(\alpha, \beta)$ and $\thetab^*(\alpha, \beta)$, respectively. Then, by denoting 
\begin{align}
\hcalF \coloneqq& \left\{j\in [p]: \max_{\thetab \in \Theta}\left\{\left|\frac{1}{n}[\Xb \thetab]_j\right|\right\} \leq \beta \right\},\label{sub-F}\\
\hcalR \coloneqq&  \left\{i\in [n]: \max_{\w \in \calW} \{1-\langle \w, \xb_i  \rangle\}<0 \right \},\label{sub-R} \\
\hcalL \coloneqq& \left \{i\in [n]: \min_{\w \in \calW} \{1-\langle \w, \xb_i  \rangle\}>\gamma \right \},\label{sub-L} 
\end{align}
%
since it is easy to know that $\hcalF\subset \calF, \hcalR \subset \calR, \mbox{ and } \hcalL \subset \calL$, the rules in (\ref{screening-rule-origin}) can be relaxed as:
\begin{align}
&\mbox{(i) } j \in \hcalF \Rightarrow [\w^*(\alpha, \beta)]_j = 0, \tag{R1} \label{screening-rule-relaxed-1} \\
&\mbox{(ii) } \left\{
\begin{array}{cc}
i\in \hcalR \Rightarrow& [\theta^*(\alpha, \beta)]_i = 0, \\
i\in \hcalL  \Rightarrow& [\theta^*(\alpha, \beta)]_i = 1. 
\end{array}
\right.\tag{R2} \label{screening-rule-relaxed-2}
\end{align}
In view of Rules \ref{screening-rule-relaxed-1} and \ref{screening-rule-relaxed-2}, we sketch the development of SIFS as follows.
\begin{enumerate}
	\item[\textbf{Step 1:}] Derive estimations $\calW$ and $\Theta$ such that $\w^*(\alpha, \beta)\in\calW$ and $\thetab^*(\alpha, \beta)\in\Theta$, respectively.
	\item[\textbf{Step 2:}] Develop SIFS by deriving the relaxed screening rules \ref{screening-rule-relaxed-1} and \ref{screening-rule-relaxed-2}, i.e., by solving the optimization problems in Eqs. (\ref{sub-F}), (\ref{sub-R}) and (\ref{sub-L}).
\end{enumerate}

\section{Estimate the Primal and Dual Optima}\label{sec:estimation}
In this section, we first show  that the primal and dual optima admit closed-form solutions for specific values of $\alpha$ and $\beta$ (Section \ref{ssec:abmax}). Then, in Sections \ref{ssec:estimation_primal} and \ref{ssec:estimation_dual}, we present accurate estimations of the primal and dual optima, respectively. We would like to point out that in order to extend the optimum estimation results below to multi-class sparse SVM and avoid redundancy, we consider the optimum estimations for the primal and dual problems in more general forms.

\subsection{Effective Intervals of Parameters $\alpha$ and $\beta$ }\label{ssec:abmax}
Below we show two things. One is that if the value of $\beta$ is sufficiently large, no matter what $\alpha$ is, the primal solution is $0$. The other is for any $\beta$, if $\alpha$ is large enough, the primal and dual optima admit closed-form solutions.
\begin{lemma} \label{beta-max-and-alpha-max} 
	Let $\beta_{\rm max} = \|\frac{1}{n}\Xb\mathbf{1}\|_{\infty}$ and $\alpha_{\rm max}(\beta)=\frac{1}{1-\gamma}\max_{i\in [n]}\big\{ \langle \xb_i, \calS_{\beta}(\frac{1}{n} \Xb \mathbf{1})\rangle\big\}$. Then,\\
	\textup{(i)} for $\alpha>0$ and $\beta \geq \beta_{\rm max}$, we have
	\begin{align}
	&\w^*(\alpha, \beta) = \mathbf{0}, \hspace{4mm} \thetab^*(\alpha, \beta) = \mathbf{1};\nonumber
	\end{align}
	\textup{(ii) }for all $\alpha \in [\max\{\alpha_{\rm max}(\beta),0\}, \infty)\cap(0,\infty)$,  we have 
	\begin{align}
	\w^*(\alpha, \beta) = \frac{1}{\alpha}\calS_{\beta}\left(\frac{1}{n}\Xb \mathbf{1}\right),\hspace{4mm}\thetab^*(\alpha, \beta) =  \mathbf{1}. \label{eqn:close-form-solution} 
	\end{align}  
\end{lemma}

By Lemma \ref{beta-max-and-alpha-max}, we only need to consider the cases where $\beta\in(0, \beta_{\rm max}]$ and $\alpha \in (0, \alpha_{\rm max}(\beta)]$.  

\subsection{Primal Optimum Estimation}\label{ssec:estimation_primal}
 In Section \ref{sec:introduction}, we mention that the proposed SIFS consists of IFS and ISS, which can identify the inactive features and samples, respectively. We also mention that an alternating application of IFS and ISS can improve the estimation accuracy of the primal and dual optimum estimations, which can in turn make ISS and IFS more effective in identifying inactive samples and features, respectively. We would like to show that discarding inactive features by IFS leads to a more accurate estimation of the primal optimum.
 
We consider the following  general problem (\ref{eqn:general-primal}): 
\begin{align}
\min_{\w\in \R^p} P(\w;\alpha, \beta) = L(\w)+ \frac{\alpha}{2}\|\w\|^2+ \beta ||\w||_1, \tag{g-P$^*$}\label{eqn:general-primal}
\end{align}
where $L(\w): \R^p \rightarrow \R_+$ is smooth and convex. Exploiting the strong convexity of the objective function, we obtain the optimum estimation in the following lemma. 

\begin{lemma}\label{lemma:primal-estimation}
		Suppose that the optimum $\w^*(\alpha_0,\beta_0)$ of problem \textup{(\ref{eqn:general-primal})} at $(\alpha_0, \beta_0)$ with $\beta_0\in(0, \beta_{\rm max}]$ and $\alpha_0 \in (0, \alpha_{\rm max}(\beta_0)]$ is known.  Consider problem \textup{(\ref{eqn:general-primal})} with parameters $\alpha>0$ and $\beta_0$. Let $\hcalF$ be the index set satisfying $[\w^*(\alpha, \beta_0)]_{\hcalF} = \mathbf{0}$ and define
		\begin{align}\label{eqn:c-primal}
		\mathbf{c} =& \frac{\alpha_0+\alpha}{2\alpha}[\w^*(\alpha_0,\beta_0)]_{\hcalF^c}, \\ \label{eqn:r-primal}
		r^2=&\frac{(\alpha_0-\alpha)^2}{4\alpha^2}\|\w^*(\alpha_0,\beta_0)\|^2-\frac{(\alpha_0+\alpha)^2}{4\alpha^2}\|[\w^*(\alpha_0,\beta_0)]_{\hcalF}\|^2.
		\end{align} 
		Then, the following holds:  
		\begin{align}
		[\w^*(\alpha, \beta_0)]_{\hcalF^c}\in \{\mathbf{w}:\|\mathbf{w}- \mathbf{c}\| \leq r\}.\nonumber 
		\end{align}
	\end{lemma}

For problem (\ref{eqn:primal}), let $\hcalF$ be the index set of inactive features identified by previous IFS steps. We have $[\w^*(\alpha, \beta_0)]_{\hcalF} = \mathbf{0}$. Hence, we only need to find an estimation for $[\w^*(\alpha, \beta_0)]_{\hcalF^c}$. Since problem (\ref{eqn:primal}) is a special case of problem (\ref{eqn:general-primal}), given the reference solution $\w^*(\alpha_0,\beta_0)$ and the set $\hcalF$, from Lemma \ref{lemma:primal-estimation} we have:
\begin{align}
[\w^*(\alpha, \beta_0)]_{\hcalF^c}\in\calW\coloneqq\{\mathbf{w}:\|\mathbf{w}- \mathbf{c}\| \leq r\},\label{eqn:primal-estimation}
\end{align}
 where $\c$ and $\r$ are defined in Eqs. (\ref{eqn:c-primal}) and (\ref{eqn:r-primal}), respectively. 

It shows that $[\w^*(\alpha, \beta_0)]_{\hcalF^c}$ lies in a ball of radius $r$ centered at $\mathbf{c}$. Note that, before we perform IFS, the set $\hat{\calF}$ is empty and the second term on the right hand side  of \eqref{eqn:r-primal} is thus $0$. If we apply IFS multiple times alternatively with ISS, the set $\hat{\calF}$ will be monotonically increasing. Thus, \eqref{eqn:r-primal} implies that the radius will be monotonically decreasing, leading to a more accurate primal optimum estimation. 


\subsection{Dual Optimum Estimation}\label{ssec:estimation_dual}
We consider  a general dual problem (\ref{eqn:general-dual}) below:
\begin{align}
\min_{\thetab \in [0,1]^n} D(\theta;\alpha,\beta) =\frac{1}{2\alpha} f_{\beta}(\theta) +\frac{\gamma}{2n}\|\theta\|^2- \frac{1}{n}\langle \v,{\theta} \rangle \tag{g-D$^*$},\label{eqn:general-dual}
\end{align}
where $\v\in \R^n$ and $f_{\beta}(\theta): \R^n\rightarrow \R_+$ is smooth and convex. It is obvious that problem (\ref{eqn:general-dual}) can be reduced to problem (\ref{eqn:dual}) by letting $f_{\beta}(\theta) =\left\|\calS_{\beta}\left(\frac{1}{n}\Xb \thetab\right)\right\|^2$ and $\v = \mathbf{1}$. The lemma below gives an optimum estimation of problem (\ref{eqn:general-dual}) based on the strong convexity of its objective function.
\begin{lemma}\label{lemma:dual-estimation}
	Suppose that the optimum $\theta^*(\alpha_0,\beta_0)$ of problem \textup{(\ref{eqn:general-dual})} with $\beta_0\in(0, \beta_{\rm max}]$ and $\alpha_0 \in (0, \alpha_{\rm max}(\beta_0)]$ is known.  Consider problem \textup{(\ref{eqn:general-dual})} at $(\alpha,\beta_0)$ with $\alpha>0$ and let $\hcalR$ and $\hcalL$ be two index sets satisfying $[\theta^*(\alpha,\beta_0)]_{\hat{\calR}}=\mathbf{0}$ and $[\theta^*(\alpha,\beta_0)]_{\hat{\calL}}=\mathbf{1}$, respectively. We denote $\hcalD = \hcalR \cup \hcalL$ and define
	\begin{align}
	\mathbf{c} =& \frac{\alpha-\alpha_0}{2 \gamma \alpha } [\v]_{\hcalD^c}+\frac{\alpha_0 +\alpha}{ 2 \alpha }[\theta^*(\alpha_0, \beta_0)]_{\hcalD^c}, \label{c-dual}  \\
	r^2 =&(\frac{\alpha-\alpha_0}{2\alpha})^2||\theta^*(\alpha_0,\beta_0)-\frac{1}{\gamma}\v||^2 - ||\mathbf{1}-\frac{\alpha-\alpha_0}{2\gamma \alpha}[\v]_{\hcalL}-\frac{\alpha_0 +\alpha}{ 2 \alpha }[\theta^*(\alpha_0, \beta_0)]_{\hcalL} ||^2\nonumber \\
	&-|| \frac{\alpha-\alpha_0}{2 \gamma \alpha } [\v]_{\hcalR}+\frac{\alpha_0 +\alpha}{ 2 \alpha }[\theta^*(\alpha_0, \beta_0)]_{\hcalR} ||^2.\label{r-dual}
	\end{align}
	Then, the following holds:
	\begin{align}
	[\theta^*(\alpha,\beta_0)]_{\hcalD^c}\in \{\theta: \| \theta - \mathbf{c}\|\leq r\}. \nonumber
	\end{align}
\end{lemma}

We now turn back to problem (\ref{eqn:dual}). As it is a special case of problem (\ref{eqn:general-dual}), given the reference solution $\theta^*(\alpha_0,\beta_0)$ at $(\alpha_0,\beta_0)$  and the index sets of inactive samples identified by the previous \textup{ISS} steps $\hcalR$ and $\hcalL$, using Lemma \ref{lemma:dual-estimation}, we can obtain:
\begin{align}
[\theta^*(\alpha,\beta_0)]_{\hcalD^c}\in\Theta\coloneqq \{\theta: \|\theta - \mathbf{c}\|\leq r\}, \label{eqn:dual-estimation}
\end{align}
where $\c$ and $\r$ are defined by Eqs. (\ref{c-dual}) and (\ref{r-dual}) with $\v=\mathbf{1}$, respectively. Therefore, $[\theta^*(\alpha,\beta_0)]_{\hcalD^c}$ lies in the ball $\Theta$. In view of \eqref{r-dual}, the index sets $\hcalL$ and $\hcalR$ monotonically increase and hence the last two terms on the right hand side of \eqref{r-dual} monotonically increase when we perform ISS multiple times (alternating with IFS), which implies that the ISS steps can reduce the radius and thus improve the dual optimum estimation. 

In addition, from both Lemmas \ref{lemma:primal-estimation} and \ref{lemma:dual-estimation}, we can see that the radii of $\calW$ and $\Theta$ can be potentially large when $\alpha$ is very small, which may affect our estimation accuracy. We can sidestep this issue by letting the ratio $\alpha/\alpha_0$ be a constant. That is why we space the values of $\alpha$ equally at the logarithmic scale on the parameter value path in the experiments.

\begin{remark}\label{remark-free-refrence}
	To estimate the optima $\w^*(\alpha, \beta_0)$ and $\theta^*(\alpha,\beta_0)$ of problems \textup{(\ref{eqn:primal})} and \textup{(\ref{eqn:dual})} using Lemmas \ref{lemma:primal-estimation} and \ref{lemma:dual-estimation}, we have a free reference solution pair $\mathbf{w}^*(\alpha_0,\beta_0)$ and $\theta^*(\alpha_0,\beta_0)$ with $\alpha_0=\alpha_{\rm max}(\beta_0)$. The reason is that  $\mathbf{w}^*(\alpha_0,\beta_0)$ and $\theta^*(\alpha_0,\beta_0)$ admit closed-form solutions in this setting (see Lemma \ref{beta-max-and-alpha-max}).
\end{remark} 

\section{The Proposed SIFS Screening Rule}\label{sec:SIFS}
We first present the IFS and ISS rules in Sections \ref{ssec:IFS} and \ref{ssec:ISS}, respectively. Then, in Section \ref{ssec:SIFS}, we develop the SIFS screening rule by an alternating application of IFS and ISS.


\subsection{Inactive Feature Screening (IFS)}\label{ssec:IFS}

Suppose that $\mathbf{w}^*(\alpha_0,\beta_0)$ and $\theta^*(\alpha_0,\beta_0)$ are known, we derive IFS to identify inactive features for problem (\ref{eqn:primal}) at $(\alpha,\beta_0)$ by solving the optimization problem in \eqref{sub-F} (see Section \ref{sec:opt-s} in the appendix):
\begin{align}
s^i(\alpha, \beta_0) =\max_{\theta \in \Theta}\left \{\frac{1}{n} |\langle [\xb^i]_{\hcalD^c}, \theta \rangle  + \langle  [\xb^i]_{\hcalL},\textbf{1} \rangle| \right\}, i \in \hcalF^c,\label{eqn:opt-feature}
\end{align}
where $\Theta$ is given by \eqref{eqn:dual-estimation} and $\hcalF$ and $\hcalD=\hcalR\cup\hcalL$ are the index sets of inactive features and samples that have been identified in previous screening processes, respectively. The next result shows the closed-form solution of problem (\ref{eqn:opt-feature}).
\begin{lemma}\label{lemma:opt-feature}
	Consider problem \textup{(\ref{eqn:opt-feature})}. Let $\mathbf{c}$ and $r$ be given by \eqref{c-dual} and \eqref{r-dual} with $\v = \mathbf{1}$. Then, for all $i \in \hcalF^c$, we have 
	\begin{align}
	s^i(\alpha, \beta_0)= \frac{1}{n}(|\langle [\xb^i]_{\hcalD^c}, \mathbf{c} \rangle + \langle  [\xb^i]_{\hcalL},\textbf{1} \rangle| + \|[\xb^i]_{\hcalD^c}\|r).\nonumber
	\end{align}
\end{lemma}

We are now ready to present the IFS rule.
\begin{theorem}\label{thm:feature-screening}
	\textup{[Feature screening rule IFS]} Consider problem \textup{(\ref{eqn:primal})}. We suppose that $\mathbf{w}^*(\alpha_0,\beta_0)$ and $\theta^*(\alpha_0,\beta_0)$ are known. Then:
	\begin{enumerate}
		\item[\textup{(1)}]the feature screening rule \textup{IFS} takes the form of 
		\begin{align}
		s^i(\alpha,\beta_0) \leq \beta_0\Rightarrow[\w^*(\alpha,\beta_0)]_{i} = 0, \forall i \in \hcalF^c; \tag{IFS}\label{eqn:feature-screening} 
		\end{align} 
		\item[\textup{(2)}]we can update the index set $\hcalF$ by
		\begin{align}
		\hcalF \leftarrow \hcalF \cup \Delta \hcalF \mbox{ with } \Delta\hcalF= \{i: s^i \leq \beta_0, i \in \hcalF^c \}. \label{eqn:update-F}
		\end{align}
	\end{enumerate}
\end{theorem}


Recall that (see Lemma \ref{lemma:dual-estimation}) previous sample screening results give us a tighter dual estimation, i.e., a smaller feasible region $\Theta$ for problem (\ref{eqn:opt-feature}), which results in a smaller $s^i(\alpha,\beta_0)$. It finally brings about a more powerful feature screening rule \ref{eqn:feature-screening}. This is the so called synergistic effect.
\subsection{Inactive Sample Screening (ISS)}\label{ssec:ISS}
Similar to IFS, we derive ISS to identify inactive samples by solving the optimization problems in \eqref{sub-R} and \eqref{sub-L} (see Section \ref{sec:opt-u-l} in the appendix for details):
\begin{align}
u_i(\alpha, \beta_0) = \max_{\w \in \calW} \{1- \langle [\xb_i]_{\hcalF^c}, \w \rangle \}, i\in \hcalD^c,\label{eqn:opt-sample-1}\\
l_i(\alpha, \beta_0) = \min_{\w \in \calW} \{1- \langle [\xb_i]_{\hcalF^c}, \w \rangle \}, i\in \hcalD^c, \label{eqn:opt-sample-2}
\end{align}
where $\calW$ is given by \eqref{eqn:primal-estimation} and $\hcalF$ and $\hcalD=\hcalR\cup\hcalL$ are the index sets of inactive features and samples that have been identified in previous screening processes. We show that problems (\ref{eqn:opt-sample-1}) and (\ref{eqn:opt-sample-2}) admit closed-form solutions.
\begin{lemma}\label{lemma:opt-sample}
	Consider problems \textup{(\ref{eqn:opt-sample-1})} and \textup{(\ref{eqn:opt-sample-2})}. Let $\mathbf{c}$ and $r$ be given by \eqref{eqn:c-primal} and \eqref{eqn:r-primal}. Then,
	\begin{align}
	u_i(\alpha, \beta_0) =  1 -\langle [\xb_i]_{\hcalF^c} , \mathbf{c}\rangle + \|[\xb_i]_{\hcalF^c}\|r, i\in \hcalD^c, \nonumber \\
	l_i(\alpha, \beta_0) = 1 -\langle [\xb_i]_{\hcalF^c} , \mathbf{c}\rangle - \|[\xb_i]_{\hcalF^c}\|r, i\in \hcalD^c. \nonumber 
	\end{align}
\end{lemma}

We are now ready to present the ISS rule.
\begin{theorem}\label{thm:sample-screening}
	\textup{[Sample screening rule ISS]} Consider problem (\ref{eqn:dual}) and suppose that $\mathbf{w}^*(\alpha_0,\beta_0)$ and $\theta^*(\alpha_0,\beta_0)$ are known, then:
	\begin{enumerate}
		\item[\textup{(1)}]the sample screening rule \textup{ISS} takes the form of 
		\begin{align}
		\begin{array}{cc}
		u_i(\alpha, \beta_0) <0\Rightarrow[\theta^*(\alpha, \beta_0)]_i=0,\\
		l_i(\alpha, \beta_0)> \gamma\Rightarrow[\theta^*(\alpha, \beta_0)]_i=1,
		\end{array}\forall i \in \hcalD^c; \tag{ISS} \label{eqn:sample-screening}
		\end{align} 
		\item[\textup{(2)}]we can update the  index sets $\hcalR$ and $\hcalL$ by
		\begin{align}
	\hcalR &\leftarrow \hcalR \cup \Delta\hcalR \mbox{ with } \Delta\hcalR = \{i: u_i(\alpha, \beta_0)<0, i \in \hcalD^c \}, \label{eqn:update-R}\\
	\hcalL &\leftarrow \hcalL \cup \Delta\hcalL \mbox{ with } \Delta\hcalL = \{i: l_i(\alpha, \beta_0)>\gamma, i \in \hcalD^c \}. \label{eqn:update-L}
	\end{align}
	\end{enumerate}
\end{theorem}

The synergistic effect also exists here. Recall that (see Lemma \ref{lemma:primal-estimation}), previous feature screening results lead to a smaller feasible region $\calW$ for problems (\ref{eqn:opt-sample-1}) and (\ref{eqn:opt-sample-2}), which results in smaller $u_i(\alpha,\beta_0)$ and larger $l_i(\alpha,\beta_0)$. It finally leads us to  a more accurate sample screening rule \ref{eqn:sample-screening}.
\subsection{The Proposed SIFS Rule by An Alternating Application of IFS and ISS}\label{ssec:SIFS}
In real applications, the optimal parameter values of $\alpha$ and $\beta$ are usually unknown. To determine appropriate parameter values, common approaches, like cross validation and stability selection, need to solve the model over a grid of parameter values $\{(\alpha_{i,j}, \beta_j): i \in [M], j \in [N] \}$ with $\beta_{\rm max} >\beta_1> ...> \beta_N>0$ and  $\alpha_{\rm max}(\beta_j) >\alpha_{1,j}>...>\alpha_{M,j}>0$. This can be very time-consuming. Inspired by Strong Rule \citep{tibshirani2012strong} and SAFE \citep{EVR:10}, we develop a sequential version of SIFS in Algorithm \ref{alg:framework-screening}. 
\begin{algorithm}\caption{\quad SIFS}
	\begin{algorithmic}[1]
		\STATE {\bfseries Input:} $\beta_{\rm max} >\beta_1> ...> \beta_N>0$ and  $\alpha_{\rm max}(\beta_j) = \alpha_{0,j} >\alpha_{1,j}>...>\alpha_{M,j}>0$. 
		\FOR{$j=1$ to $N$}
		\STATE Compute the first reference solution $\w^*(\alpha_{0,j}, \beta_j)$ and $\theta^*(\alpha_{0,j}, \beta_j)$ using the close-form formulas in  \eqref{eqn:close-form-solution}. 
		\FOR{$i=1$ to $M$}
		\STATE{\bfseries Initialization: } $\hcalF = \hcalR = \hcalL = \emptyset$.
		\REPEAT
		\STATE Run sample screening using rule \ref{eqn:sample-screening} based on $\w^*(\alpha_{i-1,j}, \beta_j)$. 
		\STATE Update the inactive sample sets $\hcalR$ and $\hcalL$:\\
		$~~~~\hcalR \leftarrow \hcalR \cup \Delta\hcalR \mbox{ and } \hcalL \leftarrow \hcalL \cup \Delta\hcalL.$
		\STATE Run feature screening using rule \ref{eqn:feature-screening} based on $\theta^*(\alpha_{i-1,j}, \beta_j)$.
		\STATE Update the inactive feature set $\hcalF$:\\
		$~~~~\hcalF \leftarrow \hcalF \cup \Delta\hcalF$. 
		\UNTIL {No new inactive features or samples are identified}
		\STATE Compute $\w^*(\alpha_{i,j}, \beta_j)$ and $\theta^*(\alpha_{i,j}, \beta_j)$ by solving the scaled problem.
		\ENDFOR
		\ENDFOR
		\STATE {\bfseries Output:}$\w^*(\alpha_{i,j}, \beta_j)$ and $\theta^*(\alpha_{i,j}, \beta_j), i\in [M], j \in [N]$.
	\end{algorithmic}\label{alg:framework-screening}
\end{algorithm}
Specifically, given the primal and dual optima $\mathbf{w}^*(\alpha_{i-1,j},\beta_j)$ and $\theta^*(\alpha_{i-1,j},\beta_j)$ at $(\alpha_{i-1,j},\beta_j)$, we apply SIFS to identify the inactive features and samples for problem (\ref{eqn:primal}) at $(\alpha_{i,j},\beta_j)$. Then, we perform training on the reduced data set and solve the primal and dual optima at $(\alpha_{i,j},\beta_j)$. We repeat this process until we solve problem (\ref{eqn:primal}) at all pairs of parameter values.

Note that we insert $\alpha_{0,j}$ into every sequence $\{\alpha_{i,j}: i \in [M]\}$ ( see line 1 in Algorithm \ref{alg:framework-screening}) to obtain a closed-form solution as the first reference solution. In this way, we can avoid solving problem at $(\alpha_{1,j}, \beta_j), j \in [N]$ directly (without screening), which is time consuming. At last, we would like to point out that the values $\{(\alpha_{i,j}, \beta_j): i \in [M], j \in [N] \}$ in SIFS can be specified by users arbitrarily.

\textup{SIFS} applies \textup{ISS} and \textup{IFS} in an alternating manner to reinforce their capability in identifying inactive samples and features. In Algorithm \ref{alg:framework-screening}, we apply ISS first. Of course, we can also apply IFS first. The theorem below demonstrates that the orders have no impact on the performance of SIFS. 

\begin{theorem}\label{thm:order-of-screening}
	Given the optimal solutions $\w^*(\alpha_{i-1,j},\beta_j)$ and $\theta^*(\alpha_{i-1,j},\beta_j)$ at $(\alpha_{i-1,j}, \beta_j)$ as the reference solution pair at $(\alpha_{i,j}, \beta_j)$ for SIFS, we assume SIFS with ISS first stops after applying IFS and ISS for $s$ times and denote the identified inactive features and samples as $\hcalF_s^A, \hcalR_s^A$, and $\hcalL_s^A$. Similarly, when we apply IFS first, the results are denoted as $\hcalF_t^B, \hcalR_t^B$, and $\hcalL_t^B$. Then, the followings hold:\\
	\textup{(1)} $\hcalF_s^A=\hcalF_t^B, \hcalR_s^A=\hcalR_t^B$, and $\hcalL_s^A=\hcalL_t^B$.\\
	\textup{(2)} with different orders of applying ISS and IFS, the difference between the times of ISS and IFS we need to apply in SIFS can never be larger than 1, that is, $|s-t|\leq 1$.  
\end{theorem}

\subsection{Discussion}
After developing the proposed method SIFS above, we now turn to the related work discussion in order to highlight the differences between SIFS and the existing methods and also the novelty of our method, although we have mentioned some of them in the introduction section. We divide the previous work into two parts: screening for sparse SVM and for other sparse learning models. 

To the best of our knowledge, the screening method in \citealt{shibagaki2016simultaneous} is the only method besides our SIFS,  which can simultaneously identify the inactive features and samples for sparse SVM. There are mainly three big differences between SIFS and this method. First, the techniques used in the optima estimations of SIFS and \citealt{shibagaki2016simultaneous} are quite different. To be specific, the estimations in SIFS are developed by exploiting  the reference solution pair and carefully studying the strong convexity of the objective functions and the optimum conditions of problems (\ref{eqn:primal}) and (\ref{eqn:dual}) at the current and reference parameter value pairs (see Lemmas \ref{lemma:primal-estimation} and \ref{lemma:dual-estimation} and also their proofs for the details). In contrast, \citet{shibagaki2016simultaneous} estimated the optima heavily based on the duality gap.  As we mentioned in the introduction section, duality gap is usually large in the early stages and decreases gradually, which weakens its capability in the early stages and leads to a limited overall speedup.  Second, algorithmically, \citet{shibagaki2016simultaneous} is dynamic while our method is static. In other words,  \citet{shibagaki2016simultaneous} identifies the inactive features and samples during the training process, and in our method, we do this before the training process (see steps 6 to 11 in Algorithm \ref{alg:framework-screening}), which means \citet{shibagaki2016simultaneous} needs to apply its screening rules for many times while we trigger SIFS only once. These two technical and algorithmic differences make our method outperform \citet{shibagaki2016simultaneous}, which will be verified in the experimental section. At last, we theoretically prove that in the static scenario the orders of applying the feature and sample screening rules have no impact on the final performance, while \citet{shibagaki2016simultaneous} did not give the theoretical result accordingly in dynamic screening.  

There are mainly three big differences between our SIFS and existing methods for other sparse learning models.
First, these existing methods identify either features or samples individually and would be helpless in real applications involving a huge number of samples and extremely high-dimensional features, while SIFS identifies features and samples simultaneously. Second, technically, SIFS can incorporate the feature (resp. sample) screening results of the previous steps as the prior knowledge into the primal (reps. dual) optimum estimation to obtain a more accurate estimation. This is verified with a strong theoretical guarantee (see Lemmas \ref{lemma:primal-estimation} and \ref{lemma:dual-estimation}).  At last, we give a theoretical proof (see Sections \ref{ssec:IFS} and \ref{ssec:ISS}) to show the existence of the synergistic effect between feature and sample screening for the first time in the static scenario.

Finally, we would like to point out that although the key ideas used in some of the screening methods including SIFS seem to be similar, they are developed specifically for the sparse models they focus on, which makes it nontrivial and even very difficult to reduce one method to another by simply resetting the loss and regularizer of the model. For example, SIFS cannot be reduced to \citet{wang2014safe} by setting $\alpha=0$ and letting loss be the logistic loss, although both of their dual optimum estimations are based on the strong convexity of the objective. The reason is that they  use the strong convexity in quite different ways due to their different  dual problems including the feasible regions and the expressions of the dual objectives \citep[see Theorem 2 in][Lemma \ref{lemma:dual-estimation} above, and their proofs for the details]{wang2014safe}. Moreover, we cannot reduce SIFS to the methods \citep{ogawa2014safe, wang2013scaling} for SVM by setting $\beta$ to be 0, since they are  based on the convexity of the objective while SIFS exploits the strong convexity.




\section{SIFS for Multi-class Sparse SVMs} \label{sec:SIFS-MSSVM}
In this section, we consider extending SIFS to multi-class sparse SVMs. We will briefly review the basics of multi-class sparse SVMs and then derive a series of theorems as we did for sparse SVMs above. Finally, we will present the detailed screening rule for multi-class sparse SVMs.  

\subsection{Basics of Multi-class Sparse SVMs}  
We focus on the $\ell_1$-regularized multi-class SVM with a smoothed hinge loss, which takes the form of
\begin{align}
\min_{\w\in \R^{Kp}} P(\w;\alpha, \beta) = \frac{1}{n}\sum\limits_{i=1}^{n}\ell_i(\w)+ \frac{\alpha}{2}\|\w\|^2+ \beta ||\w||_1, \tag{m-P$^*$}\label{eqn:primal-MSSVM}
\end{align}
where $\w = [\w_1; \w_2;...;\w_K] \in \R^{Kp}$ is the parameter vector to be estimated with $\w_k \in \R^p, k=1,...,K$. The loss function $\ell_i(\w)=\sum_{k\neq y_i}\ell(\w_k^\top \x_i-\w_{y_i}^\top \x_i + 1)$, with  $\{\x_i, y_i\}_{i=1}^n$ is the training data set of $K$ classes, $\x_i\in \R^p, y_i \in \{1,...,K\}$, and $\ell(\cdot)$ is the smoothed hinge loss defined in \eqref{eqn:hinge-loss}.

The following theorem presents the Lagrangian dual problem of (\ref{eqn:primal-MSSVM}) and the KKT conditions, which are essential for developing the screening rules. 

\begin{theorem}\label{thm:dual-kkt-MSSVM}
	For each sample $(\x_i, y_i)$, we define $\u_i = 1-\e_{y_i}\in \R^K$ and $\X_i=[\X_i^1, \X_i^2,...,\X_i^K] \in \R^{Kp \times K}$, where $\X_i^k=\mbox{vec}(\x_i(\e_k-\e_{y_i})^\top)\in \R^{Kp}$. Let $\u = [\u_1;...;\u_n ]\in \R^{Kn}$ and $\X = [\X_1, \X_2,..., \X_n]\in \R^{Kp\times Kn}$. Then, for the problem \textup{(\ref{eqn:primal-MSSVM})}, the followings hold:\\
	$\rm{(i)}$ the dual problem of \textup{(\ref{eqn:primal-MSSVM})} is 
	\begin{align}
	\min_{\theta \in [0,1]^{Kn} } D(\theta;\alpha,\beta) =\frac{1}{2\alpha}\left\|\calS_{\beta}\left(\frac{1}{n}\X \thetab\right)\right\|^2 +\frac{\gamma}{2n}\|\theta\|^2- \frac{1}{n}\langle \u,{\theta} \rangle \tag{m-D$^*$},\label{eqn:dual-MSSVM}
	\end{align}
	where $\theta = [\thetab_1;...;\theta_n]\mbox{ with }\theta_i \in [0,1]^K, i =1,..., n$;\\
	$\rm{(ii)}$ denote the optima of problems \textup{(\ref{eqn:primal-MSSVM})} and \textup{(\ref{eqn:dual-MSSVM})} by $\w^*(\alpha,\beta)$ and $\theta^*(\alpha,\beta)$, respectively, then,
	\begin{align}
	& \w^*(\alpha,\beta) = \frac{1}{\alpha}\calS_{\beta}\left(-\frac{1}{n}\X \theta^*(\alpha, \beta)\right), \tag{m-KKT-1} \label{eqn:KKT1-MSSVM}\\
	&[\theta_i^*(\alpha,\beta)]_k = 
	\begin{cases}
	0,\hspace{13mm} \mbox{ if } \langle \X_i^k,\w^*(\alpha,\beta)\rangle+[\u_i]_k\leq 0;\\
	1, \hspace{13mm} \mbox{ if } \langle \X_i^k, \w^*(\alpha,\beta)\rangle +[\u_i]_k\geq  \gamma;\\
	\frac{1}{\gamma } (\langle \X_i^k, \w^*(\alpha,\beta)\rangle+[\u_i]_k),\hspace{5mm} \mbox{ otherwise};
	\end{cases} k=1,2,...,K.\tag{m-KKT-2} \label{eqn:KKT2-MSSVM}
	\end{align}	
\end{theorem} 
As we did in Section \ref{sec:basics}, we can also define 3 index sets here:
\begin{align*}
\calF &= \left\{( k,j) \in [K] \times [p]:\frac{1}{n}|[\X \thetab^*(\alpha, \beta)]_{k,j}| \leq \beta \right\},\\
\calR &=\left \{(i, k) \in [n]\times [K]:\langle \X_i^k,\w^*(\alpha, \beta)\rangle+[\u_i]_k \leq 0 \right \},\\
\calL &=\left \{(i, k) \in [n]\times [K]:\langle \X_i^k,\w^*(\alpha,\beta)\rangle+[\u_i]_k  \geq \gamma \right \},
\end{align*}
which imply that 
\begin{align}
&\mbox{(i) } (k, j) \in \calF \Rightarrow [\w_k^*(\alpha, \beta)]_j = 0, \nonumber \\
&\mbox{(ii) } \left\{
\begin{array}{cc}
(i, k) \in \calR \Rightarrow& [\theta_i^*(\alpha, \beta)]_k = 0, \\
(i, k)  \in \calL  \Rightarrow& [\theta_i^*(\alpha, \beta)]_k = 1. 
\end{array}
\right.\tag{m-R} \label{screening-rule-orign-MSSVM}
\end{align}
Suppose that we are given three subsets of $\calF, \calR$, and $\calL$, then we can infer the values of many coefficients of $\w^*(\alpha, \beta)$ and $\theta^*(\alpha, \beta)$ via Rule \ref{screening-rule-orign-MSSVM}. The lemma below shows that the rest coefficients of $\w^*(\alpha, \beta)$ and $\theta^*(\alpha, \beta)$ can be recovered by solving a small sized problem. 
\begin{lemma}\label{lemma:dual-scaled-MSSVM}
	Given index sets $\hat{\calF}\subseteq {\calF}, \hat{\calR}\subseteq \calR$, and $ \hat{\calL} \subset \calL$, the followings hold:\\
	$\rm{(i)}$ $[\w^*(\alpha, \beta)]_{\hat{\calF}}=0$, $[\theta^*(\alpha, \beta)]_{\hat{\calR}} = 0$, $[\theta^*(\alpha, \beta)]_{\hat{\calL}} = 1$;\\
	$\rm{(ii)}$ let $\hat{\calD} = \hat{\calR}\cup \hat{\calL}$, $\hat{\G}_{1} ={}_{\hat{\calF}^c} [\X]_{\hat{\calD}^c}$, and $\hat{\G}_2 = {}_{\hat{\calF}^c} [\X]_{\hat{\calL}}$, where $\hat{\calF}^c=[K]\times [p]\setminus\hat{\calF}$, $\hat{\calD}^c=[n]\times [K]\setminus\hat{\calD}$, $\hat{\calL}^c=[n]\times [K]\setminus\hat{\calL}$ and $\hat{\u} = [\u]_{\hat{\calD}^c}$, then, $[\theta^*(\alpha, \beta)]_{\hat{\calD}^c}$ solves the following scaled dual problem:
	\begin{align}
	\min_{\hat{\thetab} \in [0,1]^{|\hat{\calD}^c|}}\Big\{ \frac{1}{2\alpha}\left\|\calS_{\beta}\left(\frac{1}{n}\hat{\G}_{1} \hat{\thetab}+\frac{1}{n} \hat{\G}_{2} \mathbf{1}\right)\right\|^2+ \frac{\gamma}{2n}\|\hat{\theta}\|^2-\frac{1}{n}\langle \hat{\u}, \hat{\thetab}\rangle\Big\}; \tag{m-scaled-D$^*$}\label{eqn:dual-scaled-MMSVM}
	\end{align}
	$\rm{(iii)}$ suppose that $\theta^*(\alpha,\beta)$ is known, then,  
	\begin{align}
	[\w^*(\alpha,\beta)]_{\hcalF^c} = \frac{1}{\alpha}\calS_{\beta}\left(-\frac{1}{n}{}_{\hcalF^c}[\X]\theta^*(\alpha,\beta)\right).\nonumber
	\end{align} 
\end{lemma}

Given two estimations $\calW$ and $\Theta$ for $\w^*(\alpha, \beta)$ and $\theta^*(\alpha, \beta)$, we can define three subsets of $\calF, \calR$, and $\calL$ below as we did in the binary case to relax Rule \ref{screening-rule-orign-MSSVM} into the applicable version:
\begin{align*}
\hcalF &= \left\{(k,j) \in [K] \times [p]:\max_{\theta \in \Theta} \left \{\frac{1}{n}|[\X \thetab]_{k,j}| \right \} \leq \beta \right\},\\
\hcalR &=\left \{(i, k) \in [n]\times [K]:\max_{\w \in \calW} \left \{\langle \X_i^k,\w\rangle+[\u_i]_k\right \} \leq 0 \right \},\\
\hcalL &=\left \{(i, k) \in [n]\times [K]:\min_{\w \in \calW} \left \{\langle \X_i^k,\w^*(\alpha,\beta)\rangle+[\u_i]_k  \rangle \right \} \geq \gamma \right \}.
\end{align*}

\subsection{Estimate the Primal and Dual Optima}
We first derive the effective intervals of the parameters $\alpha$ and $\beta$. 

\begin{lemma} \label{beta-max-and-alpha-max-MSSVM}
	Let $\beta_{\rm max} = \|\frac{1}{n}\X\u\|_{\infty}$ and $	\alpha_{\rm max}(\beta)=\frac{1}{1-\gamma}\max_{(i,k) \in [n]\times [K]} \big\{ \langle \X_i^k, \calS_{\beta}(\frac{1}{n} \X \u)\rangle\big\}$. Then:\\
	\textup{(i)} for $\alpha>0$ and $\beta \geq \beta_{\rm max}$, we have
	\begin{align}
	&\w^*(\alpha, \beta) = \mathbf{0}, \hspace{2mm} \thetab^*(\alpha, \beta) = \u;\nonumber
	\end{align}
	\textup{(ii) }for all $\alpha \in [\max\{\alpha_{\rm max}(\beta),0\}, \infty)\cap(0,\infty)$,  we have 
	\begin{align}
	\w^*(\alpha, \beta) = \frac{1}{\alpha}\calS_{\beta}\left(-\frac{1}{n}\X \u\right),\hspace{2mm}\thetab^*(\alpha, \beta) =  \u. \label{eqn:close-form-solution-MSSVM} 
	\end{align}  
\end{lemma}

Hence, we only need to consider the cases with $\beta \in (0, \beta_{\max}]$ and $\alpha \in (0, \alpha_{\max}(\beta)]$. 

Since problem (\ref{eqn:primal-MSSVM}) is a special case of problem (\ref{eqn:general-primal}) with $L(\w) = \frac{1}{m}\sum_{i=1}^n \ell_i(\w)$, given the reference solution $\w^*(\alpha_0,\beta_0)$ and the index set $\hcalF$ of the inactive features identified by the previous \textup{IFS} steps, we can apply Lemma \ref{lemma:primal-estimation} to obtain the estimation for $\w^*(\alpha, \beta_0)]_{\hcalF^c}$:
	\begin{align}
	[\w^*(\alpha, \beta_0)]_{\hcalF^c}\in\calW\coloneqq\{\mathbf{w}:\|\mathbf{w}- \mathbf{c}\| \leq r\},\label{eqn:primal-estimation-MSSVM}
	\end{align}
	where $\c$ and $r$ are defined in Eqs. (\ref{eqn:c-primal}) and (\ref{eqn:r-primal}), respectively.

Moreover, noting that problem (\ref{eqn:dual-MSSVM}) is also a special case of problem (\ref{eqn:general-dual}), given the reference solution $\theta^*(\alpha_0,\beta_0)$ and the index sets of inactive samples identified by the previous \textup{ISS} steps $\hcalR$ and $\hcalL$, we can obtain an estimation for $[\theta^*(\alpha,\beta_0)]_{\hcalD^c}$ from Lemma \ref{lemma:dual-estimation}:
 \begin{align}
 [\theta^*(\alpha,\beta_0)]_{\hcalD^c}\in\Theta\coloneqq \{\theta: \| \theta - \mathbf{c}\|\leq r\}, \label{eqn:dual-estimation-MSSVM}
 \end{align}
 where $\c$ and $r$ are defined by Eqs. (\ref{c-dual}) and (\ref{r-dual}) with $\v=\u$, respectively.

\subsection{Screening Rule SIFS}
Given the optima $\w^*(\alpha_0, \beta_0)$ and $\theta^*(\alpha_0, \beta_0)$, to derive the feature screening rule IFS, we need to solve the optimization problem below first: 
\begin{align}
s^{(k,j)}(\alpha, \beta_0) = \max_{\theta \in \Theta} \left \{\frac{1}{n}| \langle [\X_{k,j}]_{\hcalD^c}, \thetab \rangle + \langle [\X_{k,j}]_{\hcalL}, \mathbf{1} \rangle| \right \}, (k,j) \in \hcalF^c, \label{eqn:opt-feature-MSSVM}
\end{align}
where $\X_{k,j}$ is the $((k-1)p+j)$-th row of $\X$, $\Theta$ is given by \eqref{eqn:dual-estimation-MSSVM}, and $\hcalF$ and $\hcalD=\hcalR \cup \hcalL$ are the index sets of the inactive features and samples identified in the previous screening process. We notice that this problem has exactly the same form as problem (\ref{eqn:opt-feature}). Hence, from  Lemma \ref{lemma:opt-feature} we can obtain its closed-form solution directly. 
Now, from Rule \ref{screening-rule-orign-MSSVM}, we can obtain our IFS rule below.
\begin{itemize}
	\item The feature screening rule \textup{IFS} takes the form of 
	\begin{align}
	s^{(k,j)}(\alpha,\beta_0) \leq \beta_0\Rightarrow[\w_k^*(\alpha,\beta_0)]_{j} = 0, \forall (k,j) \in \hcalF^c. \tag{IFS}\label{eqn:feature-screening-MSSVM} 
	\end{align} 
	\item  The index set $\hcalF$ can be updated by
	\begin{align}
	\hcalF \leftarrow \hcalF \cup \Delta \hcalF \mbox{ with } \Delta \hcalF= \{(k,j): s^{(k,j)} \leq \beta_0, (k,j) \in \hcalF^c \}. \label{eqn:update-F-MSSVM}
	\end{align} 
\end{itemize}


Similarly, we need to solve the following problems to derive our sample screening rule ISS: 
\begin{align}
u_{(i,k)}(\alpha, \beta_0) = \max_{\w \in \calW} \{\langle [\X_i^k]_{\hcalF^c}, \w \rangle +[\u_i]_k\}, (i,k)\in \hcalD^c,\label{eqn:opt-sample-1-MSSVM}\\
l_{(i,k)}(\alpha, \beta_0) = \min_{\w \in \calW} \{\langle [\X_i^k]_{\hcalF^c}, \w \rangle +[\u_i]_k \}, (i,k)\in \hcalD^c, \label{eqn:opt-sample-2-MSSVM}
\end{align}
where $\calW$ is given by \eqref{eqn:primal-estimation-MSSVM}. We find that they can be solved by Lemma \ref{lemma:opt-sample} directly. Therefore, we can obtain our sample screening rule ISS below.
	\begin{itemize}
	\item The sample screening rule \textup{ISS} takes the form of 
	\begin{align}
	\begin{array}{cc}
	u_{(i,k)}(\alpha, \beta_0) \leq 0\Rightarrow[\theta_i^*(\alpha, \beta_0)]_k=0,\\
	l_{(i,k)}(\alpha, \beta_0) \geq  \gamma\Rightarrow[\theta_i^*(\alpha, \beta_0)]_k=1,
	\end{array}\forall (i,k) \in \hcalD^c. \tag{ISS} \label{eqn:sample-screening-MSSVM}
	\end{align} 
	\item We can update the index sets $\hcalR$ and $\hcalL$ by
	\begin{align}
	\hcalR &\leftarrow \hcalR \cup \Delta \hcalR \mbox{ with } \Delta \hcalR=\{(i,k): u_{(i,k)}(\alpha, \beta_0) \leq 0, (i,k) \in \hcalD^c \}, \label{eqn:update-R-MSSVM}\\
	\hcalL &\leftarrow \hcalL \cup \Delta \hcalL \mbox{ with } \Delta \hcalL = \{(i,k): l_{(i,k)}(\alpha, \beta_0) \geq \gamma, (i,k) \in \hcalD^c \}. \label{eqn:update-L-MSSVM}
	\end{align}
   \end{itemize}

The same as we did sparse SVM, we can also develop SIFS to reinforce the capabilities of ISS and IFS by applying them alternatively.  The framework of SIFS for solving the model over a grid of parameter values here is the same as that in the case of sparse SVM, i.e. Algorithm \ref{alg:framework-screening}, except for the different rules IFS and ISS, and the updates of $\hcalR, \hcalL$, and $\hcalF$.

In this version of SIFS, the order of applying IFS and ISS also has no impact on the final performance. Since the form of the theorem and its proof are nearly the same as that of Theorem \ref{thm:order-of-screening}, to avoid redundancy, we omit them in this section. 

\section{Experiments}\label{sec:experiments}
We evaluate SIFS on both synthetic and real data sets in terms of three measurements: \emph{speedup}, \emph{scaling ratio}, and \emph{rejection ratio}. Speedup is given by the ratio of the running time of the solver without screening to that with screening.

For sparse SVMs, scaling ratio is defined as $1-\frac{(n-\tilde{n})(p-\tilde{p})}{np}$, where $\tilde{n}$, $\tilde{p}$, $n$, and $p$ are the numbers of inactive samples and features identified by SIFS, sample size, and feature dimension of the data set. From steps 6 to 11 in Algorithm \ref{alg:framework-screening}, we can see that we trigger the rules ISS and IFS repetitively until no new features and samples are identified. We adopt the rejection ratios of the $i$-th triggering of ISS and IFS defined as  $\frac{\tilde{n}_i}{n_0}$ and $\frac{\tilde{p}_i}{p_0}$ to evaluate their performances in each triggering, where $\tilde{n}_i$ and $\tilde{p}_i$ are the numbers of inactive samples and features identified in the $i$-th triggering of ISS and IFS. $n_0$ and $p_0$ are the numbers of inactive samples and features.   

For multi-class sparse SVMs, we let scaling ratio be $1-\frac{(Kn-\tilde{n})(Kp-\tilde{p})}{K^2np}$, where $\tilde{n}=|\hcalR\cup\hcalL|$, $\tilde{p} = |\hcalF|$, and $n$ and $p$ are the sample size and the feature dimension of the data set. The rejection ratios of the $i$-th triggering of ISS and IFS are $\frac{\tilde{n}_i}{n_0}$ and $\frac{\tilde{p}_i}{p_0}$, respectively, where $p_0=|\calF|$, $n_0=|\calR \cup \calL|$, $\tilde{p}_i=|\Delta \hcalF|$ and $\tilde{n}_i=|\Delta \hcalR \cup \Delta \hcalL|$ with $\Delta \hcalF, \Delta \hcalR$, and $\Delta \hcalL$ are the increments of $\hcalF, \hcalR$ and $ \hcalL$ in the $i$-th  triggering of the rules IFS and ISS. Please see Eqs.~(\ref{eqn:update-F-MSSVM}),~(\ref{eqn:update-R-MSSVM}), and (\ref{eqn:update-L-MSSVM}) for the detailed definitions of $\Delta \hcalF, \Delta \hcalR$, and $\Delta \hcalL$. 

Recall that, we can integrate SIFS with any solvers for problem (\ref{eqn:primal}). In this experiment, we use Accelerated Proximal Stochastic Dual Coordinate Ascent (AProx-SDCA) \citep{shalev2016accelerated} as a baseline, as it is one of the state-of-the-arts. We choose the state-of-art screening method for sparse SVMs in \citealt{shibagaki2016simultaneous} as another baseline only in the experiments of sparse SVMs, since it cannot be applied in multi-class sparse SVMs.  As we mentioned in the introduction section that screening differs greatly from feature selection methods, it is not appropriate to make comparisons with feature selection methods.  

For each data set,  we solve problem (\ref{eqn:primal}) at a grid of turning parameter values. Specifically, we first compute $\beta_{\rm max}$ by Lemma \ref{beta-max-and-alpha-max} and then select 10 values of $\beta$ that are equally spaced at the logarithmic scale of $\beta/\beta_{\rm max}$ from $1$ to $0.05$. Then, for each value of $\beta$, we first compute $\alpha_{\rm max}(\beta)$ by Lemma \ref{beta-max-and-alpha-max} and then select $100$ values of $\alpha$ that are equally spaced at the logarithmic scale of $\alpha/\alpha_{\rm max}(\beta)$ from $1$ to $0.01$. Thus, for each data set, we solve problem (\ref{eqn:primal}) at $1,000$ pairs of parameter values in total. We write the code in C++ along with Eigen library for numerical computations. We perform all the computations on a single core of Intel(R) Core(TM) i7-5930K 3.50GHz, 128GB MEM.
\subsection{Simulation Studies with Spare SVMs}\label{sec:experiment-simu-binary}
We evaluate SIFS on 3 synthetic data sets named syn1, syn2, and syn3 with sample and feature size $(n,p)\in\{(10000,1000), (10000,10000), (1000,10000)\}$.
We present each data point as $\x=[\x_{1}; \x_{2}]$ with $\x_{1}\in \R^{0.02p}$ and $\x_{2}\in \R^{0.98p}$. We use Gaussian distributions $\calG_1 = N(\u, 0.75\I), \calG_2 = N(-\u,0.75\I)$, and $\calG_3 = N(0, 1)$ to generate the data points, where $\u = 1.5\textbf{1} $ and $\I\in \R^{0.02p\times 0.02p}$ is the identity matrix. To be precise, $\x_{1}$ for positive and negative points are sampled from $\calG_1$ and $\calG_2$, respectively. For each entry in $\x_{2}$, it has chance $\eta = 0.02$ to be sampled from $\calG_3$ and chance $1-\eta$ to be 0. 
\begin{figure}[htb!]
	\centering
	\subfigure[syn1.]{\includegraphics[width=.8\textwidth]{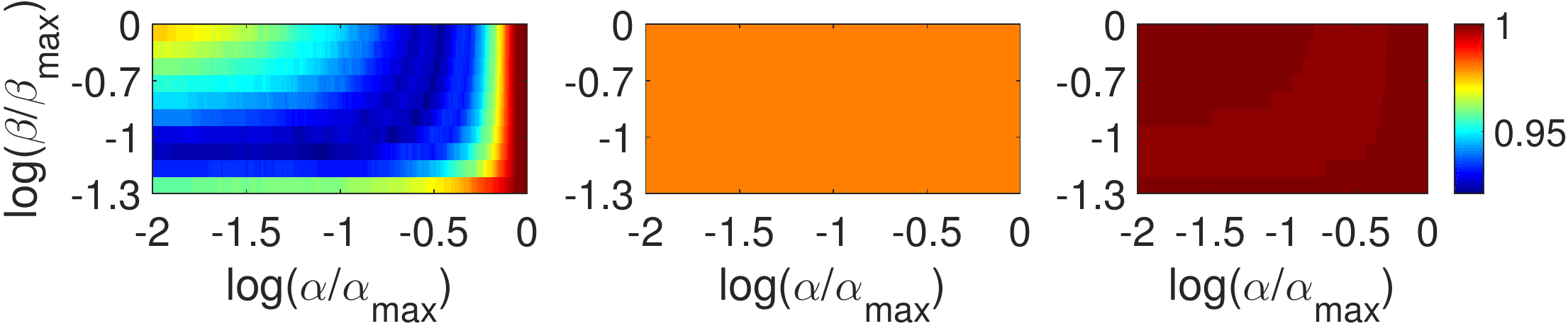}}
	\subfigure[syn2.]{\includegraphics[width=.8\textwidth]{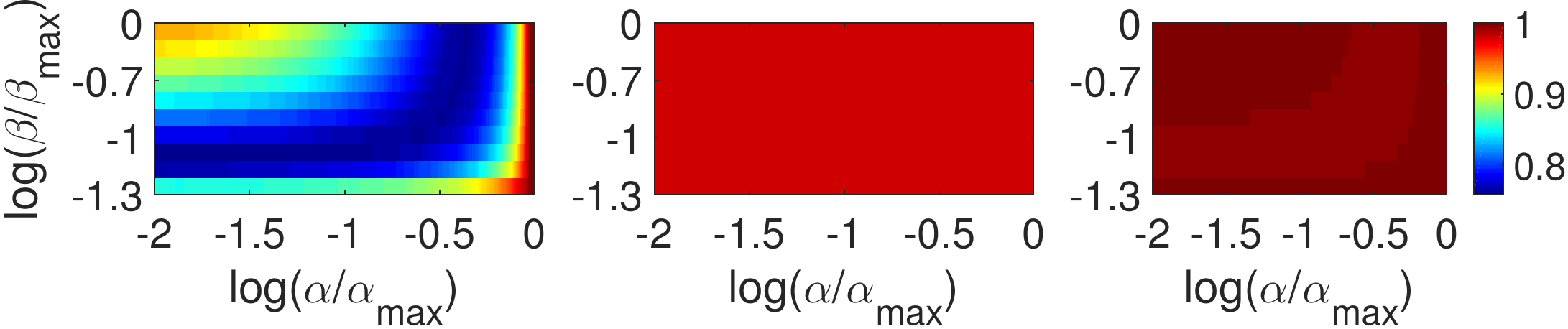}}
	\subfigure[syn3.]{\includegraphics[width=.8\textwidth]{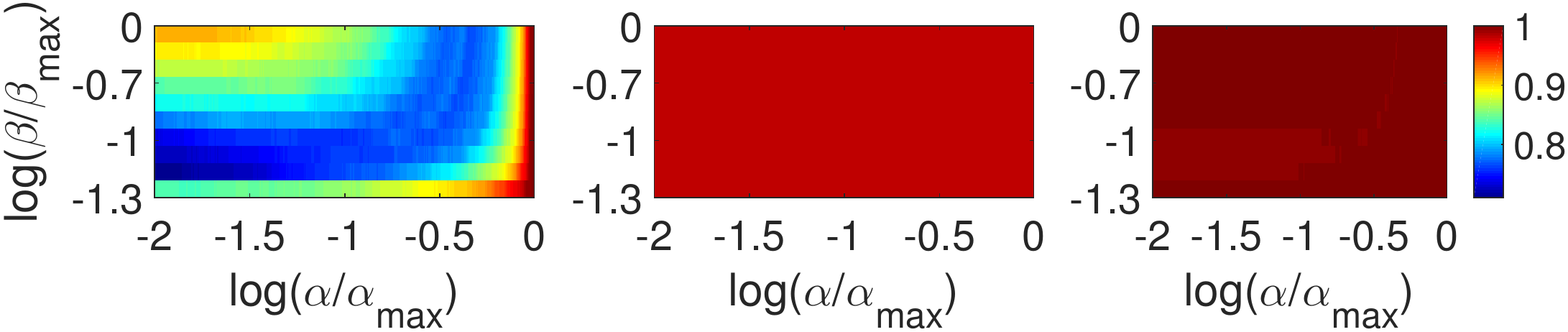}}
	\caption{Scaling ratios of ISS, IFS, and SIFS (from left to right).}\label{fig:scaling-ratio-sync}
\end{figure}

\figref{fig:scaling-ratio-sync} shows the scaling ratios by ISS, IFS, and SIFS on the synthetic data sets at $1,000$ parameter values. We can see that IFS is more effective in scaling problem size than ISS, with scaling ratios roughly $98\%$ against $70-90\%$. Moreover, SIFS, which is an alternating application of IFS and ISS, significantly outperforms ISS and IFS, with scaling ratios roughly $99.9\%$. This high scaling ratios imply that SIFS can lead to a significant speedup.

Due to the space limitation, we only report the rejection ratios of SIFS on syn2. Other results can be found in the appendix. \figref{fig:rejection-ratio-syn-2} shows that SIFS can identify most of the inactive features and samples. However, few features and samples are identified in the second and later triggerings of ISS and IFS. The reason may be that the task here is so simple that one triggering is enough.


Table \ref{table:run-time-sync} reports the running times of solver without and with IFS, ISS and SIFS for solving problem (\ref{eqn:primal}) at $1,000$ pairs of parameter values. We can see that SIFS leads to significant speedups, that is, up to $76.8$ times. Taking syn2 for example, without SIFS, the solver takes more than two hours to solve problem (\ref{eqn:primal}) at $1,000$ pairs of parameter values. However, combined with SIFS, the solver only needs less than three minutes for solving the same set of problems. From the theoretical analysis in \citealt{shalev2016accelerated} for AProx-SDCA, we can see that its computational complexity  rises proportionately to the sample size $n$ and the feature dimension $p$. From this theoretical result, we can see that the results in \figref{fig:scaling-ratio-sync} are roughly consistent with the speedups we achieved shown in Table \ref{table:run-time-sync}.
\begin{figure*}[htb!]
	\begin{center}
		\subfigure[ $\beta/\beta_{\rm{max}}$=0.05]{\includegraphics[scale=0.25]{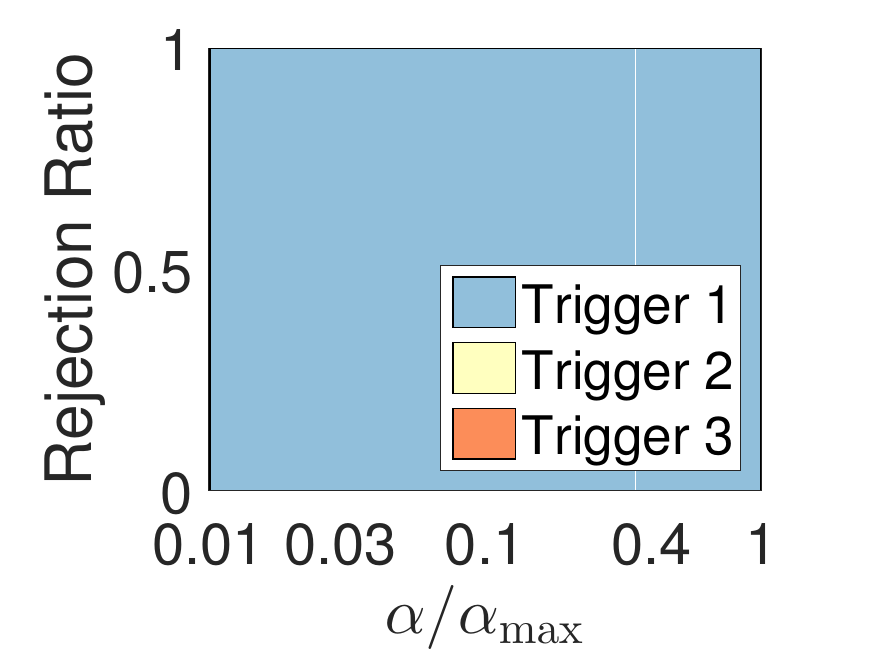}}
		\subfigure[ $\beta/\beta_{\rm{max}}$=0.1]{\includegraphics[scale=0.25]{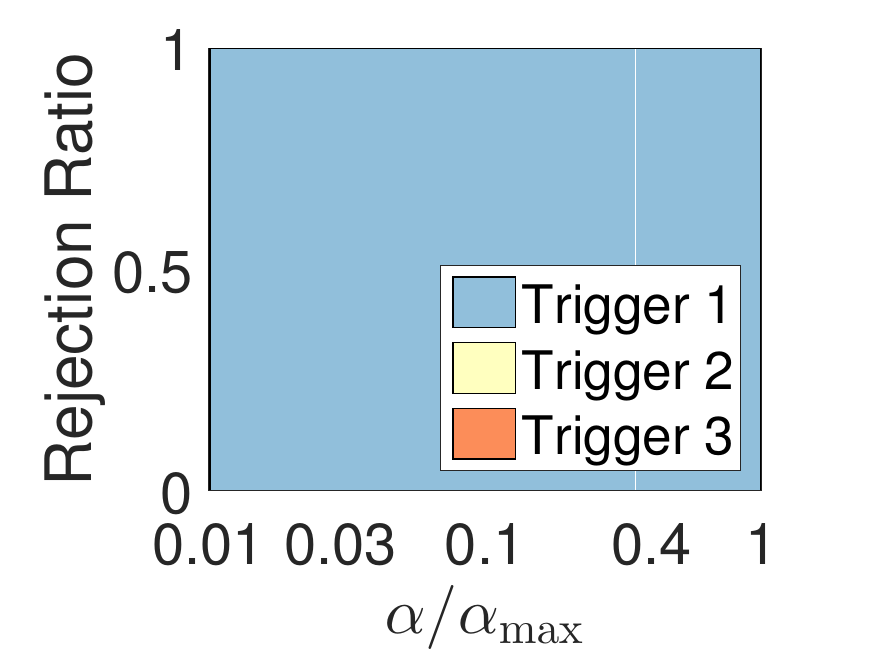}}
		\subfigure[ $\beta/\beta_{\rm{max}}$=0.5]{\includegraphics[scale=0.25]{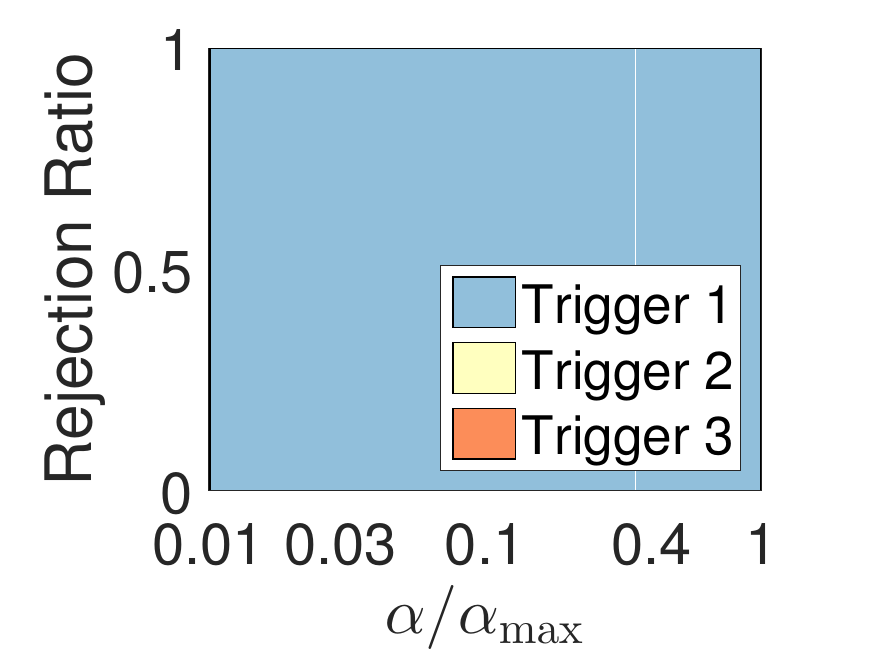}}
		\subfigure[ $\beta/\beta_{\rm{max}}$=0.9]{\includegraphics[scale=0.25]{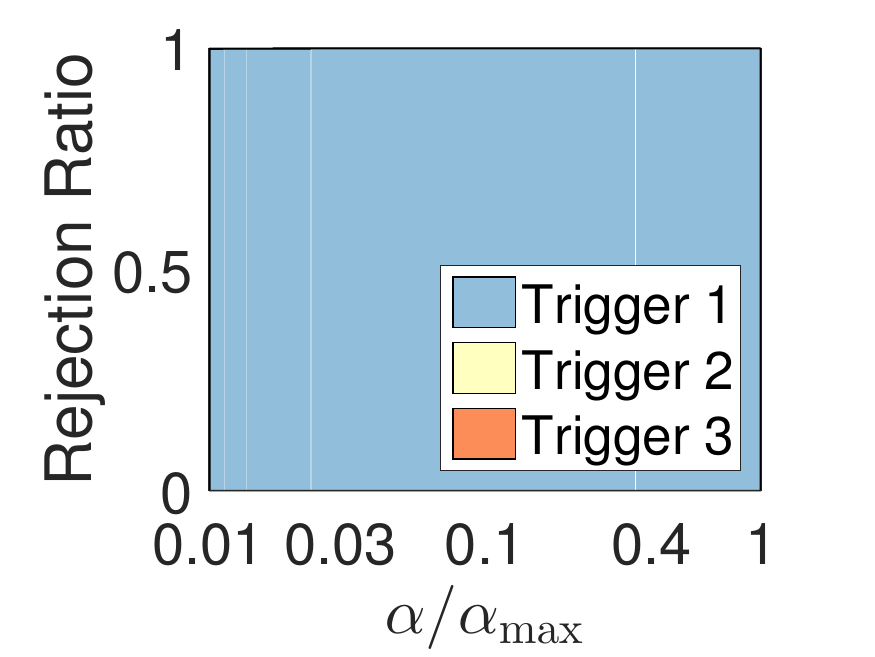}}
		\subfigure[ $\beta/\beta_{\rm{max}}$=0.05]{\includegraphics[scale=0.25]{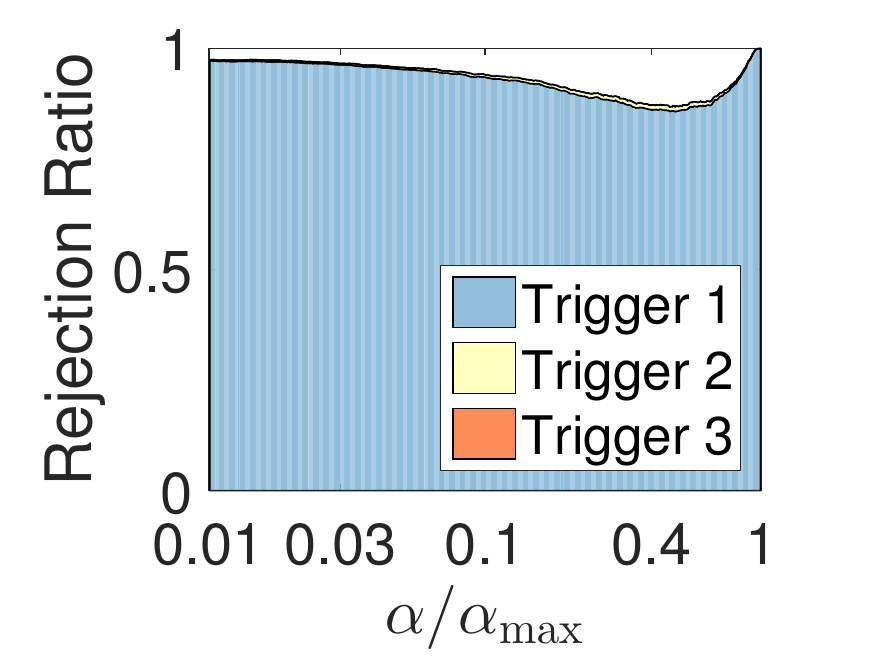}}
		\subfigure[ $\beta/\beta_{\rm{max}}$=0.1]{\includegraphics[scale=0.25]{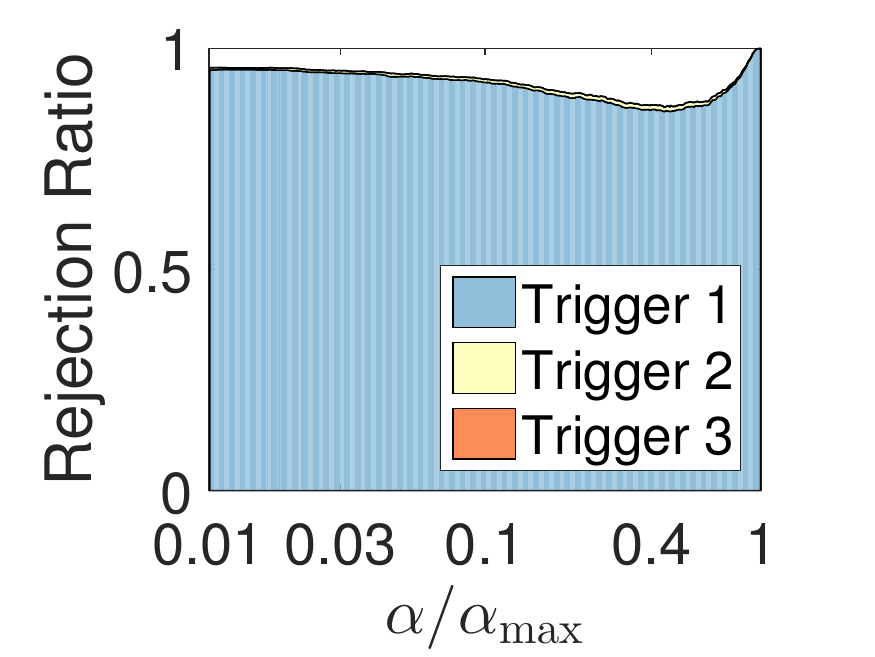}}
		\subfigure[ $\beta/\beta_{\rm{max}}$=0.5]{\includegraphics[scale=0.25]{images/toy_n_10000_d_100000_05_RL_rej_ratio}}
		\subfigure[ $\beta/\beta_{\rm{max}}$=0.9]{\includegraphics[scale=0.25]{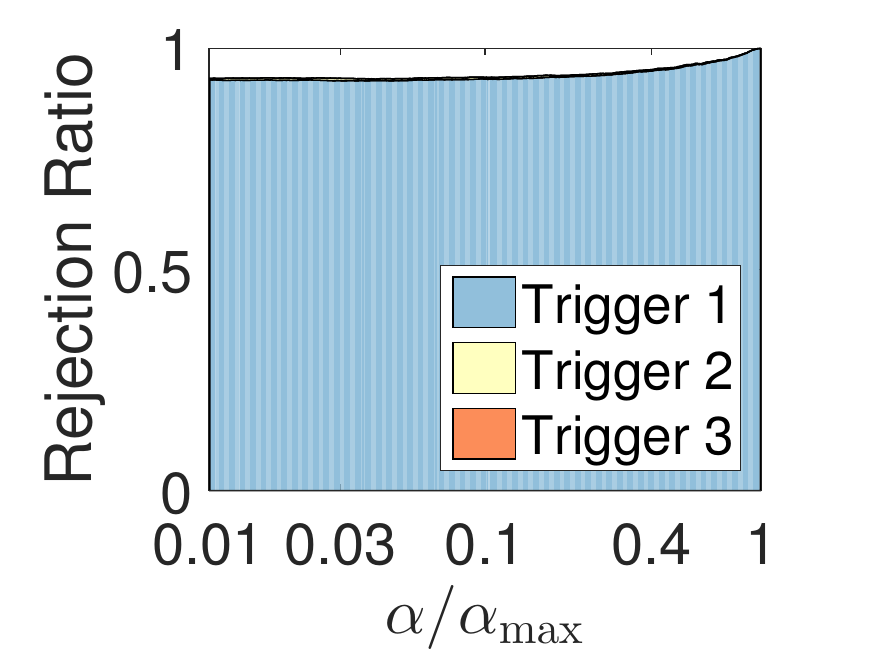}}
		\caption{Rejection ratios of SIFS on syn 2 (first row: Feature Screening, second row: Sample Screening). }
		\label{fig:rejection-ratio-syn-2}
	\end{center}
	\vspace*{-30pt}
\end{figure*}

\begin{table*}[htb!]
	\vspace*{-5pt}
	\centering
	{\footnotesize
		\begin{tabular}{|c|c|c|c|c|c|c|c|c|c|c|}
			\hline
			\multirow{2}{*}{Data} & \multirow{2}{*}{Solver} & \multicolumn{3}{c|}{ISS+Solver} & \multicolumn{3}{c|}{IFS+Solver} & \multicolumn{3}{c|}{SIFS+Solver} \\ \cline{3-11} 
			&  & ISS & Solver & Speedup & IFS & Solver & Speedup & SIFS & Solver & Speedup \\ \hline
			syn1 &499.1&4.9&27.8&15.3&2.3&42.6&11.1&8.6&6.0&\textbf{34.2} \\ \hline
			syn2 &8749.9&24.9&1496.6&5.8&23.0&288.1&28.1&92.6&70.3&\textbf{53.7} \\ \hline
			syn3 &1279.7&2.0&257.1&4.9&2.2&33.4&36.0&7.2&9.5&\textbf{76.8} \\ \hline
		\end{tabular}
	}
	\caption{Running time (in seconds) for solving problem (\ref{eqn:primal}) at $1,000$ pairs of parameter values on three synthetic data sets. }\label{table:run-time-sync}
	\vspace*{-15pt}
\end{table*}



\subsection{Experiments with Sparse SVMs on Real Datasets}
In this experiment, we evaluate the performance of SIFS on 5 large-scale real data sets: real-sim, rcv1-train, rcv1-test, url, and kddb, which are all collected from the project page of LibSVM \citep{chang2011libsvm}. See Table \ref{data_description} for a brief summary. We note that, the kddb data set has about 20 million samples with 30 million features.

Recall that,  SIFS detects the inactive features and samples in a static manner, i.e., we perform SIFS only once before the training process and hence the size of the problem we need to perform optimization on is fixed. However, the method in \citealt{shibagaki2016simultaneous} detects inactive features and samples in a dynamic manner \citep{bonnefoy2014dynamic}, i.e., they perform their method along with the training process  and thus the size of the problem would keep decreasing during the iterations. Thus, comparing SIFS with this baseline in terms of the rejection ratios is inapplicable. We compare the performance of SIFS with the method in \citealt{shibagaki2016simultaneous} in terms of speedup, i.e., the speedup gained by these two methods in solving problem (\ref{eqn:primal}) at $1,000$ pairs of parameter values. The code of the method in \citealt{shibagaki2016simultaneous} is obtained from (\url{https://github.com/husk214/s3fs}).
\begin{table*}[htb!]
	\centering
	\begin{footnotesize}
		\begin{tabular}{|l|l|l|c|}
			\hline
			Dataset & Feature size: $p$ & Sample size:$n$ & Classes \\ \hline
			real-sim & 20,958 & 72,309 & 2\\ \hline
			rcv1-train&47,236&20,242 & 2 \\ \hline
			rcv1-test & 47,236 & 677, 399 & 2\\ \hline
			url& 3,231,961& 2,396,130 & 2\\ \hline
			kddb& 29,890,095&19,264,097 & 2\\ \hline
		\end{tabular}
	\end{footnotesize}
	\caption{Statistics of the binary real data sets.}\label{data_description}
	\vspace*{-5pt}
\end{table*}

\begin{figure*}[htb!]
	\begin{center}
		\subfigure[ $\beta/\beta_{\rm{max}}$=0.05]{\includegraphics[scale=0.25]{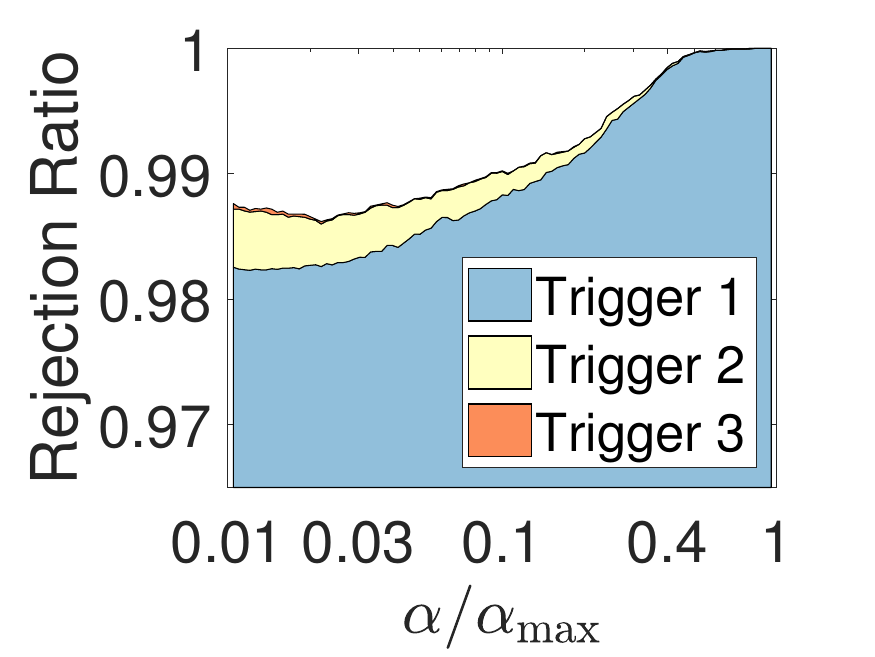}}
		\subfigure[ $\beta/\beta_{\rm{max}}$=0.1]{\includegraphics[scale=0.25]{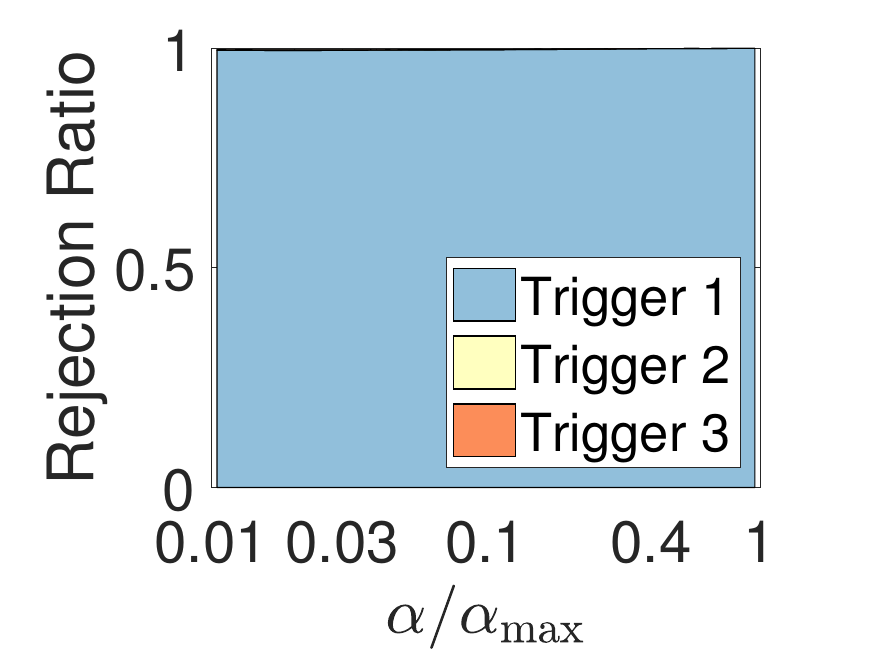}}
		\subfigure[ $\beta/\beta_{\rm{max}}$=0.5]{\includegraphics[scale=0.25]{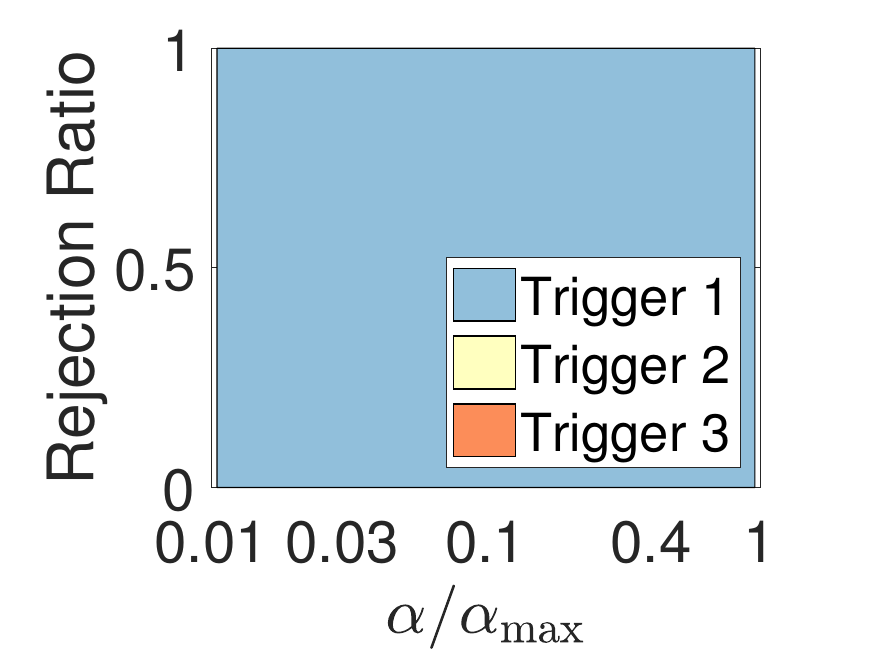}}
		\subfigure[ $\beta/\beta_{\rm{max}}$=0.9]{\includegraphics[scale=0.25]{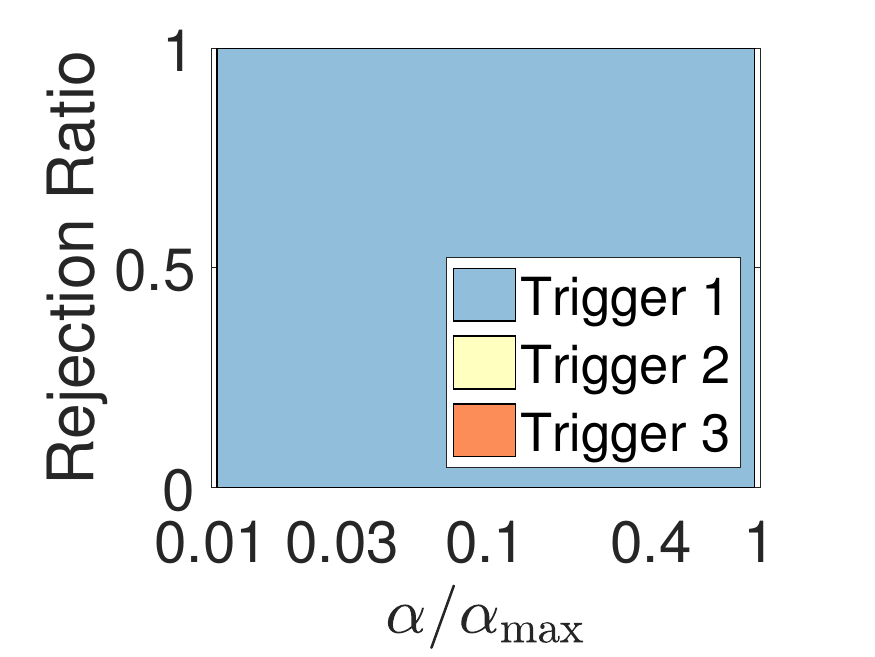}}
		\subfigure[ $\beta/\beta_{\rm{max}}$=0.05]{\includegraphics[scale=0.25]{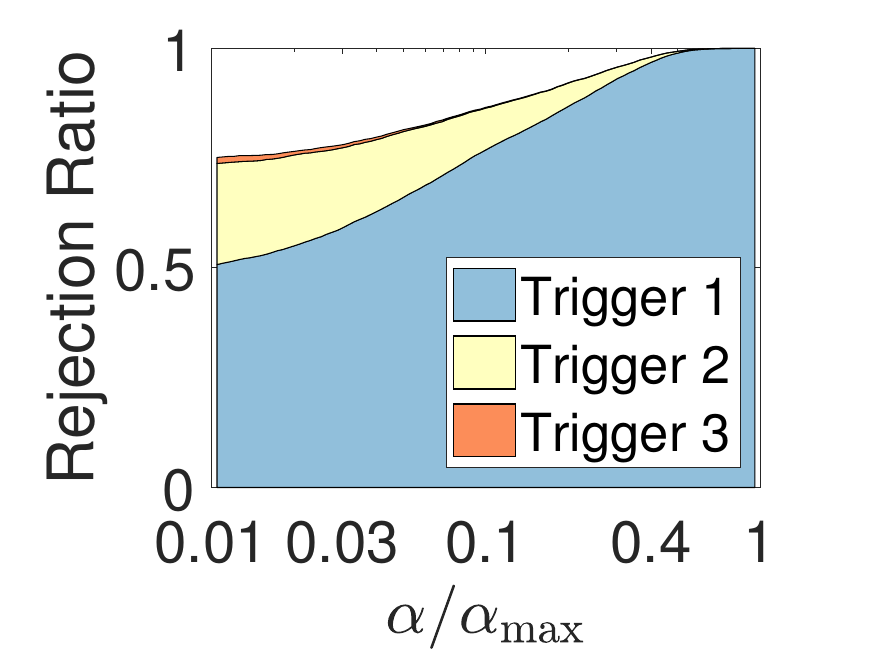}}
		\subfigure[ $\beta/\beta_{\rm{max}}$=0.1]{\includegraphics[scale=0.25]{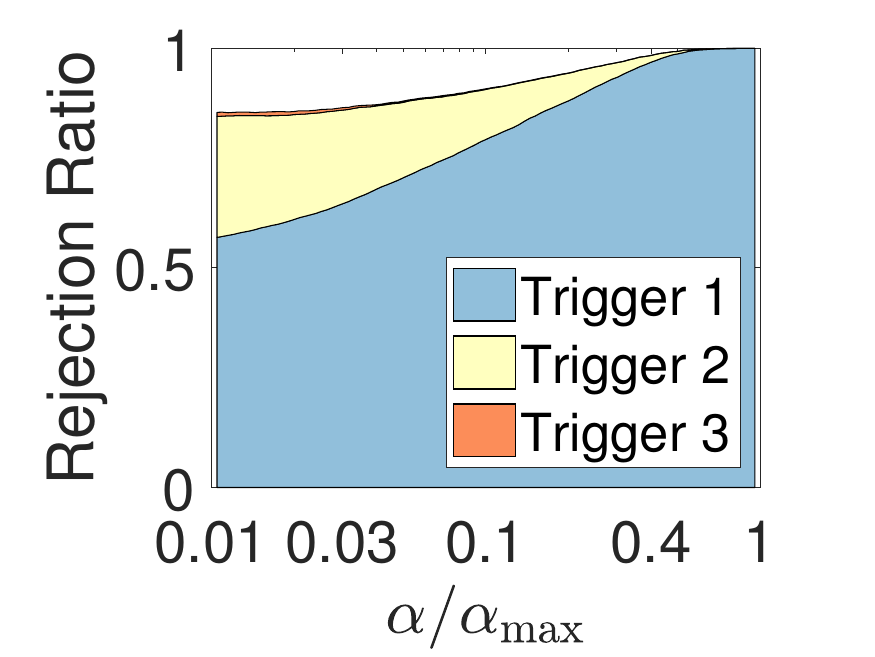}}
		\subfigure[ $\beta/\beta_{\rm{max}}$=0.5]{\includegraphics[scale=0.25]{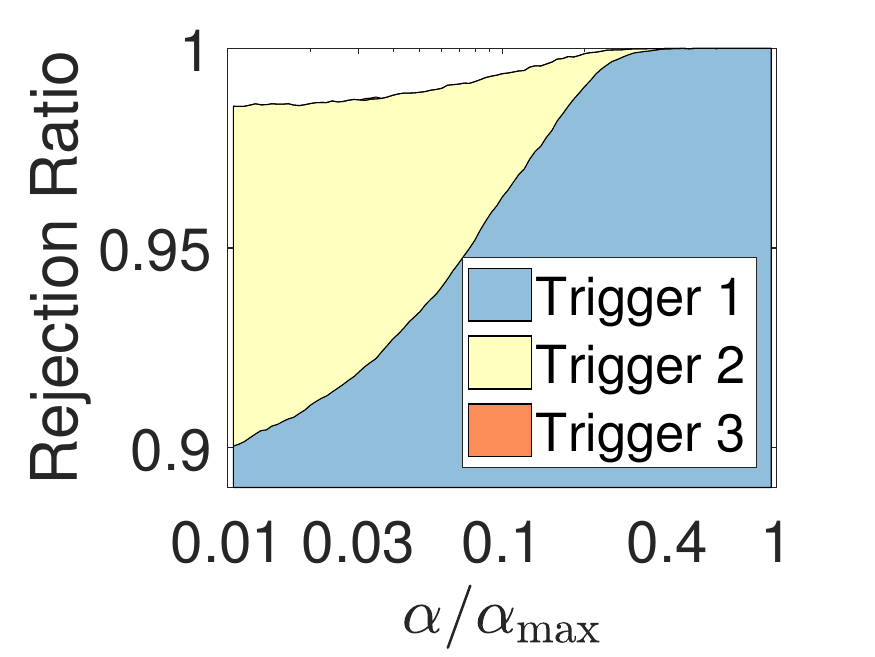}}
		\subfigure[ $\beta/\beta_{\rm{max}}$=0.9]{\includegraphics[scale=0.25]{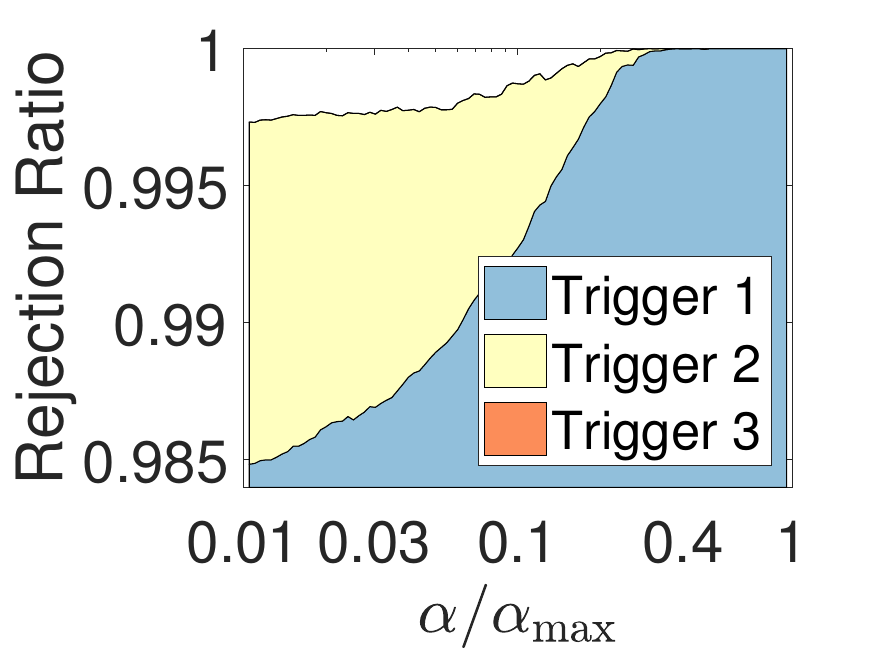}}
		\caption{Rejection ratios of SIFS on real-sim when it performs \textbf{ISS first} (first row: Feature Screening, second row: Sample Screening). }
		\label{fig:rejection-ratio-real-sim-sample-first}
	\end{center}
	\vspace*{-20pt}
\end{figure*}

\figref{fig:rejection-ratio-real-sim-sample-first} shows the rejection ratios of SIFS on the real-sim data set (other results are in the appendix). In \figref{fig:rejection-ratio-real-sim-sample-first}, we can see that some inactive features and samples are identified in the 2nd and 3rd triggerings of ISS and IFS, which verifies the necessity of the alternating application of ISS and IFS. SIFS is efficient since it always stops in 3 times of triggering. In addition, most of ($> 98\%$) the inactive features can be identified in the 1st triggering of IFS while identifying inactive samples needs to apply ISS two or more times. It may due to two reasons: 1) we run ISS first, which reinforces the capability of IFS due to the synergistic effect (see Sections \ref{ssec:IFS} and \ref{ssec:ISS}), referring to the analysis below for further verification; 2) feature screening here may be easier than sample screening.

Table \ref{Table:speedup-real} reports the running times of solver without and with the method in \citealt{shibagaki2016simultaneous} and SIFS for solving problem (\ref{eqn:primal}) at $1,000$ pairs of parameter values on real data sets. The speedup gained by SIFS is up to $300$ times on real-sim, rcv1-train and rcv1-test. Moreover, SIFS significantly outperforms the method in \citealt{shibagaki2016simultaneous} in terms of speedup---by about $30$ to $40$ times faster on the aforementioned three data sets. For data sets url and kddb, we do not report the results of the solver as the sizes of the data sets are huge and the computational cost is prohibitive. Instead, we can see that the solver with SIFS is about $25$ times faster than the solver with the method in \citealt{shibagaki2016simultaneous} on both data sets url and kddb. Let us take the data set kddb as an example. The solver with SIFS takes about $13$ hours to solve problem (\ref{eqn:primal}) for $1,000$ pairs of parameter values, while the solver with the method in \citealt{shibagaki2016simultaneous} needs $11$ days to finish the same task.

\begin{table*}[htb!]
	\centering
	{\footnotesize
		\begin{tabular}{|p{1.5cm}<{\centering}|p{1.4cm}<{\centering}|p{1.7cm}<{\centering}|p{1.7cm}<{\centering}|p{1.15cm}<{\centering}|p{1.2cm}<{\centering}|p{1.2cm}<{\centering}|p{1.2cm}<{\centering}|}
			\hline
			\multirow{2}{*}{\begin{tabular}[c]{@{}c@{}}Data\\ Set\end{tabular}} & \multirow{2}{*}{Solver} & \multicolumn{3}{p{5.9cm}<{\centering}|}{Method in \citealt{shibagaki2016simultaneous}+Solver} & \multicolumn{3}{p{3.6cm}<{\centering}|}{SIFS+Solver} \\ \cline{3-8} 
			&  & Screen & Solver & Speedup & Screen & Solver & \textbf{Speedup} \\ \hline
			real-sim & 3.93E+04&24.10&4.94E+03&7.91&60.01&140.25&\textbf{195.00}\\ \hline
			rcv1-train &2.98E+04	&10.00	&3.73E+03&7.90 &27.11&80.11&\textbf{277.10}  \\ \hline
			rcv1-test& 1.10E+06&398.00&1.35E+05&8.10&1.17E+03&2.55E+03&\textbf{295.11}\\ \hline
			url&$>$3.50E+06&3.18E+04&8.60E+05&$>$4.00&7.66E+03&2.91E+04&\textbf{$>$100}\\\hline
			kddb &$>$5.00E+06&4.31E+04&1.16E+06&$>$4.00&1.10E+04&3.6E+04&\textbf{$>$100}\\ \hline
		\end{tabular}
	}
	\caption{Running time (in seconds) for solving problem (\ref{eqn:primal}) at $1,000$ pairs of parameter values on five real data sets.}	\label{Table:speedup-real}
	\vspace*{-5pt}
\end{table*}

\begin{figure*}[htb!]
	\begin{center}
		\subfigure[ $\beta/\beta_{\rm{max}}$=0.05]{\includegraphics[scale=0.25]{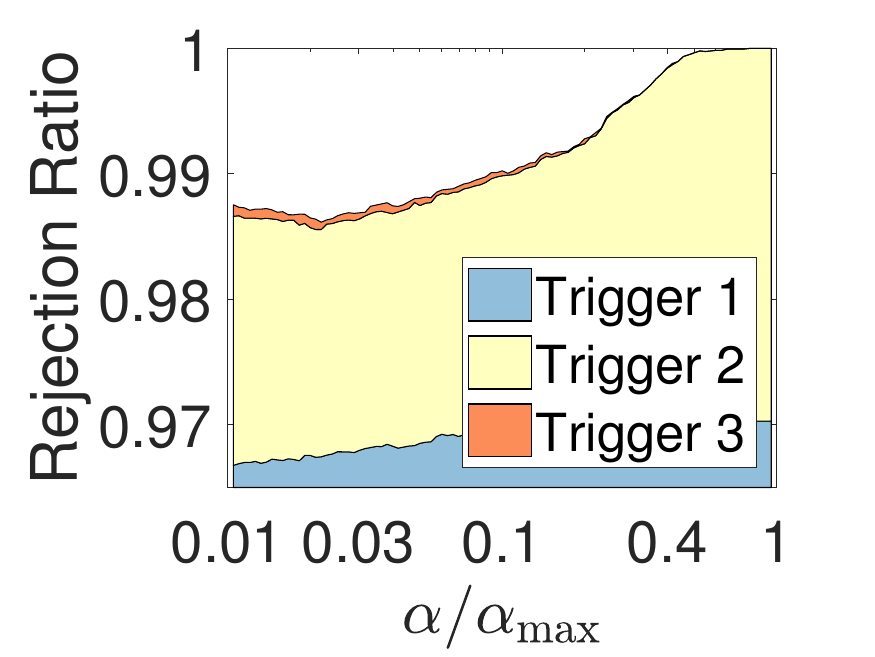}}
		\subfigure[ $\beta/\beta_{\rm{max}}$=0.1]{\includegraphics[scale=0.25]{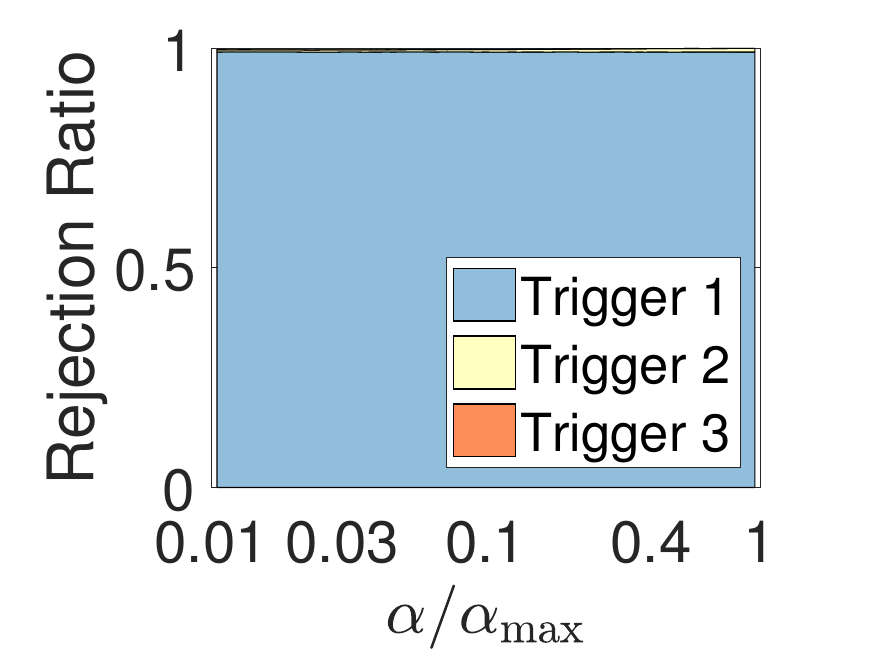}}
		\subfigure[ $\beta/\beta_{\rm{max}}$=0.5]{\includegraphics[scale=0.25]{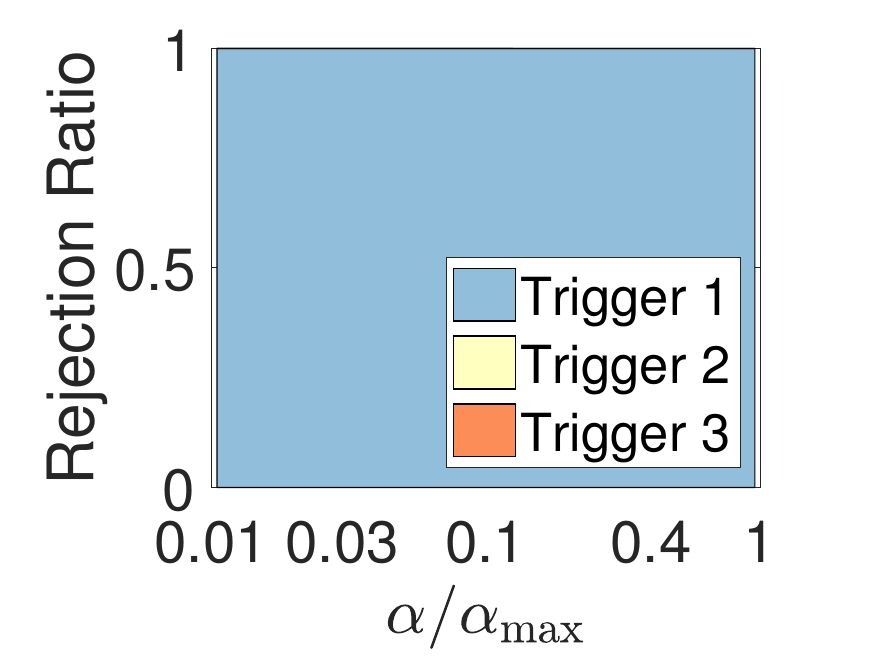}}
		\subfigure[ $\beta/\beta_{\rm{max}}$=0.9]{\includegraphics[scale=0.25]{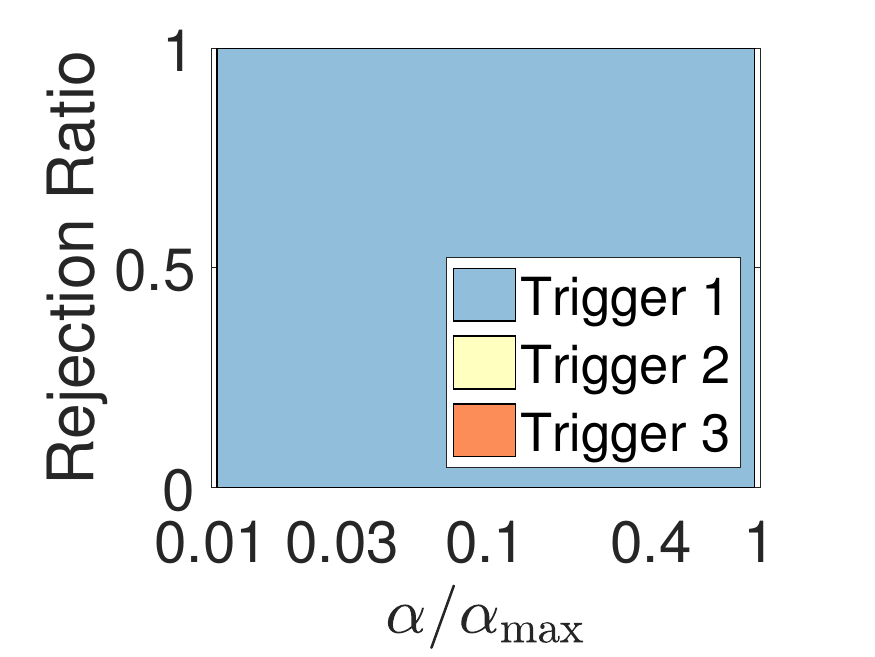}}
		\subfigure[ $\beta/\beta_{\rm{max}}$=0.05]{\includegraphics[scale=0.25]{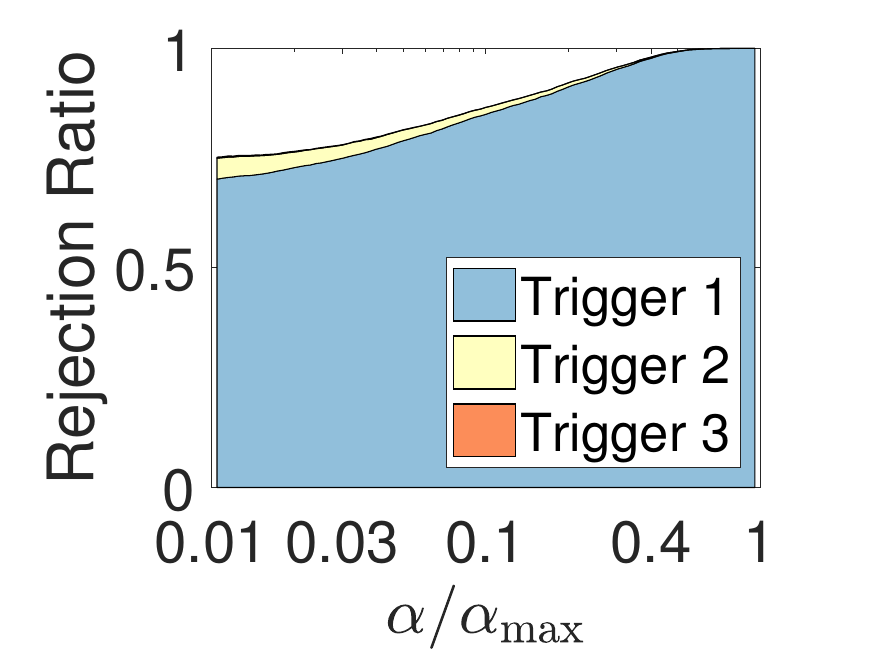}}
		\subfigure[ $\beta/\beta_{\rm{max}}$=0.1]{\includegraphics[scale=0.25]{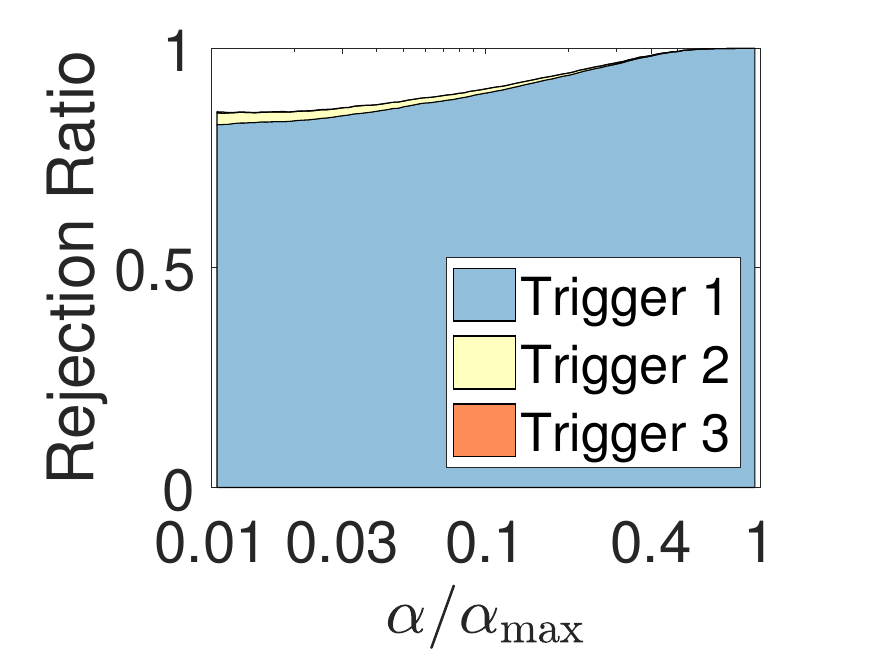}}
		\subfigure[ $\beta/\beta_{\rm{max}}$=0.5]{\includegraphics[scale=0.25]{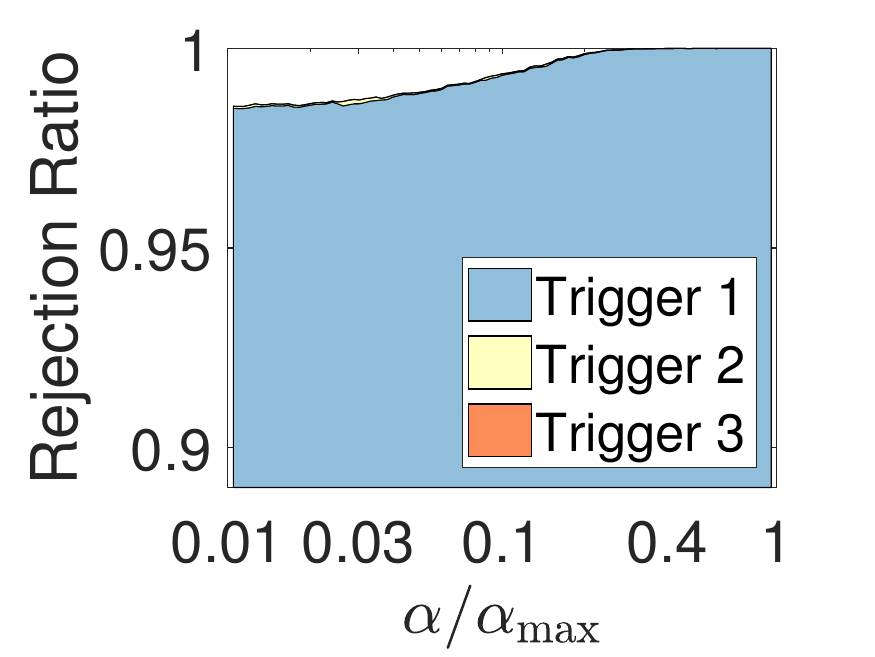}}
		\subfigure[ $\beta/\beta_{\rm{max}}$=0.9]{\includegraphics[scale=0.25]{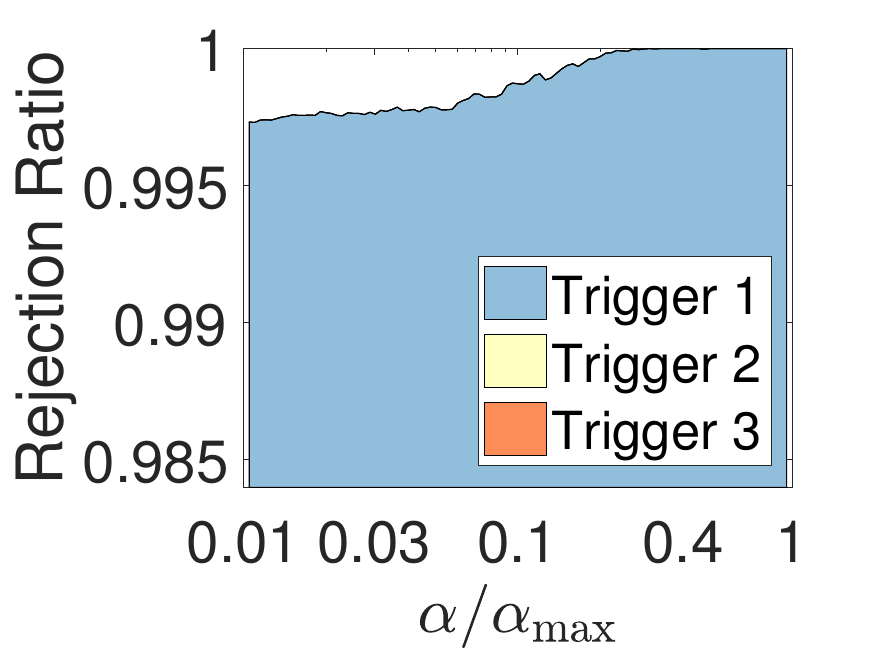}}
		\caption{Rejection ratios of SIFS on real-sim when it performs \textbf{IFS first} (first row: Feature Screening, second row: Sample Screening). }
		\label{fig:rejection-ratio-real-sim-feature-first}
	\end{center}
	\vspace*{-15pt}
\end{figure*}
\textbf{Verification of the Synergy Effect} In \figref{fig:rejection-ratio-real-sim-sample-first}, SIFS performs ISS (sample screening) first, while in \figref{fig:rejection-ratio-real-sim-feature-first}, it performs IFS (feature screening) first. All the rejection ratios of the 1st triggering of IFS in \figref{fig:rejection-ratio-real-sim-sample-first}(a)-\ref{fig:rejection-ratio-real-sim-sample-first}(d) where SIFS performs ISS first are much higher than (at least equal to) those in  \figref{fig:rejection-ratio-real-sim-feature-first}(a)-\ref{fig:rejection-ratio-real-sim-feature-first}(d) where SIFS performs IFS first. In turn, all the rejection ratios of the 1st triggering of ISS in  \figref{fig:rejection-ratio-real-sim-feature-first}(e)-\ref{fig:rejection-ratio-real-sim-feature-first}(h) where SIFS performs IFS first are also much  higher than those in \figref{fig:rejection-ratio-real-sim-sample-first}(e)-\ref{fig:rejection-ratio-real-sim-sample-first}(h) where SIFS performs ISS first. This demonstrates that the screening result of ISS can reinforce the capability of IFS and vice versa, which is the so called synergistic effect. At last, in \figref{fig:rejection-ratio-real-sim-feature-first} and \figref{fig:rejection-ratio-real-sim-sample-first}, we can see that the overall rejection ratios at the end of SIFS are exactly the same. Hence, no matter which rule (ISS or IFS) we perform first in SIFS, SIFS has the same screening performances in the end, which verifies the conclusion we present in Theorem \ref{thm:order-of-screening}.

\textbf{Performance in Solving Single Problems} In the experiments above, we solve problem (\ref{eqn:primal}) at a grid of turning parameter values. This setting is meaningful and it arises naturally in various cases, such as cross validation \citep{kohavi1995study} and feature selection \citep{meinshausen2010stability,yang2015detecting}. Therefore, the results above demonstrate that our method would be helpful in such real applications. We notice that sometimes one may be interested in the model at a specific parameter value pair. To this end, we now evaluate the performance of our method in solving  a single problem. Due to the space limitation, we solve sparse SVM at 16 specific parameter values pairs on the real-sim data set for examples, To be precise, $\beta \in \{0.05, 0.10, 0.5, 0.9\}$ and each $\beta$ has 4 values of $\alpha$ satisfying $\alpha/\alpha_{\max}(\beta) \in \{0.05, 0.1, 0.5, 0.9\}$. For each $(\alpha_i, \beta_i)$, we construct a parameter value path $\{(\alpha_{j,i}, \beta_i), j=0,...,M\}$ with $\alpha_{j,i}$ equally spaced at the logarithmic scale of $\alpha_{j,i}/\alpha_{max}(\beta_i)$ between $1$ and $\alpha_i /\alpha_{max}(\beta_i)$, which implies that $\alpha_{max}(\beta_i) = \alpha_{0,i}> \alpha_{1,i}>...>\alpha_{M,i}=\alpha_i$. Then we use AProx-SDCA integrated with SIFS to solve problem (\ref{eqn:primal}) at the parameter value pairs on the path one by one and report the total time cost. It can be expected that for $\alpha_i$ far from $\alpha_{max}(\beta_i)$, we need more points on the parameter value path in order to obtain higher rejection ratios and more significant speedups of SIFS and otherwise we need fewer ones. In this experiment, we fix $M=50$ at each $\alpha_i$ for convenience. For comparison, we solve the problem at $(\alpha_i, \beta_i)$ using AProx-SDCA with random initializations, which would be faster than solving all the problems on the path. Table \ref{table:speedup-single-problem} reports the speedups achieved by SIFS at these parameter value pairs. It shows that SIFS can still obtain significant speedups. In addition, compared with the results in Tables \ref{table:run-time-sync} and \ref{Table:speedup-real}, we can see that our method is much better at solving a problem sequence than solving a single problem.

\begin{table*}[htb!]
	\centering
	\begin{footnotesize}
		\begin{tabular}{|c|c|c|c|c|}
			\hline
			\backslashbox{$\beta/\beta_{\max}$\kern-1em}{\kern-1em$\alpha/\alpha_{\max}(\beta)$}&0.05&0.10&0.50&0.90\\ \hline
			0.05&6.31&7.14&21.04&146.66\\ \hline
			0.10&6.14&7.18&21.83&156.63\\ \hline
			0.50&5.75&7.18&20.99&154.60\\ \hline
			0.90&6.39&7.66&23.16&173.56\\ \hline
		\end{tabular}
	\end{footnotesize}
	\caption{The speedups achieved by SIFS in solving single problems}\label{table:speedup-single-problem}
		\vspace*{-5pt}
\end{table*}

\subsection{Experiments with Multi-class Sparse SVMs}
We evaluate SIFS for multi-class Sparse SVMs on 3 synthetic data sets (syn-multi1, syn-multi2, and syn-multi3) and 2 real data sets (news20 and rcv1-multiclass). Their statistics are given in Table \ref{data_description-multi}. 

The synthetic data sets are generated in a similar way as we did in Section \ref{sec:experiment-simu-binary}. Each of them has $K=5$ classes and each class has $n/K$ samples. The data points of them can be written as $\x=[\x_1; \x_2]$, where $\x_1=[\x_1^1;...;\x_1^K] \in \R^{0.02p}$ with $\x_1^k \in \R^{0.02p/K}$and $\x_2 \in \R^{0.98p}$. If $\x$ belongs to the $k$-th class, then $\x_1^k$ is sampled from a Gaussian distribution $\calG = N(\u, 0.75\I)$ with $\u=1.5\mathbf{1}$, $\I\in \R^{(0.02p/K) \times (0.02p/K)}$ and other components in $\x_1$ are from $N(0,1)$. Each entry of $\x_2$ has chance $\eta=0.2$ to be sampled from distribution $N(0,1)$ and chance $1-\eta$ to be 0. 

\begin{table*}[htb!]
	\centering
	\begin{footnotesize}
		\begin{tabular}{|l|l|l|c|}
			\hline
			Dataset & Feature size: $p$ & Sample size: $n$ & Classes \\ \hline
			syn-multi1 &1,000 &10,000& 5\\ \hline
			syn-multi2&10,000& 10,000& 5 \\ \hline
			syn-multi3&10,000 &1,000& 5 \\ \hline
			news20 & 3,993 & 62,060 & 20 \\ \hline
			rcv1-multiclass & 47,236 & 15,564 & 53 \\ \hline
		\end{tabular}
	\end{footnotesize}
	\caption{Statistics of the multi-class data sets.}\label{data_description-multi}
\end{table*}

\begin{figure}[htb!]
	\centering
	\subfigure[syn-multi1.]{\includegraphics[width=.8\textwidth]{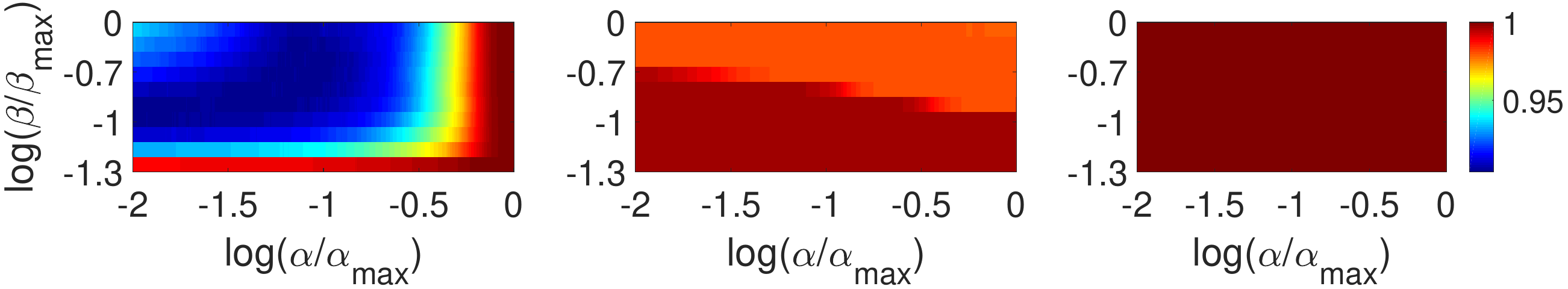}}
	\subfigure[syn-multi2.]{\includegraphics[width=.8\textwidth]{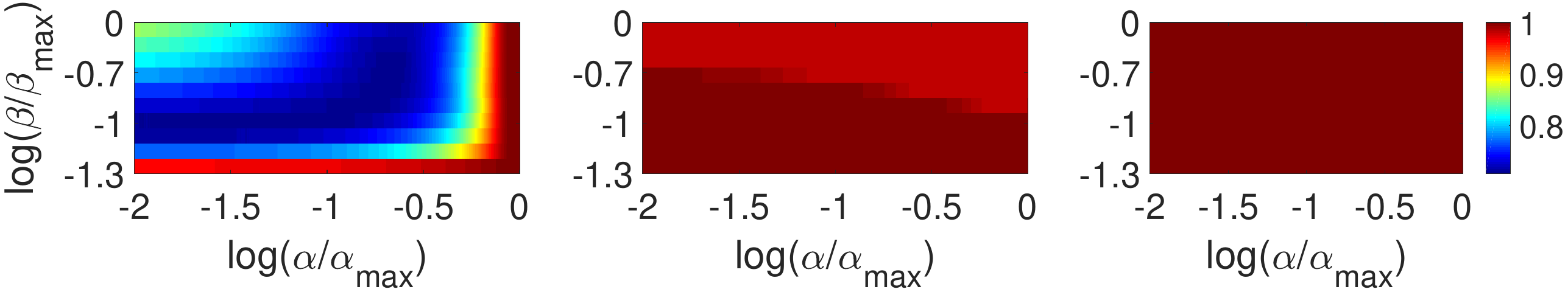}}
	\subfigure[syn-multi3.]{\includegraphics[width=.8\textwidth]{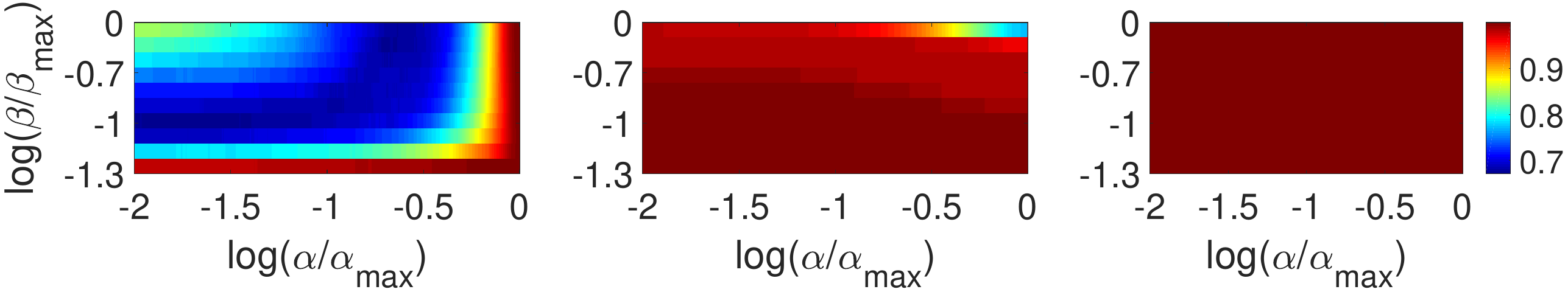}}
	\caption{Scaling ratios of ISS, IFS, and SIFS (from left to right).}\label{fig:scaling-ratio-sync-mc}
\end{figure}

\figref{fig:scaling-ratio-sync-mc} shows the scaling ratios of ISS, IFS, and SIFS on the synthetic data sets at $1,000$ pairs of parameter values. Similar to the results in sparse SVMs, SIFS totally surpasses IFS and ISS, with scaling ratios larger than $98\%$ at any parameter value pair against $70-90\%$. Thus, we can expect significant speedups by integrating SIFS with the solver. 

\begin{figure*}[htb!]
	\begin{center}
		\subfigure[ $\beta/\beta_{\rm{max}}$=0.05]{\includegraphics[scale=0.25]{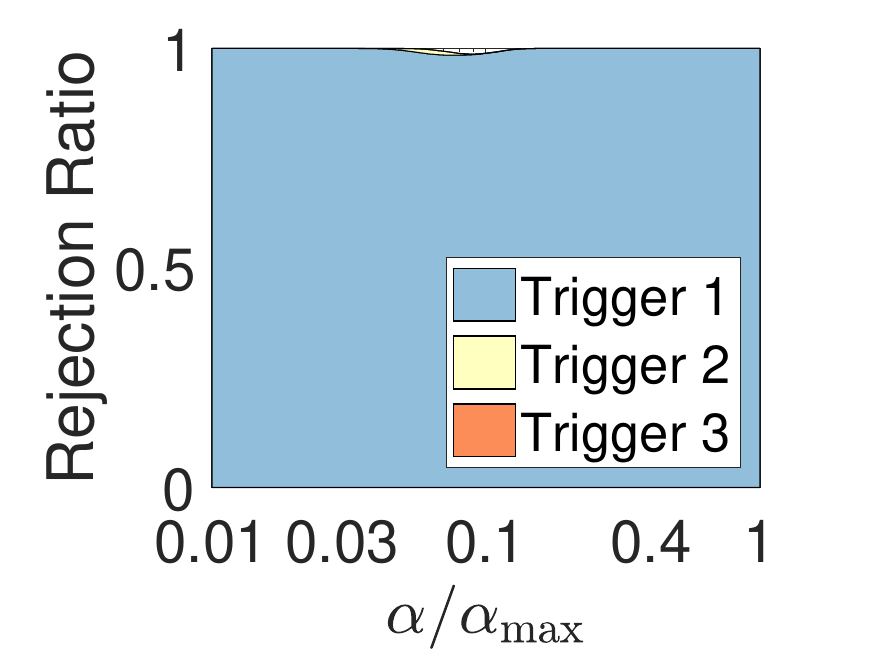}}
		\subfigure[ $\beta/\beta_{\rm{max}}$=0.1]{\includegraphics[scale=0.25]{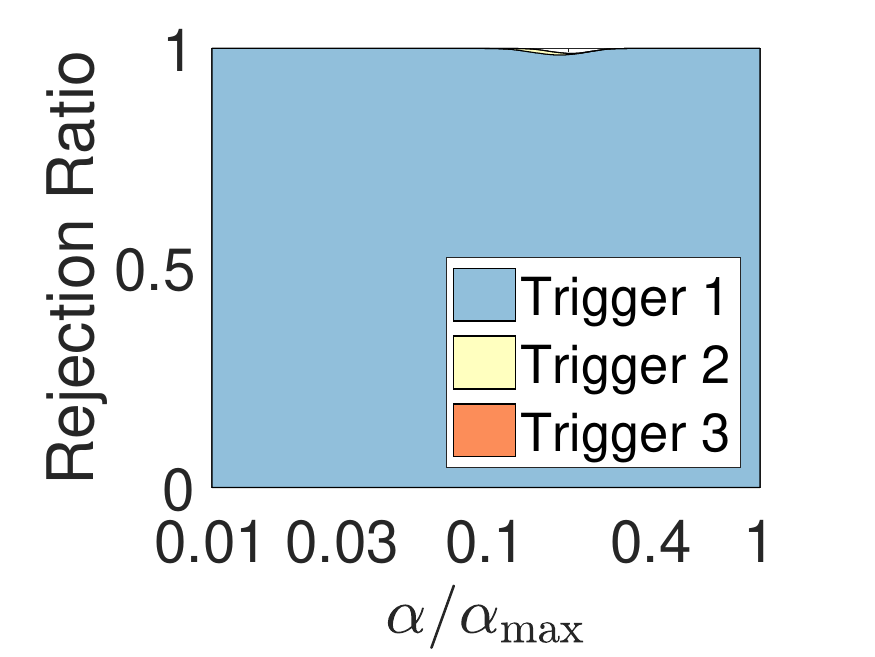}}
		\subfigure[ $\beta/\beta_{\rm{max}}$=0.5]{\includegraphics[scale=0.25]{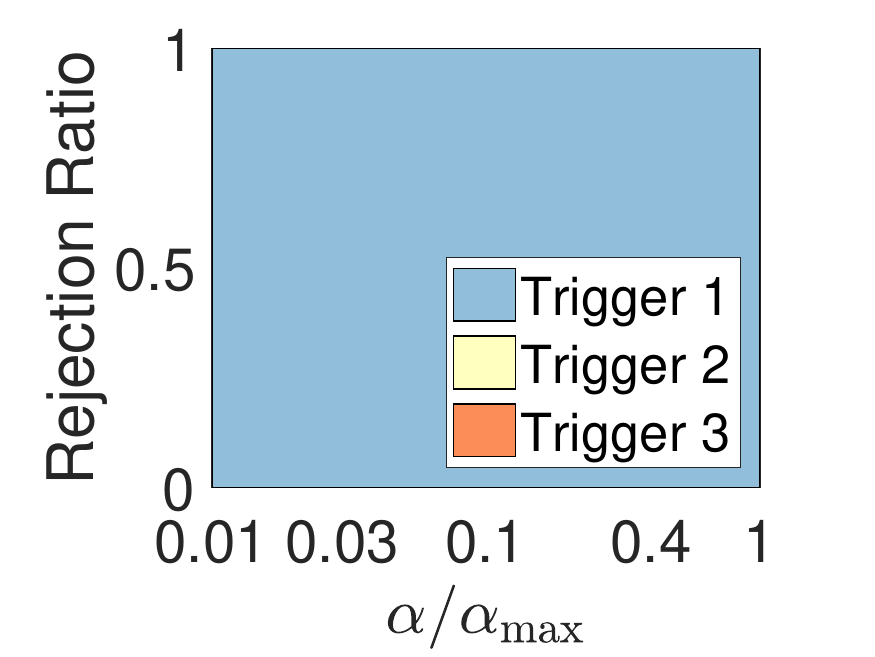}}
		\subfigure[ $\beta/\beta_{\rm{max}}$=0.9]{\includegraphics[scale=0.25]{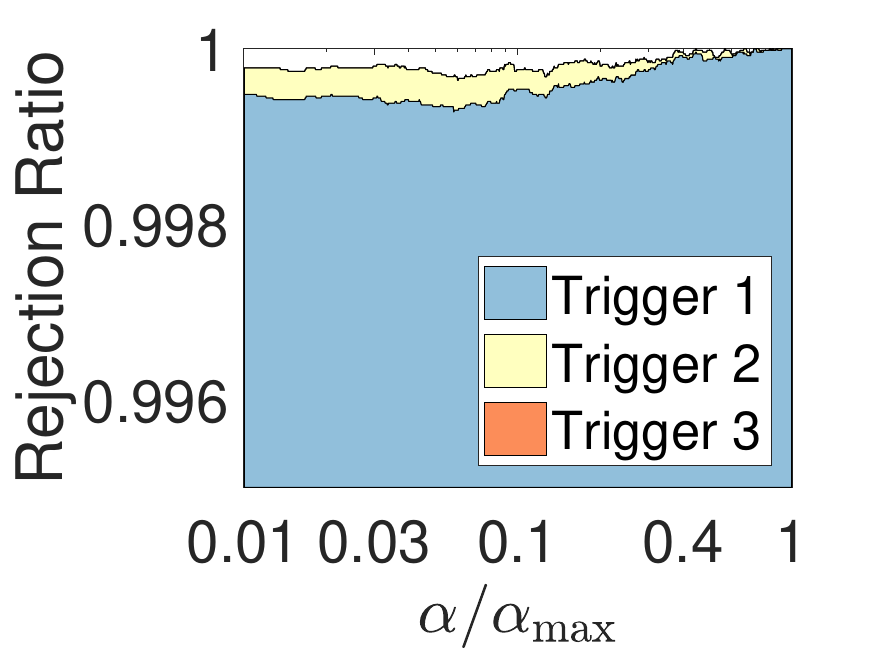}}
		\subfigure[ $\beta/\beta_{\rm{max}}$=0.05]{\includegraphics[scale=0.25]{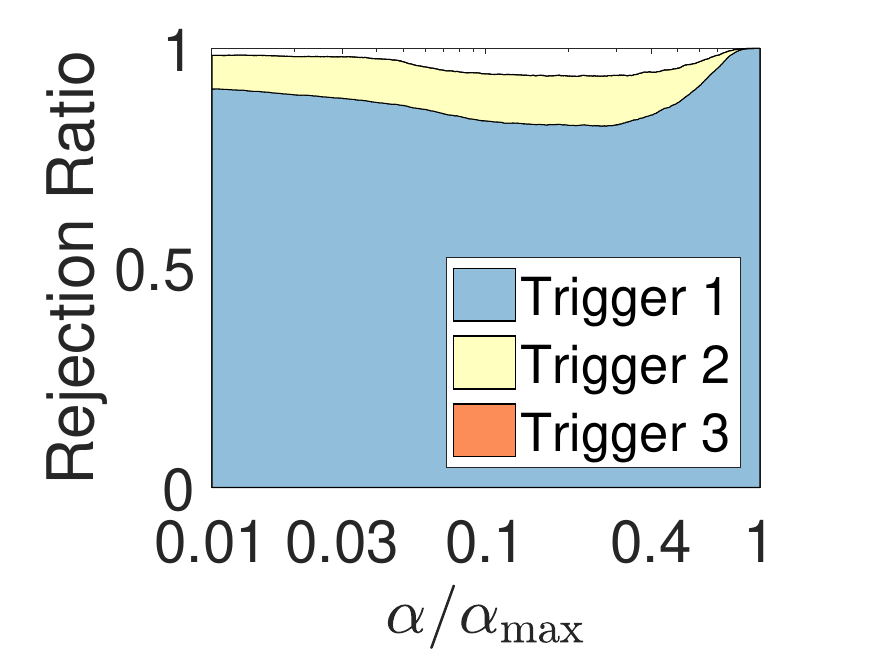}}
		\subfigure[ $\beta/\beta_{\rm{max}}$=0.1]{\includegraphics[scale=0.25]{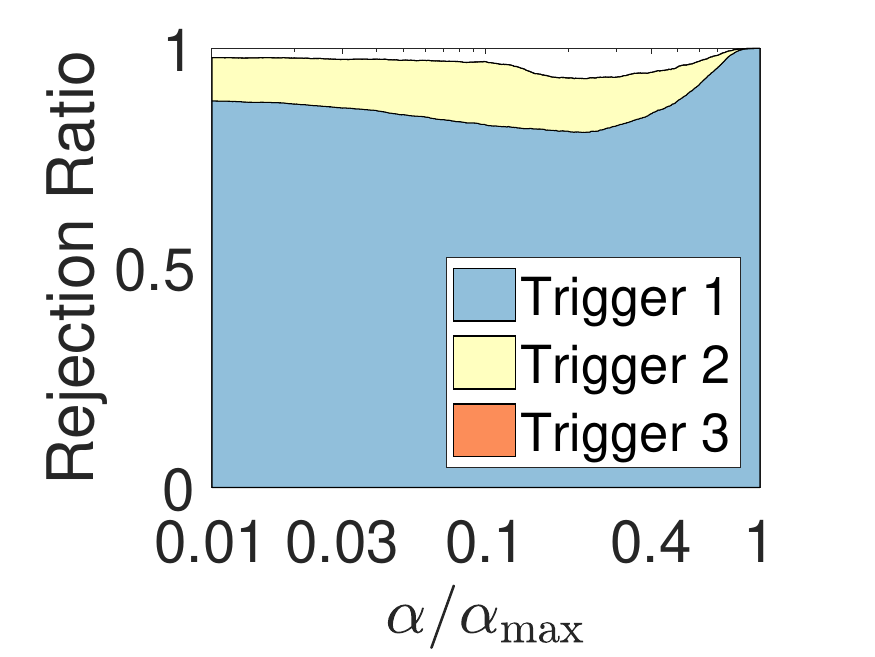}}
		\subfigure[ $\beta/\beta_{\rm{max}}$=0.5]{\includegraphics[scale=0.25]{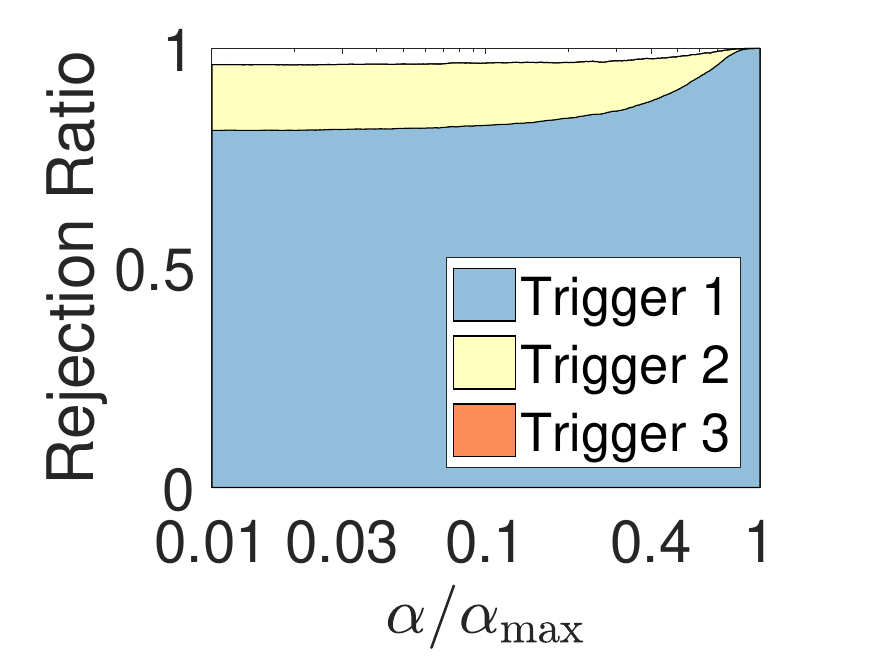}}
		\subfigure[ $\beta/\beta_{\rm{max}}$=0.9]{\includegraphics[scale=0.25]{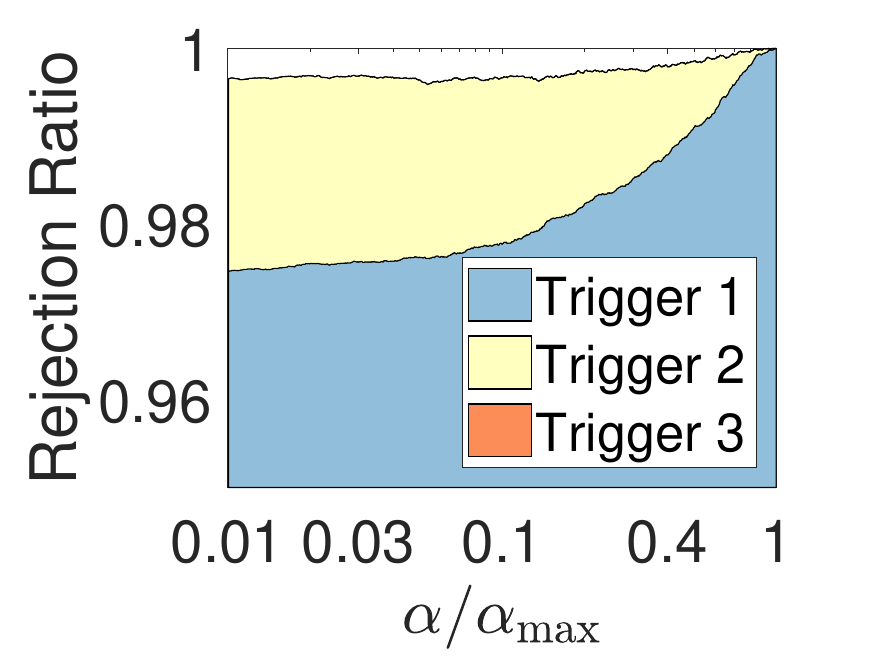}}
		\caption{Rejection ratios of SIFS on syn-multi2 (first row: Feature Screening, second row: Sample Screening). }
		\label{fig:rejection-ratio-syn-multi2}
	\end{center}
	\vspace*{-20pt}
\end{figure*}

\begin{figure*}[htb!]
	\begin{center}
		\subfigure[ $\beta/\beta_{\rm{max}}$=0.05]{\includegraphics[scale=0.25]{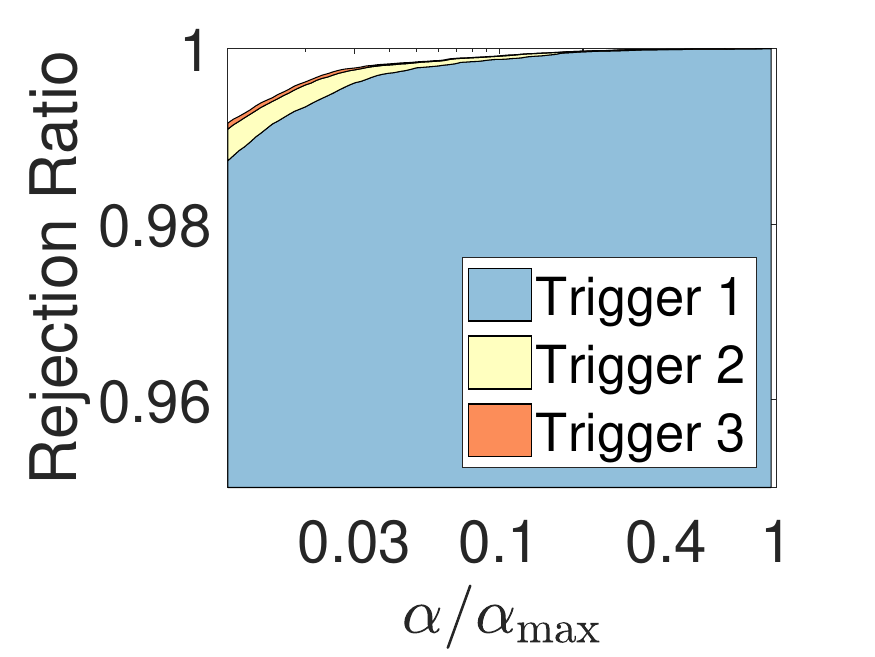}}
		\subfigure[ $\beta/\beta_{\rm{max}}$=0.1]{\includegraphics[scale=0.25]{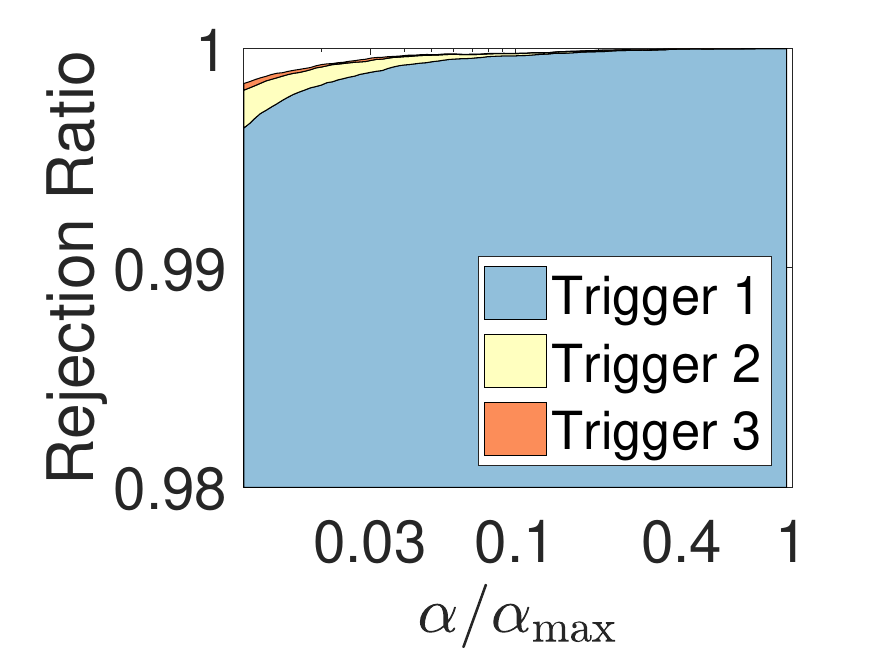}}
		\subfigure[ $\beta/\beta_{\rm{max}}$=0.5]{\includegraphics[scale=0.25]{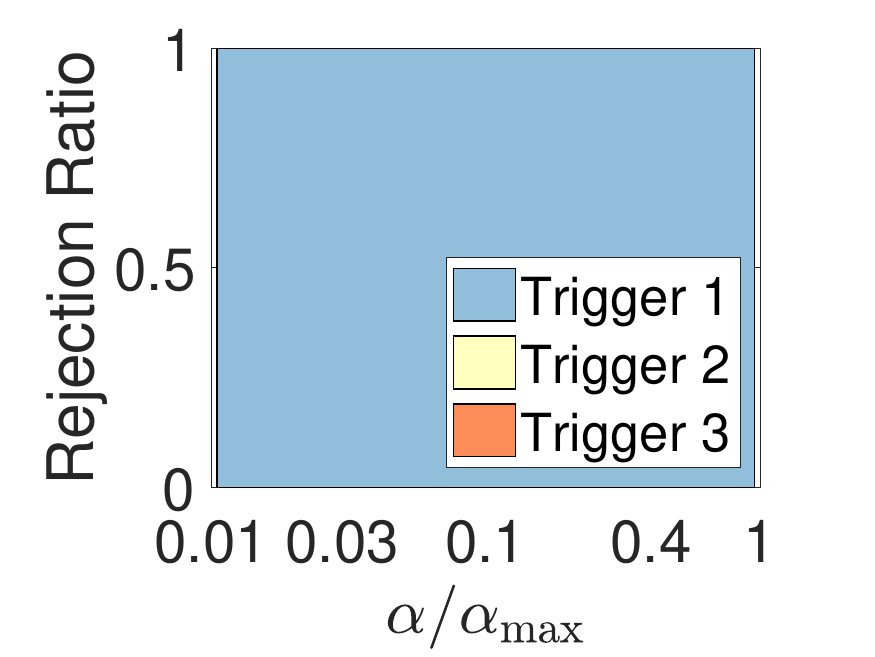}}
		\subfigure[ $\beta/\beta_{\rm{max}}$=0.9]{\includegraphics[scale=0.25]{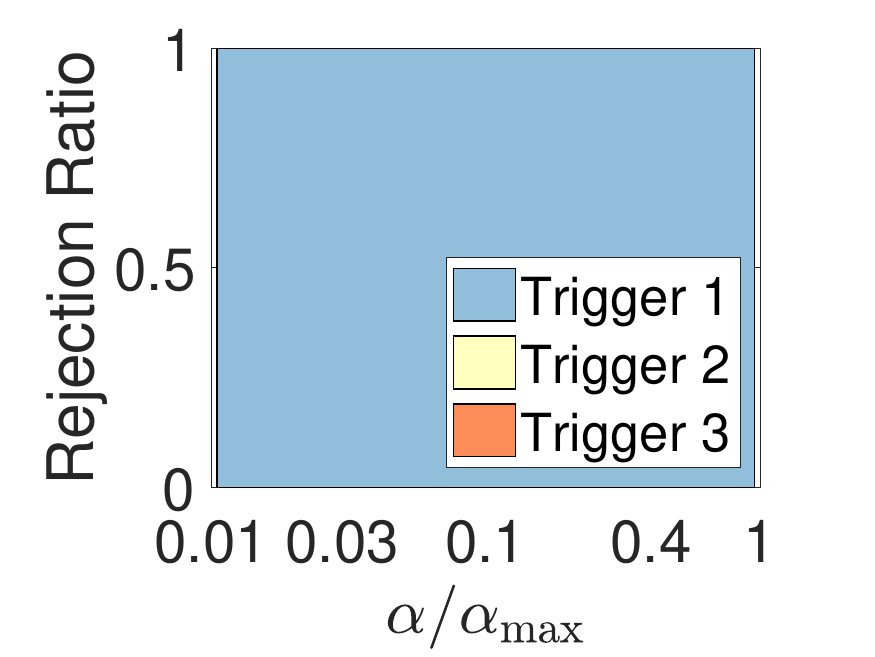}}
		\subfigure[ $\beta/\beta_{\rm{max}}$=0.05]{\includegraphics[scale=0.25]{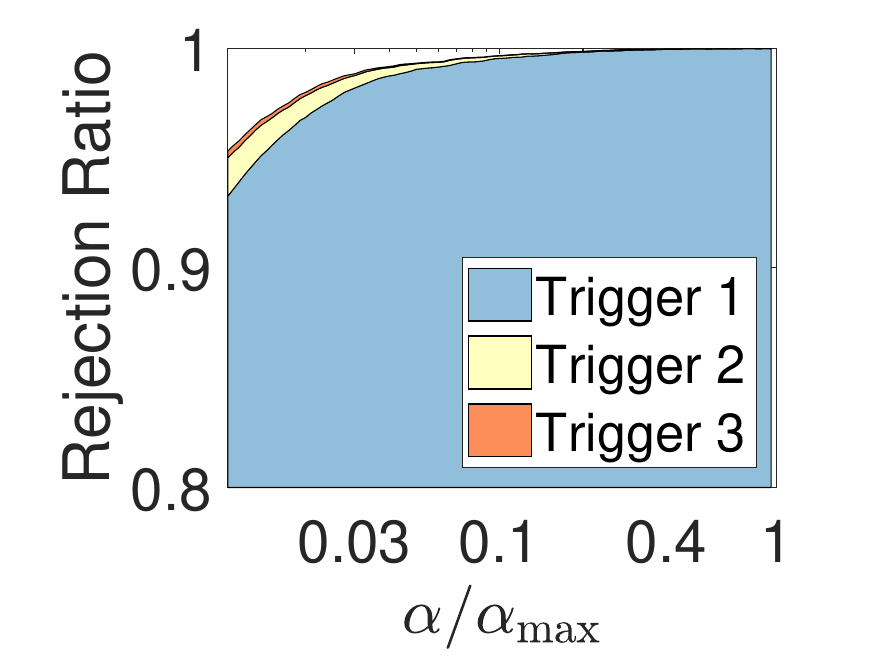}}
		\subfigure[ $\beta/\beta_{\rm{max}}$=0.1]{\includegraphics[scale=0.25]{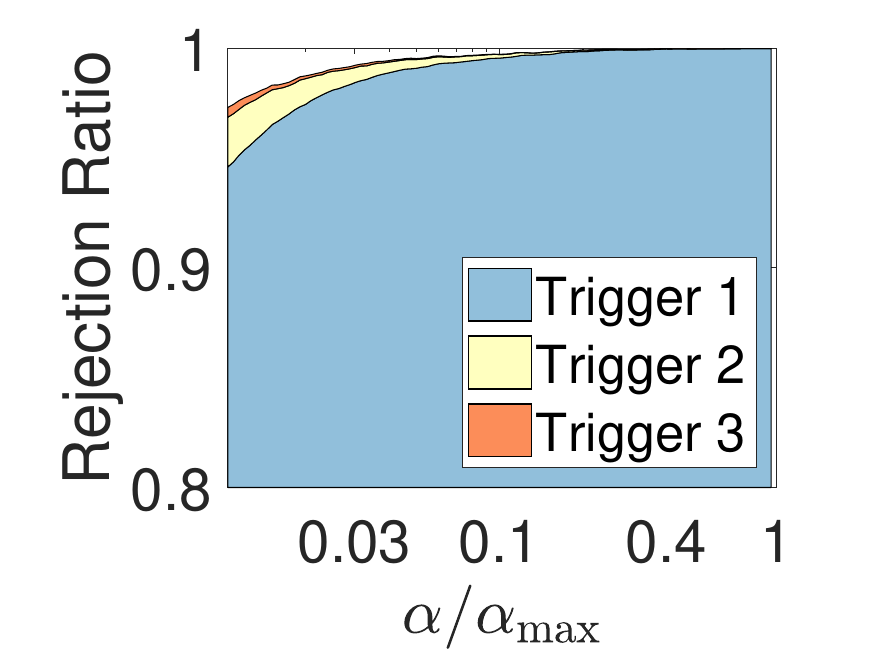}}
		\subfigure[ $\beta/\beta_{\rm{max}}$=0.5]{\includegraphics[scale=0.25]{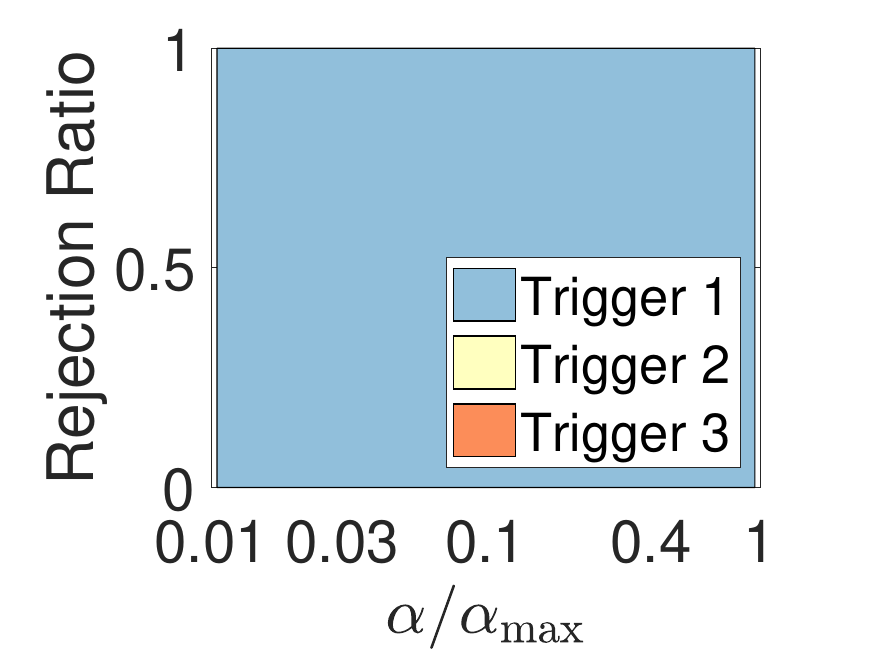}}
		\subfigure[ $\beta/\beta_{\rm{max}}$=0.9]{\includegraphics[scale=0.25]{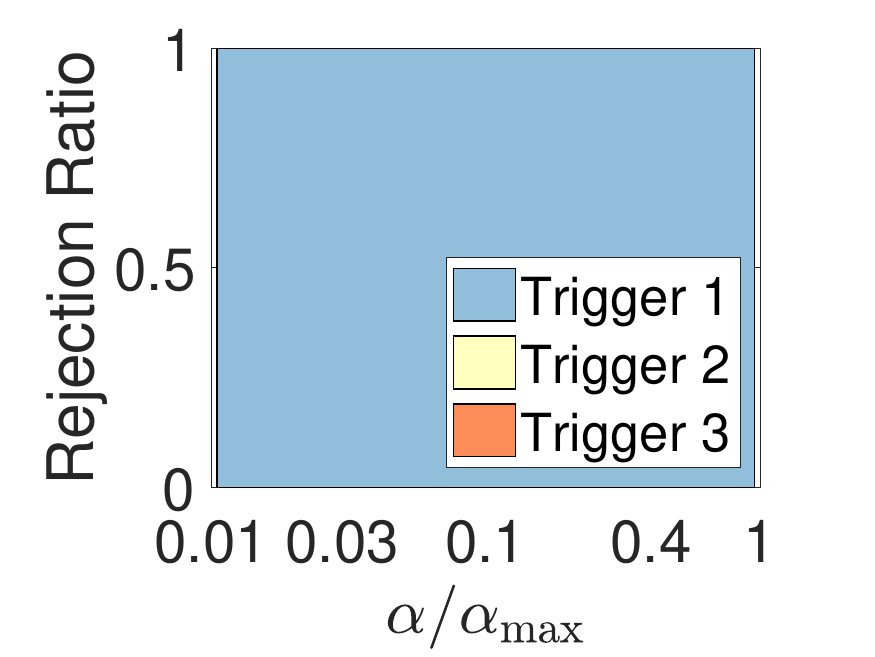}}
		\caption{Rejection ratios of SIFS on news20 (first row: Feature Screening, second row: Sample Screening). }
		\label{fig:rejection-ratio-news20}
	\end{center}
\vspace*{-30pt}
\end{figure*}

\figref{fig:rejection-ratio-syn-multi2} and \ref{fig:rejection-ratio-news20} present the rejection ratios of SIFS on the news20 data set (the results of other data sets can be found in the appendix). It indicates that, by alternatively applying IFS and ISS, we can finally identify most of the inactive samples ($>95\%$) and features ($>99\%$). The synergistic effect between IFS and ISS can also be found in these figures. 

As a result of excellent performance of SIFS in identifying inactive features and samples, we observe significant speedups gained by SIFS in Table \ref{table:run-time-MSSVM}, which are up to 200 times on news20 and 300 times on syn-multi3. Specifically, on news20, the solver without any screening method takes about 97 hours to solve the problem at $1,000$ pairs of parameter values; however, integrated with SIFS, the solver only needs less than 0.5 hour for the same task. Moreover, Table \ref{table:run-time-MSSVM} also indicates that the speedup gained by SIFS is much more significant than that gained by  IFS and ISS. This demonstrates again the great benefit of alternatively applying IFS and ISS  in SIFS. At last, we notice that the speedup gained by SIFS and IFS on syn-multi3 is far greater than those gained on syn-multi1 and syn-multi2. The reason is that the convergence rate of the solver in our problem may depend on the condition number of the data matrix. The dimension of $\w$ in syn-multi3 is much larger than  its sample size ($50,000$ over $1,000$), which leads to a large condition number and a bad convergence rate. However, SIFS and IFS can remove the inactive features from the model, which can  greatly improve the convergence rate of the solver by decreasing the condition number. This point can also be supported by the fact in the experiment on syn-multi3 that the solver with SIFS or IFS needs much fewer iterations to converge to the optima than that without SIFS or IFS.  

\begin{table*}[htb!]
	\centering
	{\footnotesize
		\begin{tabular}{|p{2.2cm}<{\centering}|p{0.8cm}<{\centering}|p{0.3cm}<{\centering}|p{0.7cm}<{\centering}|p{1.0cm}<{\centering}|p{0.5cm}<{\centering}|p{0.7cm}<{\centering}|p{1.0cm}<{\centering}|p{0.5cm}<{\centering}|p{0.7cm}<{\centering}|p{1.2cm}<{\centering}|}
			\hline
			\multirow{2}{*}{Data} & \multirow{2}{*}{Solver} & \multicolumn{3}{c|}{ISS+Solver} & \multicolumn{3}{c|}{IFS+Solver} & \multicolumn{3}{c|}{SIFS+Solver} \\ \cline{3-11} 
			&  & ISS & Solver & Speedup & IFS & Solver & Speedup & SIFS & Solver & \textbf{Speedup} \\ \hline
			syn-multi1 &9009&72&1233&6.9&90&388&18.9&178&83&\textbf{34.5}\\ \hline
			syn-multi2 &105484&412&27661&3.8&920&2606&29.9&1936&893&\textbf{37.3} \\ \hline			
			syn-multi3 &78373&44&18341&4.3&84&337&186.2&203&59&\textbf{299.9} \\ \hline
			news20 &350277&283&13950&24.6&1000&65175&5.3&1485&290&\textbf{197.3}\\ \hline 			
			rcv1-multiclass &396205&793&44639&8.7&1620&33191&11.4&2850&1277&\textbf{96.0} \\ \hline 
		\end{tabular}
	}
	\caption{Running time (in seconds) for solving problem (\ref{eqn:primal-MSSVM}) at $1,000$ pairs of parameter values on synthetic and real data sets.}
\label{table:run-time-MSSVM}
\end{table*}

\section{Conclusions}
In this paper, we proposed a novel data reduction method SIFS to simultaneously identify inactive features and samples for sparse SVMs. Our major contribution is a novel framework for an accurate estimation of the primal and dual optima based on strong convexity. To the best of our knowledge, the proposed SIFS is the first static screening method that is able to simultaneously identify inactive features and samples for sparse SVMs. An appealing feature of SIFS is that all detected features and samples are guaranteed to be irrelevant to the outputs. Thus, the model learned on the reduced data is exactly identical to that learned on the full data. To show the flexibility of the proposed SIFS, we extended it to multi-class sparse SVMs. The experimental results demonstrate that, for both sparse SVMs and multi-class sparse SVMs, SIFS can dramatically reduce the problem size and the resulting speedup can be orders of magnitude. We plan to generalize SIFS to more complicated models, e.g., SVMs with a structured sparsity-inducing penalty. 

\acks{This work was supported by the National Basic Research Program of China (973 Program) under Grant 2013CB336500, National Natural Science Foundation of China under Grant 61233011, and National Youth Top-notch Talent Support Program.}


\newpage
\appendix
\section{Detailed Proofs and More Experimental Results}
In this appendix, we first present the detailed proofs of all the theorems in the main paper and then report the rest experimental results which are omitted in the experimental section.

\subsection{Proof for Lemma \ref{lemma:dual-scaled}}
\proof of Lemma \ref{lemma:dual-scaled}:

1) It is the immediate conclusion of the analysis above. 

2) After feature screening, the primal problem (\ref{eqn:primal}) is scaled into:
\begin{align}
\min_{\wt\in \R^{|\hat{\calF}^c|}}\frac{\alpha}{2}||\wt||^2 + \beta ||\wt||_1 + \frac{1}{n}\sum\limits_{i=1}^{n}\ell(1-\langle \mbox{[}\xb_i\mbox{]}_{\hcalF^c},\wt \rangle ). \tag{scaled-$P^*$-1}\label{eqn:primal-scaled-1}
\end{align}
Thus, we can easily derive out the dual problem of (\ref{eqn:primal-scaled-1}):
\begin{align}
\min_{\tilde{\thetab} \in \mbox{[}0,\alpha\mbox{]}^n}  \tilde{D}(\tilde{\thetab};\alpha, \beta) =\frac{1}{2\alpha}||\calS_{\beta}(\frac{1}{n}{}_{\hcalF^c}\mbox{[}\Xb\mbox{]} \tilde{\thetab})||^2 +\frac{\gamma}{2n}||\tilde{\thetab}||^2 - \frac{1}{n}\langle \textbf{1},\tilde{\thetab} \rangle  \tag{scaled-$D^*$-1},\label{eqn:dual-scaled-1}
\end{align}
and the KKT conditions:
\begin{align}
&\wt^*(\alpha,\beta) = \frac{1}{\alpha}\calS_{\beta}(\frac{1}{n}{}_{\hcalF^c}\mbox{[}\Xb\mbox{]} \tilde{\theta}^*(\alpha, \beta)), \tag{scaled-KKT-1} \label{eqn:KKT1-scaled}\\
&\mbox{[}\tilde{\theta}^*(\alpha,\beta)\mbox{]}_i = \left\{
\begin{array}{ccc}
0,&\mbox{ if }1 - \langle \mbox{[}\xb_i\mbox{]}_{\hcalF^c}, \wt^*(\alpha,\beta)\rangle<0,\\
\frac{1}{\gamma } (1 - \langle \mbox{[}\xb_i\mbox{]}_{\hcalF^c}, \wt^*(\alpha,\beta)),&\mbox{ if } 0 \leq 1 - \langle \mbox{[}\xb_i\mbox{]}_{\hcalF^c}, \wt^*(\alpha,\beta) \leq \gamma,\\
1, &\mbox{ if } 1 - \langle \mbox{[}\xb_i\mbox{]}_{\hcalF^c}, \wt^*(\alpha,\beta) > \gamma,
\end{array}
\right. \tag{scaled-KKT-2} \label{eqn:KKT2-scaled}
\end{align}
Then, it is obvious that $\wt^*(\alpha, \beta) = \mbox{[}\w^*(\alpha, \beta)\mbox{]}_{\hat{\calF}^c}$, since essentially, problem (\ref{eqn:primal-scaled-1}) can be derived by substituting 0 to the weights for the eliminated features in problem (\ref{eqn:primal}) and optimizing over the rest weights. \\
Since the solutions $\w^*(\alpha, \beta)$ and $\theta^*(\alpha, \beta)$ satisfy the conditions (\ref{eqn:KKT1}) and (\ref{eqn:KKT2}) and $\langle [\xb_i\mbox{]}_{\hat{\calF}^c},\wt^*(\alpha, \beta)\rangle=\langle \xb_i, \w^*(\alpha, \beta)\rangle$ for all $i$ , we know that $\wt^*(\alpha, \beta)$ and $\theta^*(\alpha, \beta)$ satisfy the conditions (\ref{eqn:KKT1-scaled}) and (\ref{eqn:KKT2-scaled}). Thus they are the solutions of problems (\ref{eqn:primal-scaled-1}) and (\ref{eqn:dual-scaled-1}). Then, due to the uniqueness of the solution of problem (\ref{eqn:dual-scaled-1}), we have
\begin{align}
\thetab^*(\alpha, \beta) =\tilde{\thetab}^*(\alpha, \beta). 
\end{align} 

From 1) we have $\mbox{[}\tilde{\thetab}^*(\alpha, \beta)\mbox{]}_{\hat{\calR}^c} = 0$ and $\mbox{[}\tilde{\thetab}^*(\alpha, \beta)\mbox{]}_{\hat{\calL}^c} = 1$. Therefore, from the dual problem (\ref{eqn:dual-scaled}), we can see that $\mbox{[}\tilde{\thetab}^*(C,\alpha)\mbox{]}_{\hat{\calD}^c}$ can be recovered from the following problem:
\begin{align}
\min_{\hat{\thetab} \in \mbox{[}0,1\mbox{]}^{|\hat{\calD}^c|}} \frac{1}{2\alpha}||\calS_{\beta}(\frac{1}{n}\hat{\G}_{1} \hat{\thetab}+\frac{1}{n} \hat{\G}_{2} \textbf{1})||^2+ \frac{\gamma}{2n}||\hat{\theta}||^2-\frac{1}{n}\langle\textbf{1}, \hat{\thetab}\rangle.\nonumber
\end{align}
Since $\mbox{[}\tilde{\thetab}^*(\alpha, \beta)\mbox{]}_{\hat{\calD}^c}=\mbox{[}\thetab^*(\alpha, \beta)\mbox{]}_{\hat{\calD}^c}$, the proof is therefore completed. 
\endproof

\subsection{Proof for Lemma \ref{beta-max-and-alpha-max}}
\proof of Lemma \ref{beta-max-and-alpha-max}:

(i) We prove this lemma by verifying that the solutions $\w^*(\alpha, \beta) = \textbf{0}$ and $\thetab^*(\alpha, \beta) = \textbf{1}$ satisfy the 
conditions (\ref{eqn:KKT1}) and (\ref{eqn:KKT2}).

Firstly, since $\beta \geq \beta_{\max} = ||\frac{1}{n}\Xb\textbf{1}||_{\infty}$, we have $\calS_{\beta}(\frac{1}{n}\Xb \textbf{1}) = 0$. Thus $\w^*(\alpha,\beta) = \textbf{0}$ and $ \thetab^*(\alpha,\beta) = \textbf{1}$ satisfy the condition (\ref{eqn:KKT1}).

Then, for all $i\in \mbox{[}n\mbox{]}$, we have 
\begin{align}
1-\langle \xb_i, \w^*(\alpha, \beta)\rangle = 1-0> \gamma. \nonumber
\end{align}
Thus, $\w^*(\alpha,\beta) = \textbf{0}$ and $\thetab^*(\alpha,\beta) = \textbf{1}$ satisfy the condition (\ref{eqn:KKT2}). Hence, they are the solutions of the primal problem (\ref{eqn:primal}) and the dual problem (\ref{eqn:dual}), respectively.

(ii) Similar to the proof of (i), we prove this  by verifying that the solutions $\w^*(\alpha, \beta) = \frac{1}{\alpha}\calS_{\beta}(\frac{1}{n}\Xb \thetab^*(\alpha, \beta))$ and $\thetab^*(\alpha, \beta) = \textbf{1}$ satisfy the conditions (\ref{eqn:KKT1}) and (\ref{eqn:KKT2}).
\begin{enumerate}
	\item \textbf{Case 1:} $\alpha_{\max}(\beta) \leq 0$. Then for all $\alpha >0$, we have 
	\begin{align}
	&\min_{i\in \mbox{[}n\mbox{]}}\{1-\langle \xb_i, \w^*(\alpha, \beta)\rangle \}\nonumber \\
	=&\min_{i\in \mbox{[}n\mbox{]}}\{1-\frac{1}{\alpha}\langle \xb_i, \calS_{\beta}(\frac{1}{n}\Xb \thetab^*(\alpha, \beta)) \rangle \} =\min_{i\in \mbox{[}n\mbox{]}}\{1-\frac{1}{\alpha}\langle \xb_i, \calS_{\beta}(\frac{1}{n} \Xb \textbf{1})\rangle \}\nonumber\\
	= & 1 - \frac{1}{\alpha} \max_{i\in \mbox{[}n\mbox{]}}\langle \xb_i, \calS_{\beta}(\frac{1}{n} \Xb \textbf{1})\rangle = 1-(1-\gamma)\frac{1}{\alpha}\alpha_{\max}(\beta) \nonumber \\
	\geq&  1 > \gamma.\nonumber
	\end{align}
	Then, $\calL = \mbox{[}n\mbox{]}$, and $\w^*(\alpha, \beta) = \frac{1}{\alpha}\calS_{\beta}(\frac{1}{n}\Xb \thetab^*(\alpha, \beta))$ and $\thetab^*(\alpha, \beta) = \textbf{1}$ satisfy the conditions (\ref{eqn:KKT1}) and (\ref{eqn:KKT2}). Hence, they are the optimal solution of the primal and dual problems (\ref{eqn:primal})  and (\ref{eqn:dual}).
	\item \textbf{Case 2:} $\alpha_{\max}(\beta) >0$. Then for any $\alpha \geq  \alpha_{\max}(\beta)$, we have
	\begin{align}
	&\min_{i\in \mbox{[}n\mbox{]}}\{1-\langle \xb_i, \w^*(\alpha, \beta)\rangle \}\nonumber \\
	=&\min_{i\in \mbox{[}n\mbox{]}}\{1-\frac{1}{\alpha}\langle \xb_i, \calS_{\beta}(\frac{1}{n}\Xb \thetab^*(\alpha, \beta)) \rangle \} = \min_{i\in \mbox{[}n\mbox{]}}\{1-\frac{1}{\alpha}\langle \xb_i, \calS_{\beta}(\frac{1}{n} \Xb \textbf{1})\rangle \}\nonumber\\
	= & 1 - \frac{1}{\alpha} \max_{i\in \mbox{[}n\mbox{]}}\langle \xb_i, \calS_{\beta}(\frac{1}{n} \Xb \textbf{1})\rangle = 1 - (1-\gamma)\frac{1}{\alpha}\alpha_{\max}(\beta)\geq  1- (1-\gamma) = \gamma.  \nonumber
	\end{align}
	Thus, $\calE\cup \calL = \mbox{[}n\mbox{]}$, and $\w^*(\alpha, \beta) = \frac{1}{\alpha}\calS_{\beta}(\frac{1}{n}\Xb \thetab^*(\alpha, \beta))$ and $\thetab^*(\alpha, \beta) =  \textbf{1}$ satisfy the conditions (\ref{eqn:KKT1}) and (\ref{eqn:KKT2}). Hence, they are the optimal solution of the primal and dual problems (\ref{eqn:primal})  and (\ref{eqn:dual}). 
\end{enumerate}
The proof is complete.
\endproof

\subsection{Proof for Lemma \ref{lemma:primal-estimation}}
\proof of Lemma \ref{lemma:primal-estimation}:

Due to the $\alpha$-strong convexity of the objective $P(\w;\alpha, \beta)$, we have 
\begin{align}
&P(\w^*(\alpha_0, \beta_0); \alpha, \beta_0)\geq P(\w^*(\alpha, \beta_0); \alpha, \beta_0)+\frac{\alpha}{2}||\w^*(\alpha_0, \beta_0)-\w^*(\alpha, \beta_0)||^2,\nonumber \\
&P(\w^*(\alpha, \beta_0); \alpha_0, \beta_0)\geq P(\w^*(\alpha_0, \beta_0); \alpha_0, \beta_0)+\frac{\alpha_0}{2}||\w^*(\alpha_0, \beta_0)-\w^*(\alpha, \beta_0)||^2,\nonumber
\end{align}
which are equivalent to 

\begin{align}
&\frac{\alpha}{2}||\w^*(\alpha_0, \beta_0)||^2 + \beta_0 ||\w^*(\alpha_0, \beta_0)||_1 + L(\w^*(\alpha_0, \beta_0))\nonumber \\
&\geq \frac{\alpha}{2}||\w^*(\alpha, \beta_0)||^2 + \beta_0 ||\w^*(\alpha, \beta_0)||_1 + L(\w^*(\alpha, \beta_0))+\frac{\alpha}{2}||\w^*(\alpha_0, \beta_0)-\w^*(\alpha, \beta_0)||^2, \nonumber \\
&\frac{\alpha_0}{2}||\w^*(\alpha, \beta_0)||^2 + \beta_0 ||\w^*(\alpha, \beta_0)||_1 + L(\w^*(\alpha, \beta_0))\nonumber \\
&\geq \frac{\alpha_0}{2}||\w^*(\alpha_0, \beta_0)||^2 + \beta_0 ||\w^*(\alpha_0, \beta_0)||_1 + L(\w^*(\alpha_0, \beta_0))+\frac{\alpha_0}{2}||\w^*(\alpha_0, \beta_0)-\w^*(\alpha, \beta_0)||^2.\nonumber 
\end{align}
Adding the above two inequalities together, we obtain 
\begin{align}
&\frac{\alpha-\alpha_0}{2}||\w^*(\alpha_0, \beta_0)||^2\geq \frac{\alpha-\alpha_0}{2}||\w^*(\alpha, \beta_0)||^2+\frac{\alpha_0+\alpha}{2}||\w^*(\alpha_0, \beta_0)-\w^*(\alpha, \beta_0)||^2\nonumber \\
&\Rightarrow ||\w^*(\alpha, \beta_0)- \frac{\alpha_0+ \alpha}{2\alpha}\w^*(\alpha_0, \beta_0)||^2 \leq \frac{(\alpha-\alpha_0)^2}{4\alpha^2}||\w^*(\alpha_0,\beta_0)||^2.\label{eqn:primal-estimation-temp1} 
\end{align}
Substituting the prior that $\mbox{[}\w^*(\alpha, \beta_0)\mbox{]}_{\hcalF} = 0$ into \eqref{eqn:primal-estimation-temp1}, we obtain 
\begin{align}
&||\mbox{[}\w^*(\alpha, \beta_0)\mbox{]}_{\hcalF^c}- \frac{\alpha_0+\alpha}{2\alpha}\mbox{[}\w^*(\alpha_0,\beta_0)\mbox{]}_{\hcalF^c}||^2 \nonumber \\
\leq& \frac{(\alpha-\alpha_0)^2}{4\alpha^2}||\w^*(\alpha_0,\beta_0)||^2-\frac{(\alpha_0+\alpha)^2}{4\alpha^2}||\mbox{[}\w^*(\alpha_0,\beta_0)\mbox{]}_{\hcalF}||^2.\nonumber
\end{align}
The proof is complete. 

\endproof

\subsection{Proof for Lemma \ref{lemma:dual-estimation}}
\proof of  Lemma \ref{lemma:dual-estimation}:
Firstly, we need to extend the definition of $D(\theta; \alpha,\beta)$ to $\R^n$:
\begin{align}
\tilde{D}(\theta; \alpha, \beta) = \left\{
\begin{array}{ccc}
D(\theta; \alpha, \beta),&\mbox{ if } \theta \in \mbox{[}0,1\mbox{]}^n,\\
+\infty,&\mbox{ otherwise. } 
\end{array}
\right. 
\end{align} 
 Due to the strong convexity of objective $\tilde{D}(\theta;\alpha,\beta)$, we have 
\begin{align}
& \tilde{D}(\theta^*(\alpha_0, \beta_0), \alpha, \beta_0) \geq \tilde{D}(\theta^*(\alpha,\beta_0),\alpha,\beta_0) +\frac{\gamma}{2n}||\theta^*(\alpha_0,\beta_0)-\theta^*(\alpha, \beta_0)||^2,\nonumber \\
& \tilde{D}(\theta^*(\alpha, \beta_0), \alpha_0, \beta_0) \geq \tilde{D}(\theta^*(\alpha_0,\beta_0),\alpha_0,\beta_0) +\frac{\gamma}{2n}||\theta^*(\alpha_0,\beta_0)-\theta^*(\alpha, \beta_0)||^2.\nonumber 
\end{align}
Since $\theta^*(\alpha_0, \beta_0), \theta^*(\alpha, \beta_0) \in \mbox{[}0,1\mbox{]}^n$, the above inequalities are equivalent to  
\begin{align}
&\frac{1}{2\alpha}f_{\beta_0}(\theta^*(\alpha_0, \beta_0)) +\frac{\gamma}{2n}||\theta^*(\alpha_0, \beta_0)||^2 - \frac{1}{n}\langle \textbf{1},{\theta^*(\alpha_0, \beta_0)} \rangle \nonumber \\
&\geq \frac{1}{2\alpha}f_{\beta_0}(\theta^*(\alpha, \beta_0)) +\frac{\gamma}{2n}||\theta^*(\alpha, \beta_0)||^2 - \frac{1}{n}\langle \textbf{1},\theta^*(\alpha, \beta_0) \rangle +\frac{\gamma}{2n}||\theta^*(\alpha_0, \beta_0)-\theta^*(\alpha, \beta_0)||^2,\nonumber\\
&\frac{1}{2\alpha_0}f_{\beta_0}(\theta^*(\alpha, \beta_0)) +\frac{\gamma}{2n}||\theta^*(\alpha, \beta_0)||^2 - \frac{1}{n}\langle \textbf{1},{\theta^*(\alpha, \beta_0)} \rangle \nonumber \\
& \geq  \frac{1}{2\alpha_0}f_{\beta_0}( \theta^*(\alpha_0, \beta_0)) +\frac{\gamma}{2n}||\theta^*(\alpha_0, \beta_0)||^2 - \frac{1}{n}\langle \textbf{1},\theta^*(\alpha_0, \beta_0) \rangle +\frac{\gamma}{2n}||\theta^*(\alpha_0, \beta_0)-\theta^*(\alpha, \beta_0)||^2.\nonumber
\end{align}
Adding the above two inequalities, we obtain 
\begin{align}
&\frac{\gamma (\alpha-\alpha_0)}{2n}||\theta^*(\alpha_0, \beta_0)||^2 - \frac{\alpha-\alpha_0}{n}\langle \textbf{1},{\theta^*(\alpha_0, \beta_0)} \rangle \nonumber \\
&\geq \frac{\gamma (\alpha-\alpha_0)}{2n}||\theta^*(\alpha, \beta_0)||^2 - \frac{\alpha-\alpha_0}{n}\langle \textbf{1},\theta^*(\alpha, \beta_0) \rangle  + \frac{\gamma(\alpha_0 +\alpha)}{2n}||\theta^*(\alpha_0, \beta_0)-\theta^*(\alpha, \beta_0)||^2.\nonumber
\end{align}
That is equivalent to 
\begin{align}
&||\theta^*(\alpha,\beta_0)||^2- \langle \frac{\alpha-\alpha_0}{\gamma \alpha } \textbf{1}+\frac{\alpha_0 +\alpha}{ \alpha }\theta^*(\alpha_0, \beta_0) ,\theta^*(\alpha, \beta_0) \rangle\nonumber \\
&\leq -\frac{ \alpha_0 }{ \alpha }||\theta^*(\alpha_0, \beta_0)||^2 - \frac{\alpha-\alpha_0}{\gamma \alpha }\langle \textbf{1},{\theta^*(\alpha_0, \beta_0)}\rangle. 
\end{align}
That is 
\begin{align}
||\theta^*(\alpha,\beta_0)-(\frac{\alpha-\alpha_0}{2 \gamma \alpha } \textbf{1}+\frac{\alpha_0 +\alpha}{ 2 \alpha }\theta^*(\alpha_0, \beta_0)) ||^2 
\leq (\frac{\alpha-\alpha_0}{2\alpha})^2||\theta^*(\alpha_0,\beta_0)-\frac{1}{\gamma}\textbf{1}||^2.\label{eqn:dual-estimation-temp1}
\end{align}
Substituting the priors that $\mbox{[}\theta^*(\alpha, \beta_0)\mbox{]}_{\hcalR} = 0$ and $ \mbox{[}\theta^*(\alpha, \beta_0)\mbox{]}_{\hcalL} = 1$ into \eqref{eqn:dual-estimation-temp1}, we have 
\begin{align}
&||\mbox{[}\theta^*(\alpha,\beta_0)\mbox{]}_{\hcalD^c} - (\frac{\alpha-\alpha_0}{2 \gamma \alpha } \textbf{1}+\frac{\alpha_0 +\alpha}{ 2 \alpha }\mbox{[}\theta^*(\alpha_0, \beta_0)\mbox{]}_{\hcalD^c})||^2 \nonumber \\
&\leq (\frac{\alpha-\alpha_0}{2\alpha})^2||\theta^*(\alpha_0,\beta_0)-\frac{1}{\gamma}\textbf{1}||^2 - ||\frac{(2\gamma-1)\alpha+\alpha_0}{2 \gamma \alpha } \textbf{1}-\frac{\alpha_0 +\alpha}{ 2 \alpha }\mbox{[}\theta^*(\alpha_0, \beta_0)\mbox{]}_{\hcalL} ||^2\nonumber \\
&-|| \frac{\alpha-\alpha_0}{2 \gamma \alpha } \textbf{1}+\frac{\alpha_0 +\alpha}{ 2 \alpha }\mbox{[}\theta^*(\alpha_0, \beta_0)\mbox{]}_{\hcalR} ||^2. \nonumber 
\end{align}
The proof is complete.
\endproof

\subsection{Proof for Lemma \ref{lemma:opt-feature}}\label{sec:opt-s}
Before the proof of Lemma \ref{lemma:opt-feature}, we should prove that the optimization problem in (\ref{sub-F}) is equivalent to 
\begin{align}
s^i(\alpha, \beta_0) =\max_{\theta \in \Theta}\left \{\frac{1}{n} |\langle \mbox{[}\xb^i\mbox{]}_{\hcalD^c}, \theta \rangle  + \langle  \mbox{[}\xb^i\mbox{]}_{\hcalL},\textbf{1} \rangle| \right\}, i \in \hcalF^c.
\end{align}
To avoid notational confusion, we denote the feasible region $\Theta$ and $\theta$ in  \eqref{sub-F}  as $\tilde{\Theta}$ and $\tilde{\theta}$, respectively. Then,
\begin{align}
&\max_{\tilde{\thetab} \in \tilde{\Theta}}\left\{\left|\frac{1}{n}\mbox{[}\Xb \tilde{\thetab}\mbox{]}_i\right|\right\} = \max_{\tilde{\thetab} \in \tilde{\Theta}}\left\{\frac{1}{n}\left|\xb^i \tilde{\thetab}\right|\right\} \nonumber\\
=&\max_{\tilde{\thetab} \in \tilde{\Theta}}\left\{\frac{1}{n}\left|\mbox{[}\xb^i\mbox{]}_{\hcalD^c} \mbox{[}\tilde{\thetab}\mbox{]}_{\hcalD^c}+\mbox{[}\xb^i\mbox{]}_{\hcalL} \mbox{[}\tilde{\thetab}\mbox{]}_{\hcalL}+\mbox{[}\xb^i\mbox{]}_{\hcalR} \mbox{[}\tilde{\thetab}\mbox{]}_{\hcalR}\right|\right\} \nonumber\\
=&\max_{\tilde{\thetab} \in \tilde{\Theta}}\left \{\frac{1}{n} |\langle \mbox{[}\xb^i\mbox{]}_{\hcalD^c}, \mbox{[}\tilde{\thetab}\mbox{]}_{\hcalD^c} \rangle  + \langle  \mbox{[}\xb^i\mbox{]}_{\hcalL},\textbf{1} \rangle| \right\}\nonumber\\
=&\max_{\thetab \in \Theta}\left \{\frac{1}{n} |\langle \mbox{[}\xb^i\mbox{]}_{\hcalD^c}, \thetab \rangle  + \langle  \mbox{[}\xb^i\mbox{]}_{\hcalL},\textbf{1} \rangle| \right\}=s^i(\alpha, \beta_0).\nonumber
\end{align}
The last two equations hold since $\mbox{[}\theta\mbox{]}_{\hcalL} = \textbf{1}$, $\mbox{[}\theta\mbox{]}_{\hcalR} = 0$, and $\mbox{[}\theta_{\hcalD^c}\mbox{]}\in \Theta$. 

\proof of Lemma \ref{lemma:opt-feature}:
\begin{align}
s^i(\alpha, \beta_0) =&\max_{\theta \in B(\mathbf{c},r)} \{\frac{1}{n} |\langle \mbox{[}\xb^i\mbox{]}_{\hcalD^c}, \theta \rangle  + \langle  \mbox{[}\xb^i\mbox{]}_{\hcalL},\textbf{1} \rangle| \}\nonumber\\
=& \max_{\eta \in B(\mathbf{0},r)} \{\frac{1}{n} |\langle \mbox{[}\xb^i\mbox{]}_{\hcalD^c}, \mathbf{c} \rangle  + \langle  \mbox{[}\xb^i\mbox{]}_{\hcalL},\textbf{1}\rangle + \langle \mbox{[}\xb^i\mbox{]}_{\hcalD^c}, \eta  \rangle| \}\nonumber \\
=& \frac{1}{n} \big(|\langle \mbox{[}\xb^i\mbox{]}_{\hcalD^c}, \mathbf{c} \rangle  + \langle  \mbox{[}\xb^i\mbox{]}_{\hcalL},\textbf{1} \rangle| +\| \mbox{[}\xb^i\mbox{]}_{\hcalD^c}\| r\big). \nonumber
\end{align}
The last equality holds since $-\| \mbox{[}\xb^i\mbox{]}_{\hcalD^c}\| r \leq \langle \mbox{[}\xb^i\mbox{]}_{\hcalD^c}, \eta  \rangle \leq \| \mbox{[}\xb^i\mbox{]}_{\hcalD^c}\| r$.\\
The proof is complete. 
\endproof

\subsection{Proof for Theorem \ref{thm:feature-screening}}
\proof of Theorem \ref{thm:feature-screening}:\\ 
(1) It can be obtained from the rule \ref{screening-rule-relaxed-1}.\\
(2) It is from the  definition of $\hcalF$.
\endproof

\subsection{Proof for Lemma \ref{lemma:opt-sample}}\label{sec:opt-u-l}
Firstly, we need to point out that the optimization problems in Eqs. (\ref{sub-R}) and (\ref{sub-L}) are equivalent to the problems:
\begin{align}
u_i(\alpha, \beta_0) = \max_{\w \in \calW} \{1- \langle \mbox{[}\xb_i\mbox{]}_{\hcalF^c}, \w \rangle \}, i\in \hcalD^c,\nonumber\\
l_i(\alpha, \beta_0) = \min_{\w \in \calW} \{1- \langle \mbox{[}\xb_i\mbox{]}_{\hcalF^c}, \w \rangle \}, i\in \hcalD^c.\nonumber
\end{align}
They follow from the fact that $ \mbox{[}\w\mbox{]}_{\hcalF^c} \in \calW$ and 
\begin{align}
&\{1-\langle \w, \xb_i  \rangle\}\nonumber\\ 
=& \{1-\langle \mbox{[}\w\mbox{]}_{\hcalF^c}, \mbox{[}\xb_i\mbox{]}_{\hcalF^c}  \rangle-\langle \mbox{[}\w\mbox{]}_{\hcalF}, \mbox{[}\xb_i\mbox{]}_{\hcalF}  \rangle\}\nonumber \\
=&\{1-\langle \mbox{[}\w\mbox{]}_{\hcalF^c}, \mbox{[}\xb_i\mbox{]}_{\hcalF^c}  \rangle\}\mbox{ (since $\mbox{[}\w\mbox{]}_{\hcalF}=0$)}.\nonumber
\end{align}

\proof of Lemma \ref{lemma:opt-sample}:
\begin{align}
u_i(\alpha, \beta_0) =& \max_{\w \in B(\mathbf{c},r)} \{1- \langle \mbox{[}\xb_i\mbox{]}_{\hcalF^c}, \w \rangle \}\nonumber \\
= &\max_{\eta \in B(\mathbf{0},r)} \{1- \langle \mbox{[}\xb_i\mbox{]}_{\hcalF^c}, \mathbf{c} \rangle-  \langle \mbox{[}\xb_i\mbox{]}_{\hcalF^c}, \eta \rangle\} \nonumber\\
= & 1- \langle \mbox{[}\xb_i\mbox{]}_{\hcalF^c}, \mathbf{c} \rangle + \max_{\eta \in B(\mathbf{0},r)} \{-  \langle \mbox{[}\xb_i\mbox{]}_{\hcalF^c}, \eta \rangle\}\nonumber\\
= & 1- \langle \mbox{[}\xb_i\mbox{]}_{\hcalF^c}, \mathbf{c} \rangle + \|\mbox{[}\xb_i\mbox{]}_{\hcalF^c} \|r.\nonumber
\end{align}
\begin{align}
l_i(\alpha, \beta_0) =& \min_{\w \in B(\mathbf{c},r)} \{1- \langle \mbox{[}\xb_i\mbox{]}_{\hcalF^c}, \w \rangle \}\nonumber \\
= &\min_{\eta \in B(\mathbf{0},r)} \{1- \langle \mbox{[}\xb_i\mbox{]}_{\hcalF^c}, \mathbf{c} \rangle-  \langle \mbox{[}\xb_i\mbox{]}_{\hcalF^c}, \eta \rangle\} \nonumber\\
= & 1- \langle \mbox{[}\xb_i\mbox{]}_{\hcalF^c}, \mathbf{c} \rangle + \min_{\eta \in B(\mathbf{0},r)} \{-  \langle \mbox{[}\xb_i\mbox{]}_{\hcalF^c}, \eta \rangle\}\nonumber\\
= & 1- \langle \mbox{[}\xb_i\mbox{]}_{\hcalF^c}, \mathbf{c} \rangle - \|\mbox{[}\xb_i\mbox{]}_{\hcalF^c} \|r.\nonumber
\end{align}
The proof is complete. 
\endproof

\subsection{Proof for Theorem \ref{thm:sample-screening}}
\proof of Theorem \ref{thm:sample-screening}:\\
(1) It can be obtained from the rule \ref{screening-rule-relaxed-2}.\\
(2) It is from the  definitions of $\hcalR$ and $\hcalL$. 
\endproof

\subsection{Proof for Theorem \ref{thm:order-of-screening}}
\proof of Theorem \ref{thm:order-of-screening}:\\
(1) Given the reference solutions pair $\w^*(\alpha_{i-1,j}, \beta_j)$ and $\theta^*(\alpha_{i-1,j}, \beta_j)$, if we do ISS first in SIFS and apply ISS and IFS for infinite times. If after $s$ times of triggering, no new inactive features or samples are identified, then we can denote the sequence of $\hcalF, \hcalR$, and $\hcalL$ as:
\begin{align}
&\hcalF_0^A=\hcalR_0^A=\hcalL_0^A=\emptyset \overset{ISS}\longrightarrow \hcalF_1^A, \hcalR_1^A, \hcalL_1^A \overset{IFS}\longrightarrow \hcalF_2^A, \hcalR_2^A, \hcalL_2^A \overset{ISS}\longrightarrow ...\hcalF_{s}^A, \hcalR_s^A, \hcalL_s^A \overset{IFS / ISS}\longrightarrow ...\label{seq:ISS-first}\\
&\mbox{with }\hcalF_{s}^A=\hcalF_{s+1}^A=\hcalF_{s+2}^A=..., \hcalR_s^A=\hcalR_{s+1}^A=\hcalR_{s+2}^A=...\mbox{ and } \hcalL_s^A=\hcalL_{s+1}^A=\hcalL_{s+2}^A=...\label{seq:ISS-first-A}
\end{align}
In the same way, if we do IFS first in SIFS and no new inactive features or samples are identified after $t$ times of triggering of ISS and IFS, then the sequence can be denoted as: 
\begin{align}
&\hcalF_0^B=\hcalR_0^B=\hcalL_0^B=\emptyset \overset{IFS}\longrightarrow \hcalF_1^B, \hcalR_1^B, \hcalL_1^B \overset{ISS}\longrightarrow \hcalF_2^B, \hcalR_2^B, \hcalL_2^B \overset{IFS}\longrightarrow ...\hcalF_{t}^B, \hcalR_t^B, \hcalL_t^B \overset{IFS / ISS}\longrightarrow ...\label{seq:IFS-first}\\
&\mbox{with }\hcalF_{t}^B=\hcalF_{t+1}^B=\hcalF_{t+2}^B=..., \hcalR_t^B=\hcalR_{t+1}^B=\hcalR_{t+2}^B=...\mbox{and } \hcalL_t^B=\hcalL_{t+1}^B=\hcalL_{t+2}^B=...\label{seq:ISS-first-B}
\end{align}
We first prove that $\hcalF_k^B \subseteq \hcalF_{k+1}^A, \hcalR_k^B\subseteq \hcalR_{k+1}^A$ and $\hcalL_k^B \subseteq \hcalL_{k+1}^A$ hold for all $k\geq0$ by induction.\\

1) When $k=0$, the equalities $\hcalF_0^B \subseteq \hcalF_{1}^A, \hcalR_0^B\subseteq \hcalR_{1}^A$ and $\hcalL_0^B \subseteq \hcalL_{1}^A$ hold since $\hcalF_0^B=\hcalR_0^B=\hcalL_0^B=\emptyset$.

2) If $\hcalF_k^B \subseteq \hcalF_{k+1}^A, \hcalR_k^B\subseteq \hcalR_{k+1}^A$ and $\hcalL_k^B \subseteq \hcalL_{k+1}^A$ hold, by the synergistic effect of ISS and IFS, we have that $\hcalF_{k+1}^B \subseteq \hcalF_{k+2}^A, \hcalR_{k+1}^B\subseteq \hcalR_{k+2}^A$ and $\hcalL_{k+1}^B \subseteq \hcalL_{k+2}^A$ hold. 

Thus, $\hcalF_k^B \subseteq \hcalF_{k+1}^A, \hcalR_k^B\subseteq \hcalR_{k+1}^A$ and $\hcalL_k^B \subseteq \hcalL_{k+1}^A$ hold for all $k\geq0$.

Similar to the analysis in (1), we can also prove that $\hcalF_k^A \subseteq \hcalF_{k+1}^B, \hcalR_k^A\subseteq \hcalR_{k+1}^B$ and $\hcalL_k^A \subseteq \hcalL_{k+1}^B$ hold for all $k\geq0$.

Combining (1) and (2), we can acquire
\begin{align}
\hcalF_0^B \subseteq \hcalF_{1}^A \subseteq \hcalF_{2}^B \subseteq \hcalF_{3}^A....\label{seq:F1}\\
\hcalF_0^A \subseteq \hcalF_{1}^B \subseteq \hcalF_{2}^A \subseteq \hcalF_{3}^B.... \label{seq:F2} \\
\hcalR_0^B \subseteq \hcalR_{1}^A \subseteq \hcalR_{2}^B \subseteq \hcalR_{3}^A....\label{seq:R1}\\
\hcalR_0^A \subseteq \hcalR_{1}^B \subseteq \hcalR_{2}^A \subseteq \hcalR_{3}^A....\label{seq:R2} \\
\hcalL_0^B \subseteq \hcalL_{1}^A \subseteq \hcalL_{2}^B \subseteq \hcalL_{3}^A....\label{seq:L1} \\
\hcalL_0^A \subseteq \hcalL_{1}^B \subseteq \hcalL_{2}^A \subseteq \hcalL_{3}^B....\label{seq:L2}
\end{align} 
By the first equality of Eqs. (\ref{seq:ISS-first-A}), (\ref{seq:F1}), and (\ref{seq:F2}), we can obtain $\hcalF_{p}^A=\hcalF_{t}^B$. Similarly, we can obtain $\hcalR_{p}^A=\hcalR_{t}^B$ and $\hcalL_{p}^A=\hcalL_{t}^B$.  

(2) If $p$ is odd, then by Eqs. (\ref{seq:F1}), (\ref{seq:R1}, and (\ref{seq:L1}), we have $\hcalF_s^A \subseteq \hcalF_{p+1}^B, \hcalR_s^A \subseteq \hcalR_{p+1}^B$, and $\hcalL_s^A \subseteq \hcalL_{p+1}^B$. Thus $q\leq p+1$.

Else if $p$ is even, then by Eqs. (\ref{seq:F2}), (\ref{seq:R2}), and (\ref{seq:L2}), we have $\hcalF_s^A \subseteq \hcalF_{p+1}^B, \hcalR_s^A \subseteq \hcalR_{p+1}^B$, and $\hcalL_s^A \subseteq \hcalL_{p+1}^B$. Thus $q\leq p+1$.

Doing the same analysis for $q$, we can obtain $p\leq q+1$. 

Hence, $|p-q|\leq 1$.

The proof is complete.
\endproof

\subsection{Proof for Theorem \ref{thm:dual-kkt-MSSVM}}
\proof of Theorem \ref{thm:dual-kkt-MSSVM}:

We notice $(\X_i^k)^\top\w = \w_k^\top\x_i - \w_{y_i}^\top\x_i$. By denoting $\phi_i(\a)= \sum_{k=1}^K\ell([\u_i]_k+[\a]_k)$, it is easy to verify that $\ell_i(\w)=\phi_i(\X_i^\top\w)$. Thus, the problem (\ref{eqn:primal-MSSVM}) can be rewritten as: 
\begin{align}
\min_{\w\in \R^{Kp}} P(\w;\alpha, \beta) = \frac{1}{n}\sum\limits_{i=1}^{n}\phi_i(\X_i^\top\w)+ \frac{\alpha}{2}\|\w\|^2+ \beta ||\w||_1.\nonumber
\end{align}
Let $\z_i= \X_i^\top \w $. The primal problem (\ref{eqn:primal}) is then equivalent to
\begin{align}
&\min_{\w\in \R^p, \z \in \R^n} \frac{\alpha}{2}||\w||^2 + \beta ||\w||_1 + \frac{1}{n}\sum\limits_{i=1}^{n}\phi_i(\z_i) ,\nonumber\\
&~~~~~~~~~~~\mbox{s.t. } \z_i = \X_i^\top \w, i = 1,2,..,n. \nonumber
\end{align}
The Lagrangian then becomes
\begin{align}
L(\w,\z,\theta) =& \frac{\alpha}{2}||\w||^2 + \beta ||\w||_1 + \frac{1}{n}\sum\limits_{i=1}^{n}\phi_i(\z_i)+ \frac{1}{n}\sum_{i=1}^n\langle \X_i^\top\w - \z_i, \theta_i\rangle \nonumber \\
=& \underbrace{ \frac{\alpha}{2}||\w||^2 + \beta ||\w||_1 +\frac{1}{n}\langle \X \theta, \w \rangle }_{:=f_1(\w)} +\underbrace{ \frac{1}{n}\sum\limits_{i=1}^{n}(\phi_i(\z_i)-\langle \z_i, \theta_i \rangle)}_{:=f_2(\z)}.\label{eqn:dual-Lagrange-MSSVM}
\end{align}
We first consider the subproblem $\min_{\w} L(\w,\z,\theta)$:
\begin{align}
& 0\in \partial_{\w} L(\w,\z, \thetab) = \partial_{\w} f_1(\w) = \alpha \w+\frac{1}{n}\X \thetab + \beta \partial ||\w||_1 \Leftrightarrow \nonumber\\
&\frac{1}{n}\X \thetab \in -\alpha \w  -\beta \partial ||\w||_1 \Rightarrow \w = -\frac{1}{\alpha}\calS_{\beta}(\frac{1}{n}\X \thetab). \label{eqn:primal-KKT1-MSSVM}
\end{align}
By substituting \eqref{eqn:primal-KKT1-MSSVM} into $f_1(\w)$, we obtain 
\begin{align}
f_1(\w) =  \frac{\alpha}{2}||\w||^2 + \beta ||\w||_1-\langle \alpha \w+\beta\partial||\w||_1, \w \rangle= -\frac{\alpha}{2} ||\w||^2 = -\frac{1}{2\alpha}||\calS_{\beta}(\frac{1}{n}\X \thetab)||^2. \label{eqn:min-f1-MSSVM}
\end{align}
Then, we consider the problem $\min_{\z} L(\w,\z,\thetab)$:
\begin{align}
0 =& \nabla_{[\z_i]_k} L(\w,\z,\thetab) = \nabla_{[\z_i]_k} f_2(\z) = \left\{
\begin{array}{ccc}
-\frac{1}{n}\mbox{[}\theta_i\mbox{]}_k,&\mbox{if }\mbox{[}\z_i\mbox{]}_k+[\u_i]_k<0,\\
\frac{1}{\gamma n} (\mbox{[}\z_i\mbox{]}_k+[\u_i]_k)-\frac{1}{n}\mbox{[}\theta_i\mbox{]}_k,&\mbox{if } 0 \leq \mbox{[}\z_i\mbox{]}_k+[\u_i]_k \leq \gamma,\\
\frac{1}{n}-\frac{1}{n}\mbox{[}\theta_i\mbox{]}_k, &\mbox{if } \mbox{[}\z_i\mbox{]}_k +[\u_i]_k> \gamma.
\end{array}
\right.  \nonumber\\
\Rightarrow &\mbox{[}\theta_i\mbox{]}_k = \left\{
\begin{array}{ccc}
0,&\mbox{ if }\mbox{[}\z_i\mbox{]}_k+[\u_i]_k<0,\\
\frac{1}{\gamma } (\mbox{[}\z_i\mbox{]}_k+[\u_i]_k),&\mbox{ if } 0 \leq \mbox{[}\z_i\mbox{]}_k+[\u_i]_k \leq \gamma,\\
1, &\mbox{ if } \mbox{[}\z_i\mbox{]}_k +[\u_i]_k> \gamma.
\end{array}
\right.  \label{eqn:primal-KKT2-MSSVM}
\end{align}
Thus, we have
\begin{align}
f_2(\z) = \left\{
\begin{array}{ccc}
-\frac{\gamma}{2n}||\theta||^2+\frac{1}{n}\langle \u, \theta\rangle,&\mbox{ if }\mbox{[}\theta_i\mbox{]}_k\in \mbox{[}0,1\mbox{]}, \forall i\in \mbox{[}n\mbox{]}, k \in [K],\\
-\infty, &\mbox{ otherwise } .
\end{array}
\right.  \label{eqn:min-f2-MSSVM}
\end{align}
Combining Eqs.(\ref{eqn:dual-Lagrange-MSSVM}), (\ref{eqn:min-f1-MSSVM}), and (\ref{eqn:min-f2-MSSVM}), we obtain the dual problem:
\begin{align}
\min_{\thetab \in \mbox{[}0,1\mbox{]}^{Kn}} \frac{1}{2\alpha}||\calS_{\beta}(\frac{1}{n}\X \thetab)||^2 +\frac{\gamma}{2n}||\theta||^2 - \frac{1}{n}\langle \u,{\theta} \rangle.\nonumber 
\end{align}
\endproof
\subsection{Proof for Lemma \ref{lemma:dual-scaled-MSSVM}}
Lemma  \ref{lemma:dual-scaled-MSSVM} can be proved quite similarly with Lemma \ref{lemma:dual-scaled}. Therefore, we omit this proof here. 
\subsection{Proof for Lemma \ref{beta-max-and-alpha-max-MSSVM}}
\proof of Lemma \ref{beta-max-and-alpha-max-MSSVM}:

(i) We prove this lemma by verifying that the solutions $\w^*(\alpha, \beta) = \textbf{0}$ and $\thetab^*(\alpha, \beta) = \u$ satisfy the 
conditions (\ref{eqn:KKT1-MSSVM}) and (\ref{eqn:KKT2-MSSVM}).

Firstly, since $\beta \geq \beta_{\max} = ||\frac{1}{n}\X \u||_{\infty}$, we have $\calS_{\beta}(\frac{1}{n}\X \u) = 0$. Thus, $\w^*(\alpha,\beta) = \textbf{0}$ and $ \theta^*(\alpha,\beta) = \u$ satisfy the condition (\ref{eqn:KKT1-MSSVM}). Then, for all $i\in \mbox{[}n\mbox{]}$, we have 
\begin{align}
\langle \X_i^k,\w^*(\alpha,\beta)\rangle+[\u_i]_k = [\u_i]_k = \begin{cases}
1,\hspace{3mm} \mbox{ if }  k \neq y_i;\\
0,\hspace{3mm} \mbox{ if } k = y_i;
\end{cases} \forall i =1,...,n, k= 1, ..., K. \nonumber
\end{align}
Thus, $\w^*(\alpha,\beta) = \textbf{0}$ and $\thetab^*(\alpha,\beta) = \u$ satisfy the condition (\ref{eqn:KKT2-MSSVM}). Hence, they are the solutions of the primal problem (\ref{eqn:primal-MSSVM}) and the dual problem (\ref{eqn:dual-MSSVM}), respectively.

(ii) Similar to the proof of (i), we prove this by verifying that the solutions $\w^*(\alpha, \beta) = \frac{1}{\alpha}\calS_{\beta}(-\frac{1}{n}\X \u)$ and $\thetab^*(\alpha, \beta) = \u$ satisfy the conditions (\ref{eqn:KKT1-MSSVM}) and (\ref{eqn:KKT2-MSSVM}).
\begin{enumerate}
	\item \textbf{Case 1:} $\alpha_{\max}(\beta) \leq 0$. Then for all $\alpha >0$, we have 
	\begin{align}
	&\mbox{if } k = y_i, ~~~\langle \X_i^k, \w^*(\alpha, \beta)\rangle + \u_i^k = \u_i^k = 0,\nonumber\\
	&\mbox{if } k \neq  y_i,~~~  \langle  \X_i^k, \w^*(\alpha, \beta)\rangle + \u_i^k = 1 - \langle  \X_i^k, \frac{1}{\alpha}\calS_{\beta}(\frac{1}{n}\X \u)\rangle \nonumber\\
	&~~~~~~~~~~~~~~~~~~~~~~~~~~~~~~~~~~~~~~~~~= 1 - \frac{1}{\alpha} \langle  \X_i^k, \calS_{\beta}(\frac{1}{n}\X \u)\rangle \geq 1 > \gamma.\nonumber
	\end{align}
	Thus, $\w^*(\alpha, \beta) = \frac{1}{\alpha}\calS_{\beta}(-\frac{1}{n}\X \u)$ and $\thetab^*(\alpha, \beta) = \u$ satisfy the conditions (\ref{eqn:KKT1-MSSVM}) and (\ref{eqn:KKT2-MSSVM}). Hence, they are the optimal solution of problems (\ref{eqn:primal-MSSVM})  and (\ref{eqn:dual-MSSVM}).
	\item \textbf{Case 2:} $\alpha_{\max}(\beta) >0$. Then for any $\alpha \geq  \alpha_{\max}(\beta)$, we have
	\begin{align}
	&\mbox{if } k = y_i, ~~~\langle \X_i^k, \w^*(\alpha, \beta)\rangle + \u_i^k = \u_i^k = 0,\nonumber\\
	&\mbox{if } k \neq  y_i,~~~  \langle  \X_i^k, \w^*(\alpha, \beta)\rangle + \u_i^k = 1 - \langle  \X_i^k, \frac{1}{\alpha}\calS_{\beta}(\frac{1}{n}\X \u)\rangle \nonumber\\
	&~~~~~~~~~~~~~~~~~~~~~~~~~~~~~~~~~~~~~~~~~= 1 - \frac{1}{\alpha} \langle  \X_i^k, \calS_{\beta}(\frac{1}{n}\X \u)\rangle \nonumber \\
	&~~~~~~~~~~~~~~~~~~~~~~~~~~~~~~~~~~~~~~~~~\geq 1 - \frac{\alpha_{\max}(\beta)}{\alpha} (1-\gamma) \geq  1-(1-\gamma) = \gamma.\nonumber
	\end{align}
	Thus, $\w^*(\alpha, \beta) = \frac{1}{\alpha}\calS_{\beta}(-\frac{1}{n}\X \u)$ and $\thetab^*(\alpha, \beta) =  \u$ satisfy the conditions (\ref{eqn:KKT1-MSSVM}) and (\ref{eqn:KKT2-MSSVM}). Hence, they are the optimal solutions of the primal and dual problems (\ref{eqn:primal-MSSVM})  and (\ref{eqn:dual-MSSVM}). 
\end{enumerate}
The proof is complete.
\endproof

\subsection{More Experimental Results}
\begin{figure*}[htb!]
	\begin{center}
		\subfigure[ $\beta/\beta_{\rm{max}}$=0.05]{\includegraphics[scale=0.21]{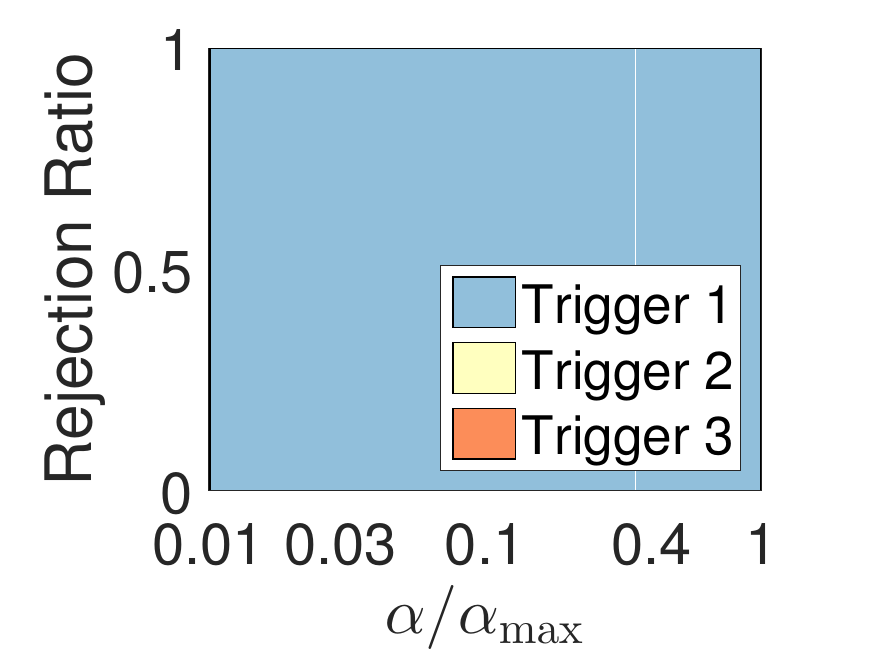}}
		\subfigure[ $\beta/\beta_{\rm{max}}$=0.1]{\includegraphics[scale=0.21]{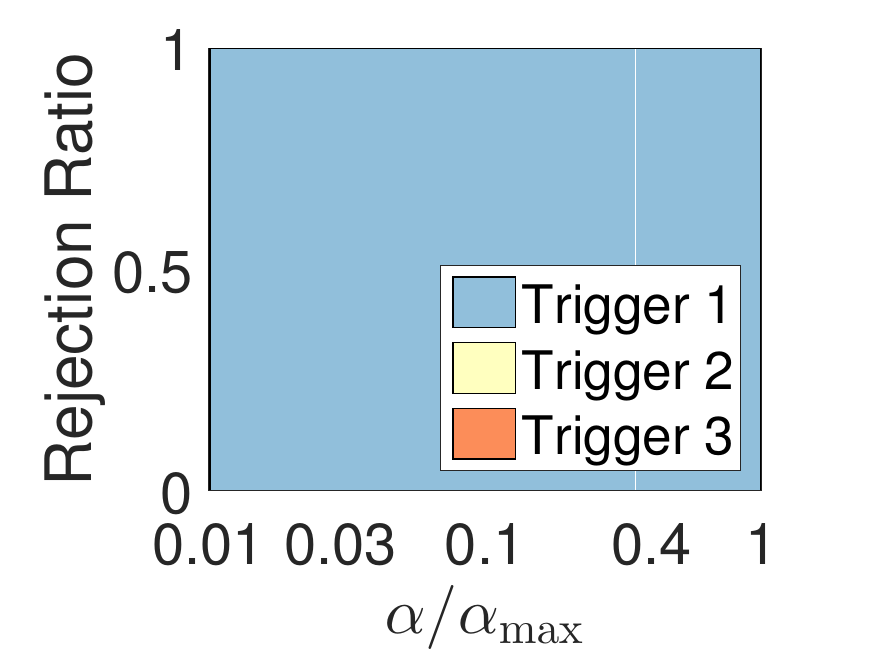}}
		\subfigure[ $\beta/\beta_{\rm{max}}$=0.5]{\includegraphics[scale=0.21]{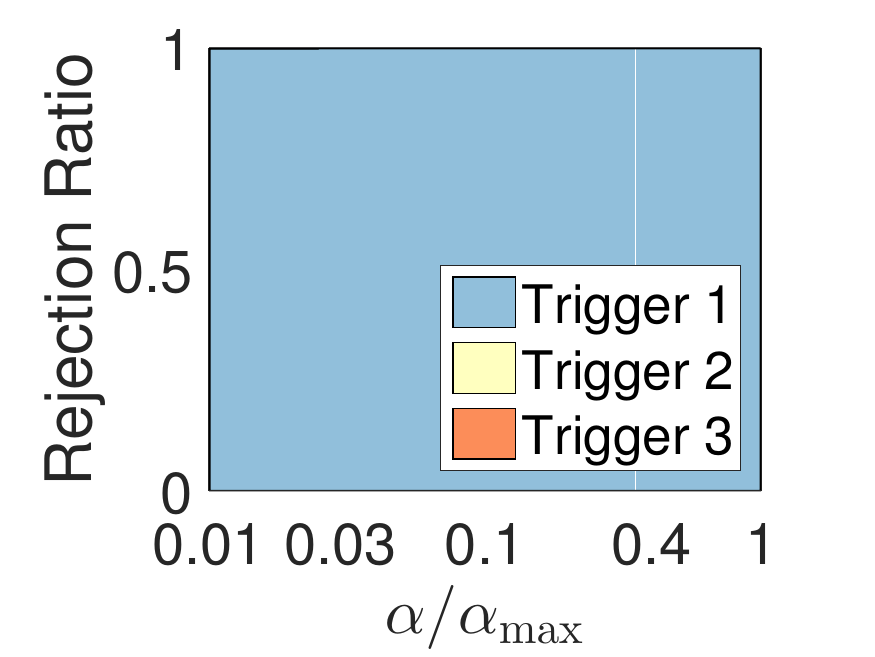}}
		\subfigure[ $\beta/\beta_{\rm{max}}$=0.9]{\includegraphics[scale=0.21]{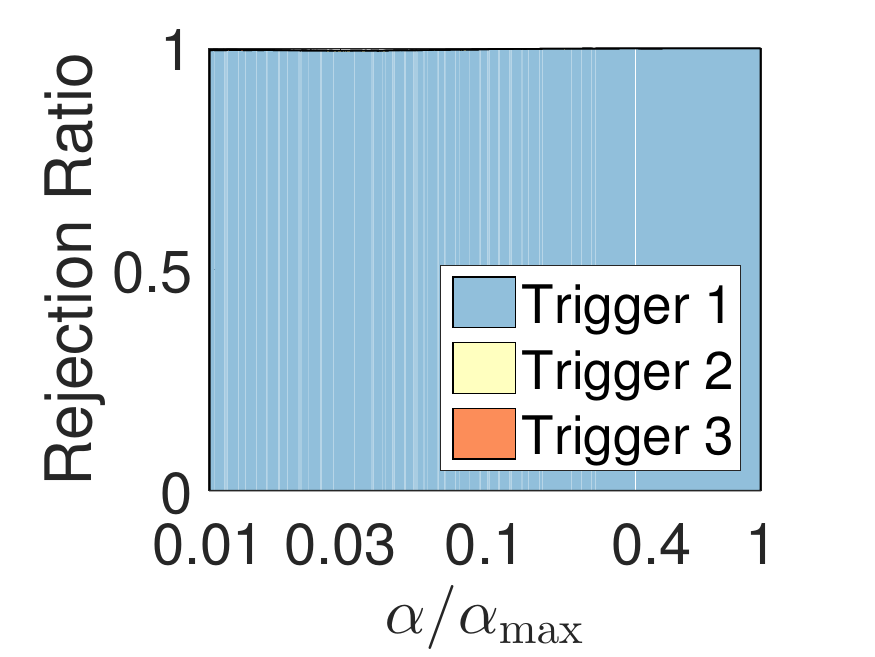}}
		\subfigure[ $\beta/\beta_{\rm{max}}$=0.05]{\includegraphics[scale=0.21]{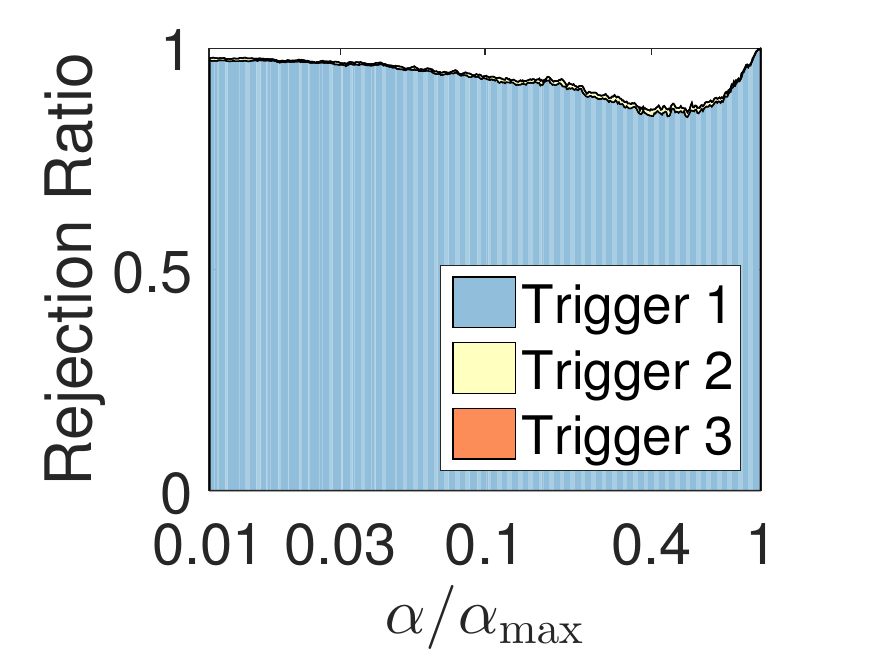}}
		\subfigure[ $\beta/\beta_{\rm{max}}$=0.1]{\includegraphics[scale=0.21]{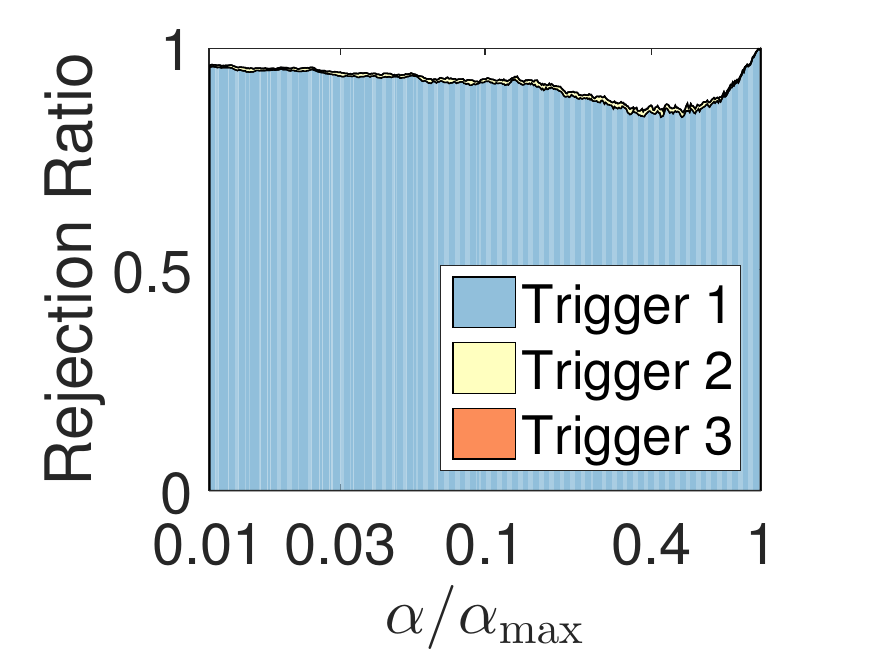}}
		\subfigure[ $\beta/\beta_{\rm{max}}$=0.5]{\includegraphics[scale=0.21]{images/toy_n_1000_d_100000_05_RL_rej_ratio}}
		\subfigure[ $\beta/\beta_{\rm{max}}$=0.9]{\includegraphics[scale=0.21]{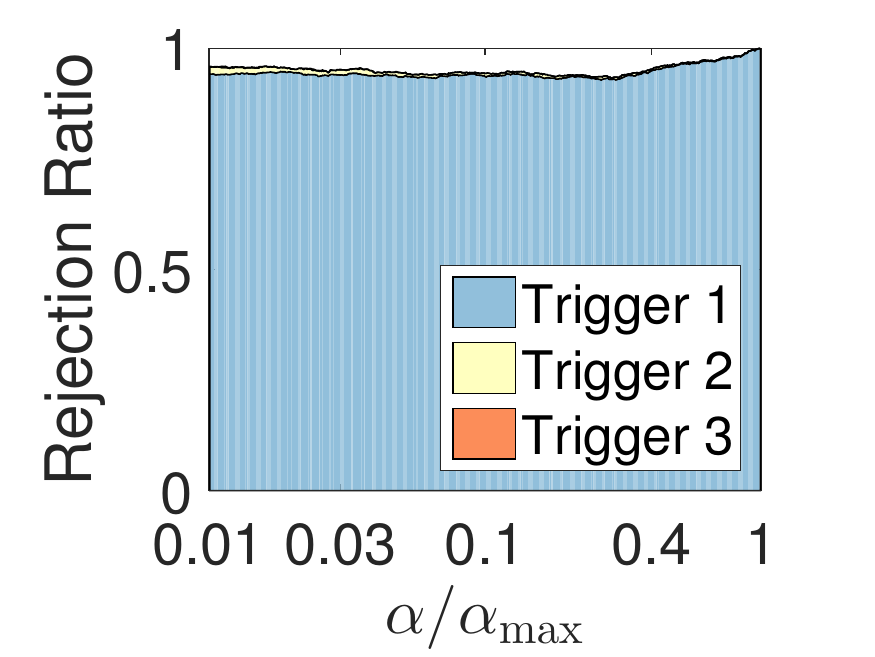}}
		\caption{ Rejection ratios of SIFS on syn3. }
		\label{fig:rejection-ratio-syn-3}
	\end{center}
	\vspace*{-20pt}
\end{figure*}

\begin{figure*}[htb!]
	\begin{center}
		\subfigure[ $\beta/\beta_{\rm{max}}$=0.05]{\includegraphics[scale=0.21]{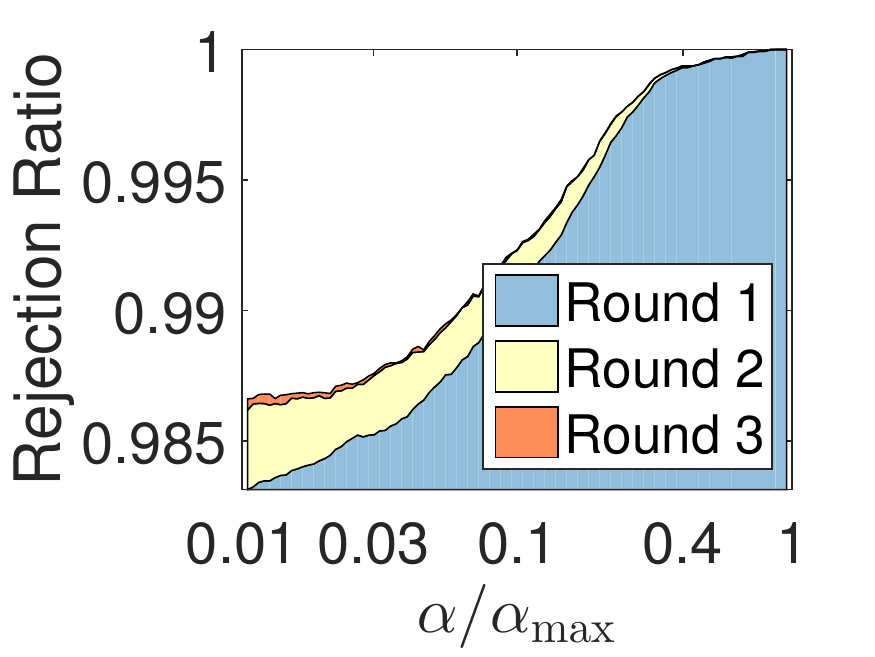}}
		\subfigure[ $\beta/\beta_{\rm{max}}$=0.1]{\includegraphics[scale=0.21]{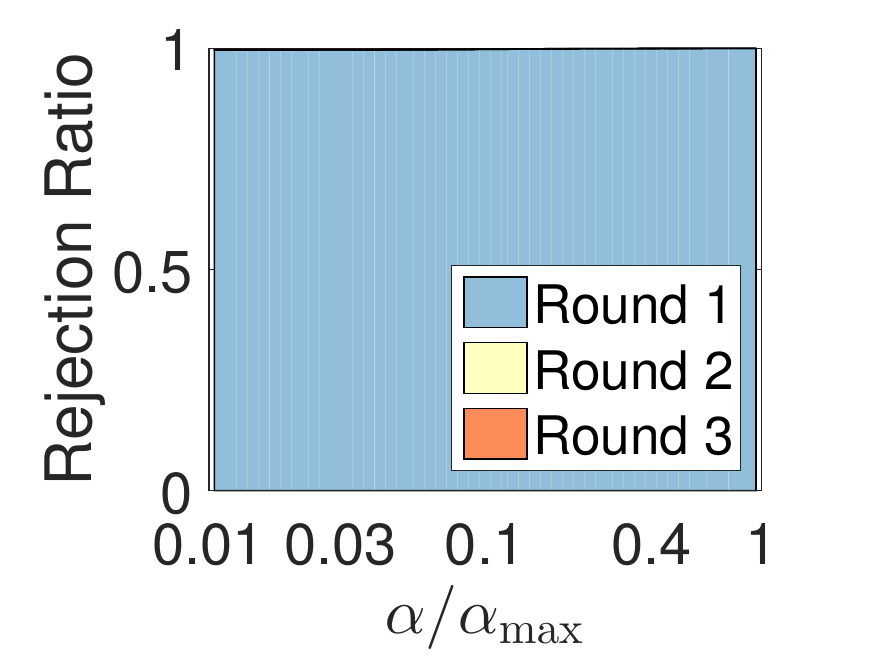}}
		\subfigure[ $\beta/\beta_{\rm{max}}$=0.5]{\includegraphics[scale=0.21]{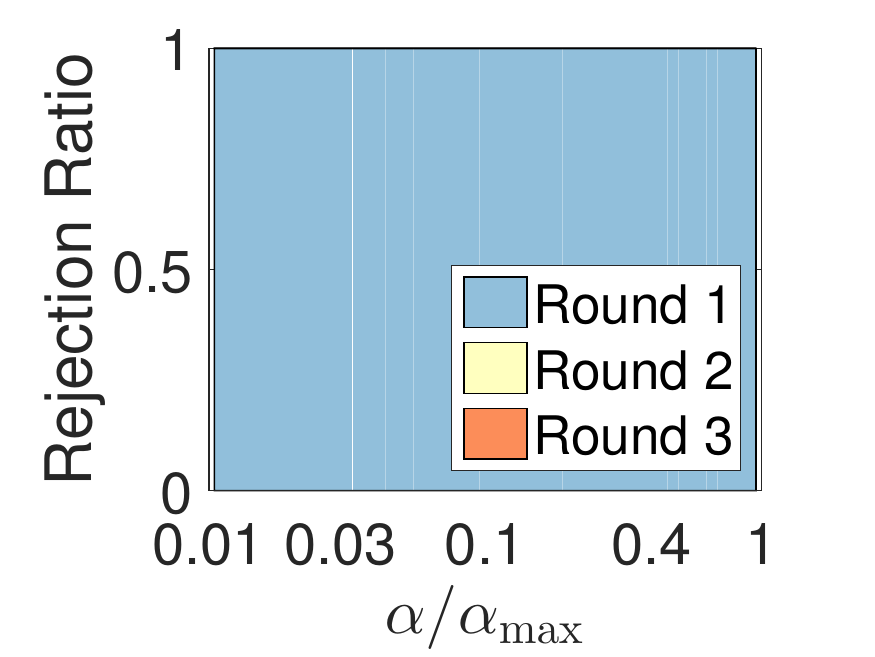}}
		\subfigure[ $\beta/\beta_{\rm{max}}$=0.9]{\includegraphics[scale=0.21]{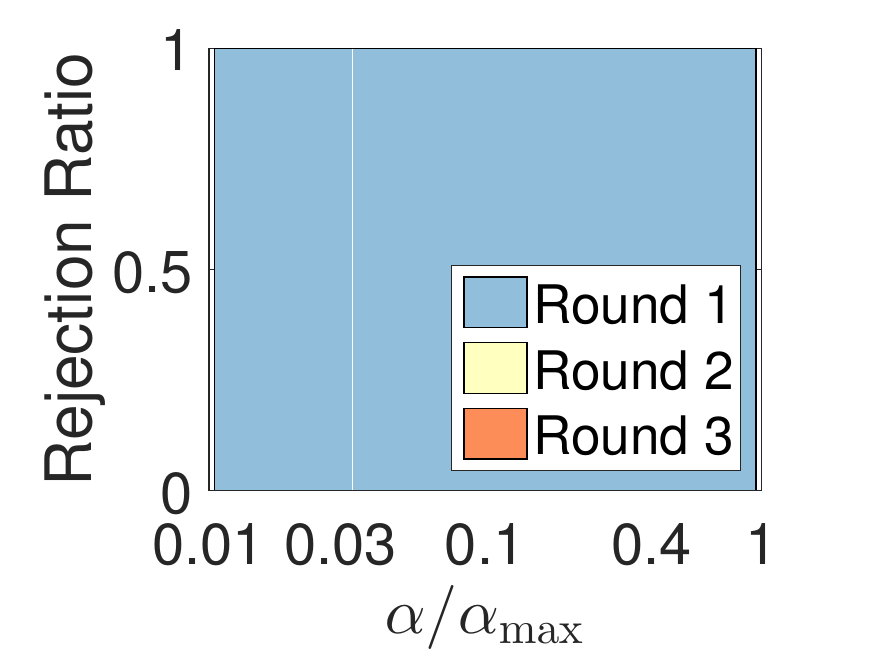}}
		\subfigure[ $\beta/\beta_{\rm{max}}$=0.05]{\includegraphics[scale=0.21]{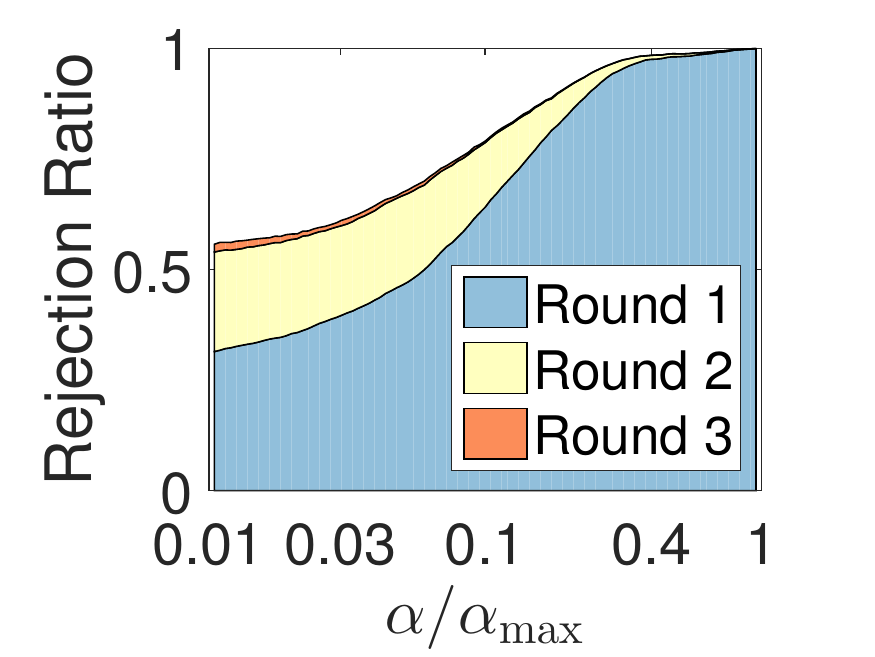}}
		\subfigure[ $\beta/\beta_{\rm{max}}$=0.1]{\includegraphics[scale=0.21]{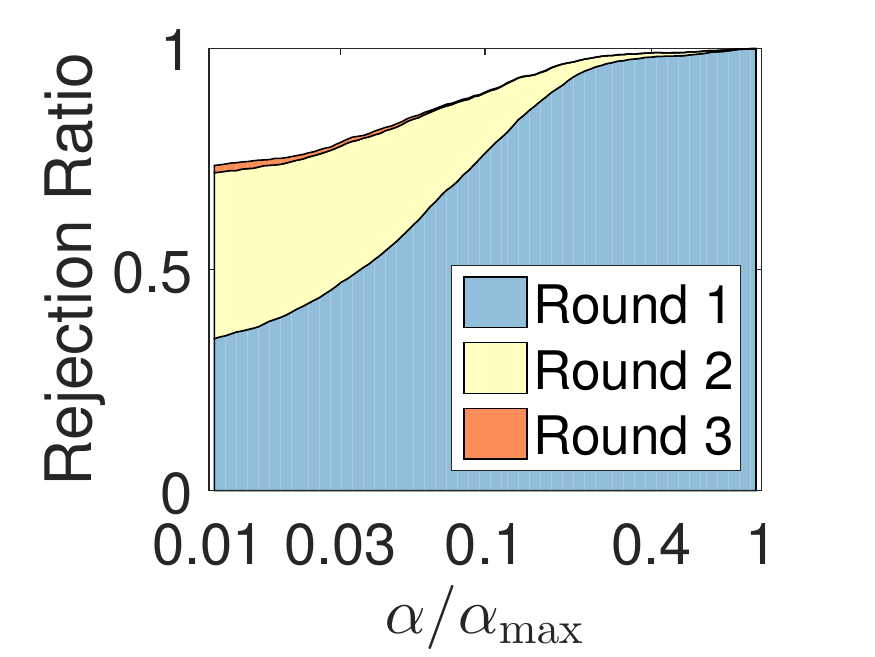}}
		\subfigure[ $\beta/\beta_{\rm{max}}$=0.5]{\includegraphics[scale=0.21]{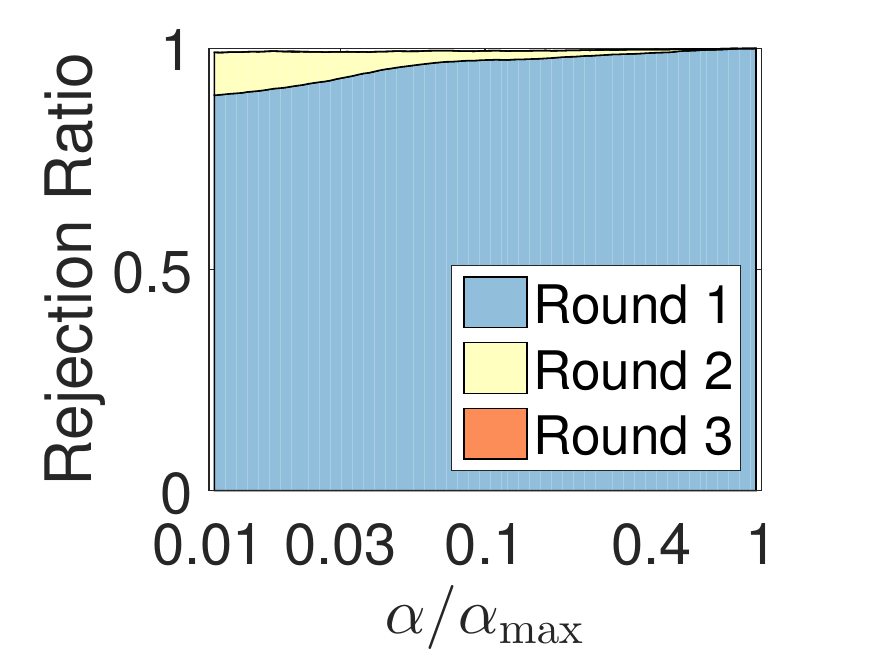}}
		\subfigure[ $\beta/\beta_{\rm{max}}$=0.9]{\includegraphics[scale=0.21]{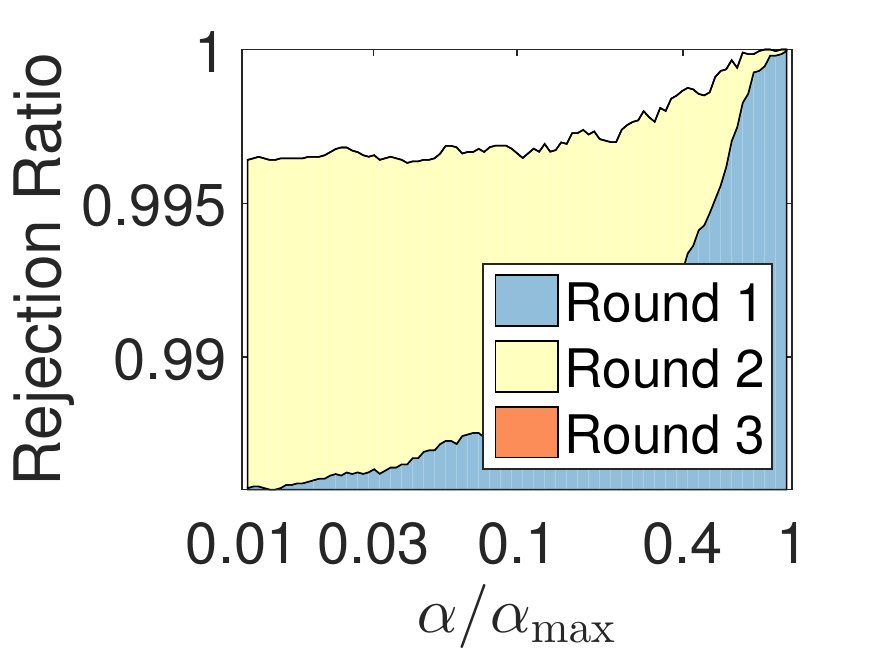}}
		\caption{ Rejection ratios of SIFS on rcv1-train. }
		\label{fig:rejection-ratio-rcv1-train}
	\end{center}
	\vspace*{-20pt}
\end{figure*}

\begin{figure*}[htb!]
	\begin{center}
		\subfigure[ $\beta/\beta_{\rm{max}}$=0.05]{\includegraphics[scale=0.21]{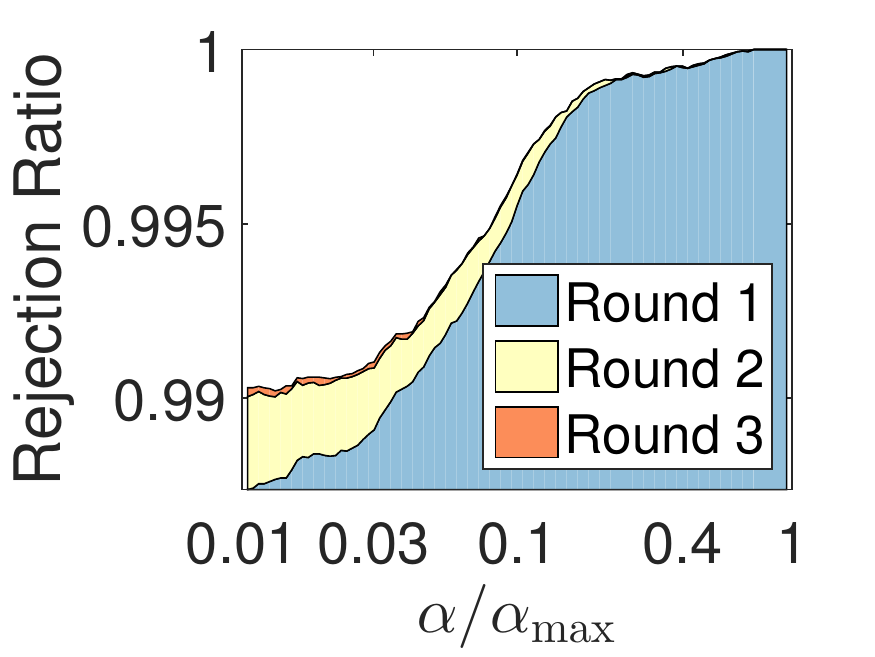}}
		\subfigure[ $\beta/\beta_{\rm{max}}$=0.1]{\includegraphics[scale=0.21]{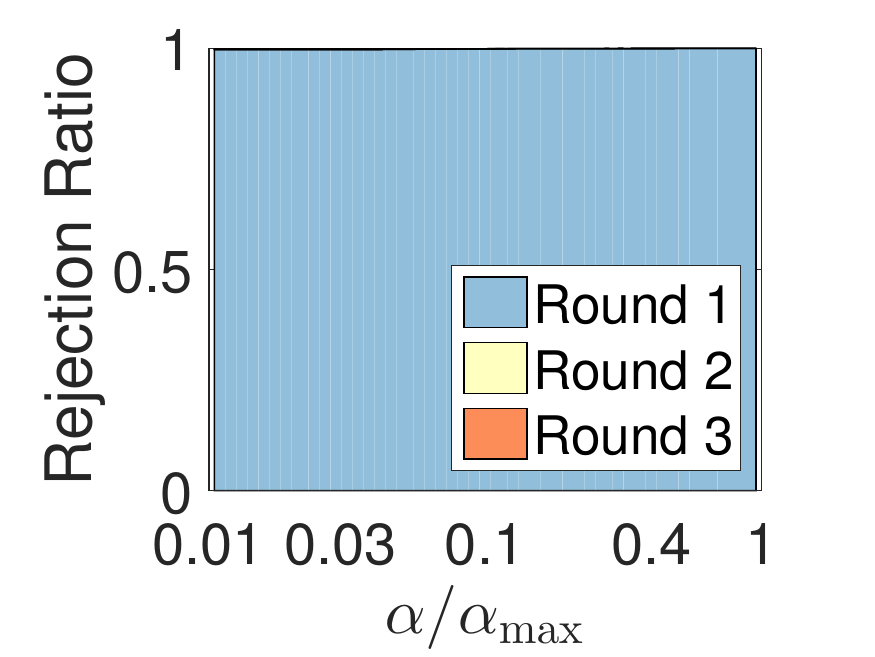}}
		\subfigure[ $\beta/\beta_{\rm{max}}$=0.5]{\includegraphics[scale=0.21]{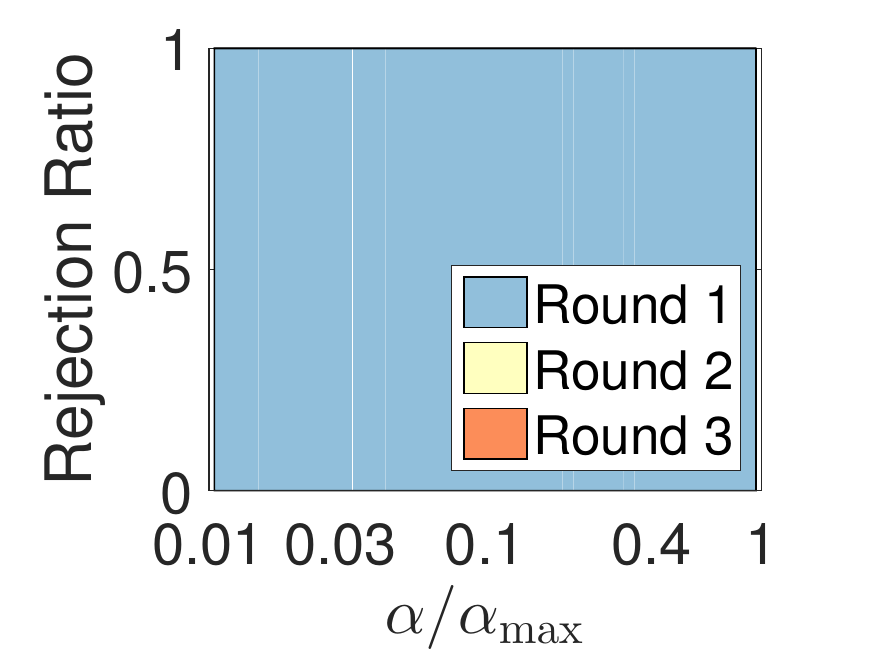}}
		\subfigure[ $\beta/\beta_{\rm{max}}$=0.9]{\includegraphics[scale=0.21]{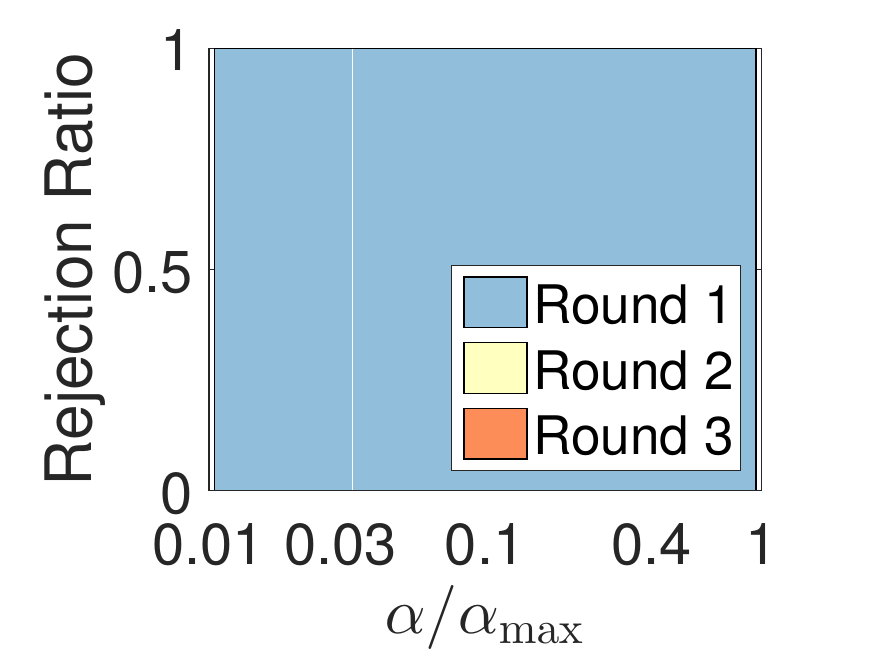}}
		\subfigure[ $\beta/\beta_{\rm{max}}$=0.05]{\includegraphics[scale=0.21]{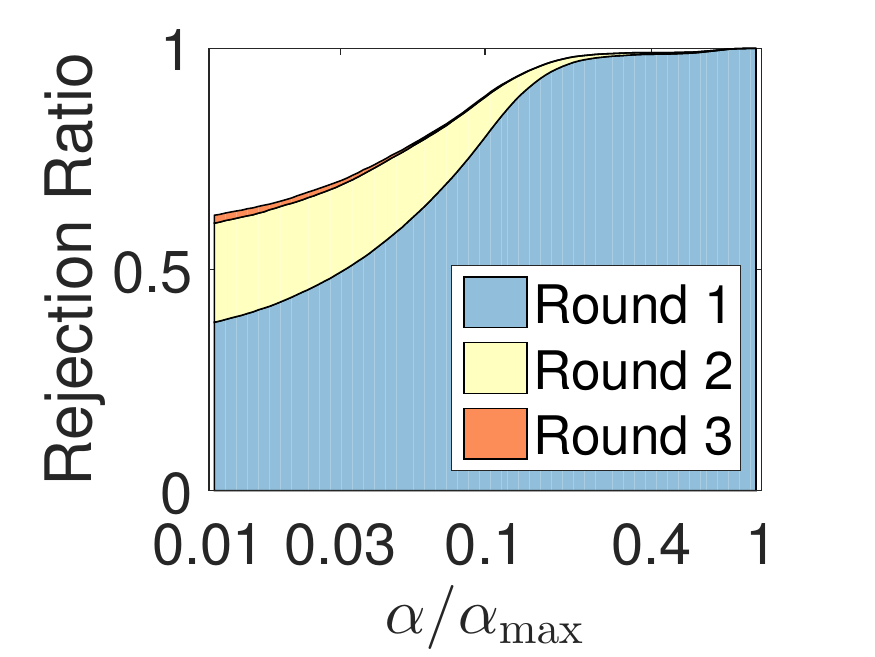}}
		\subfigure[ $\beta/\beta_{\rm{max}}$=0.1]{\includegraphics[scale=0.21]{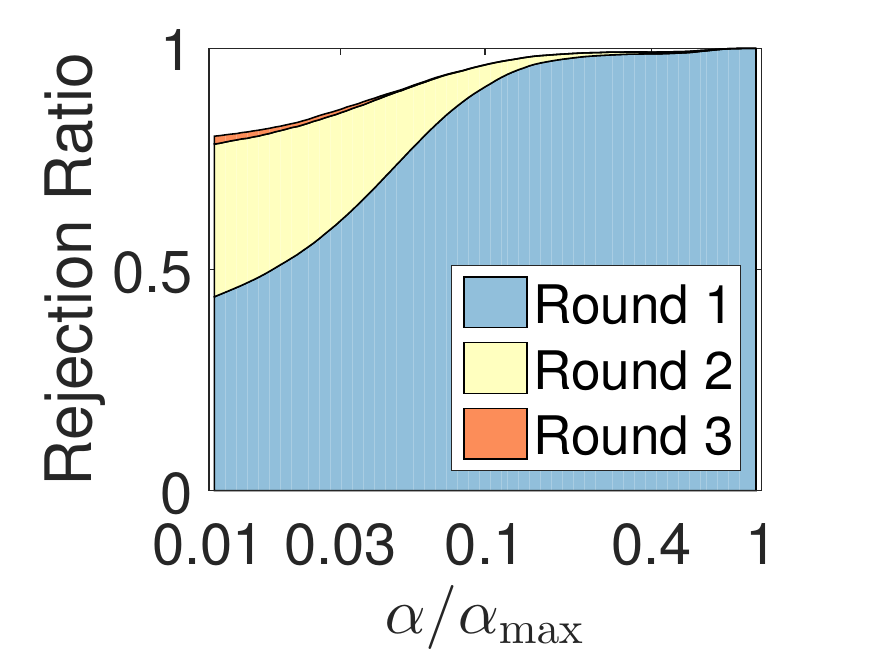}}
		\subfigure[ $\beta/\beta_{\rm{max}}$=0.5]{\includegraphics[scale=0.21]{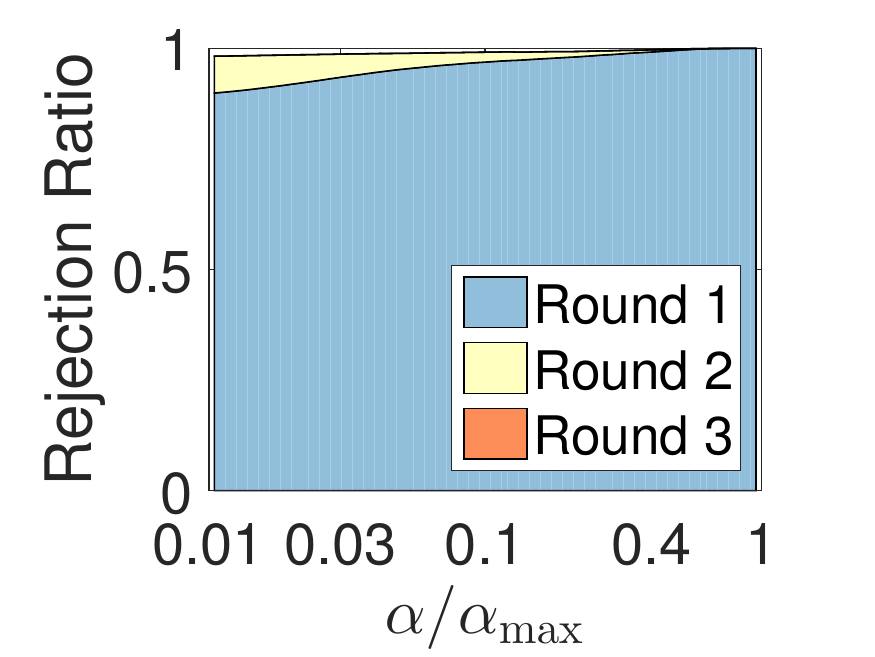}}
		\subfigure[ $\beta/\beta_{\rm{max}}$=0.9]{\includegraphics[scale=0.21]{images/rcv1_train_binary09_RL_rej_ratio}}
		\caption{ Rejection ratios of SIFS on rcv1-test. }
		\label{fig:rejection-ratio-rcv1-test}
	\end{center}
	\vspace*{-20pt}
\end{figure*}

\begin{figure*}[htb!]
	\begin{center}
		\subfigure[ $\beta/\beta_{\rm{max}}$=0.05]{\includegraphics[scale=0.21]{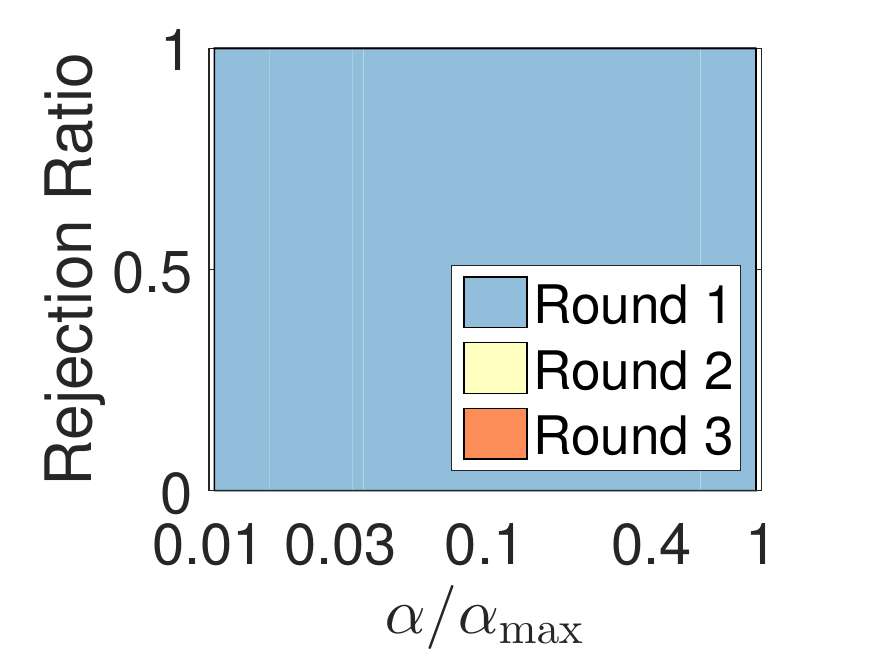}}
		\subfigure[ $\beta/\beta_{\rm{max}}$=0.1]{\includegraphics[scale=0.21]{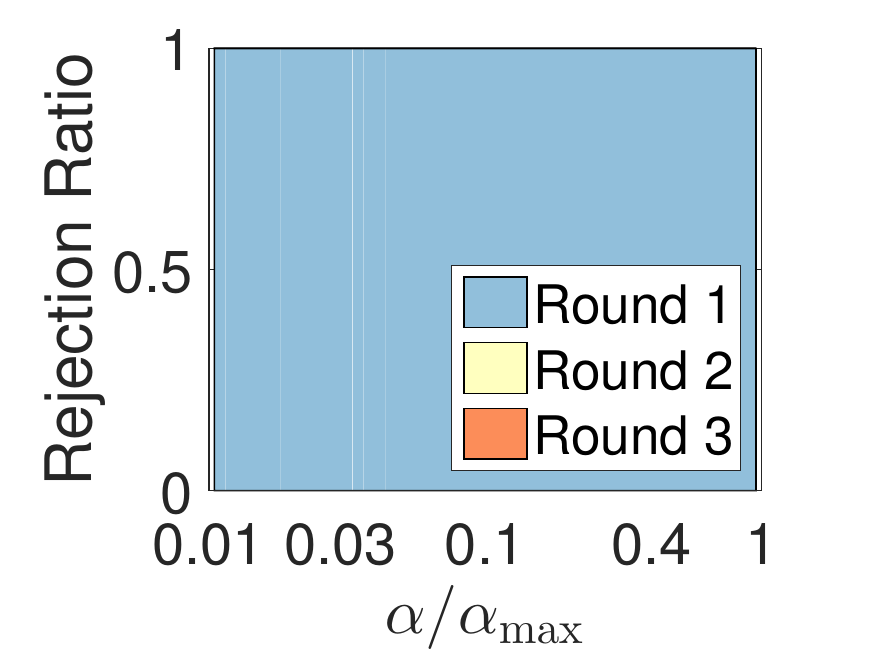}}
		\subfigure[ $\beta/\beta_{\rm{max}}$=0.5]{\includegraphics[scale=0.21]{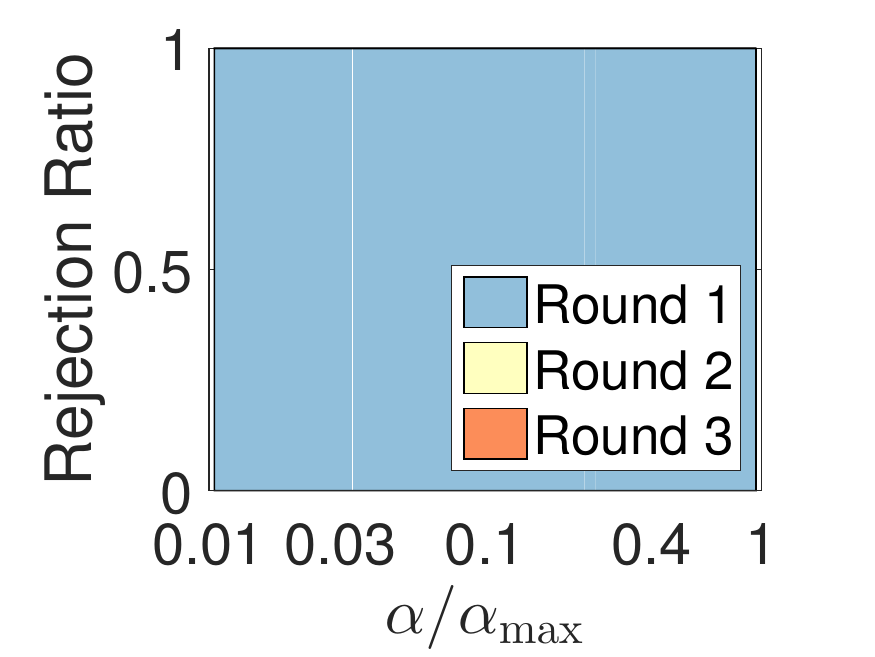}}
		\subfigure[ $\beta/\beta_{\rm{max}}$=0.9]{\includegraphics[scale=0.21]{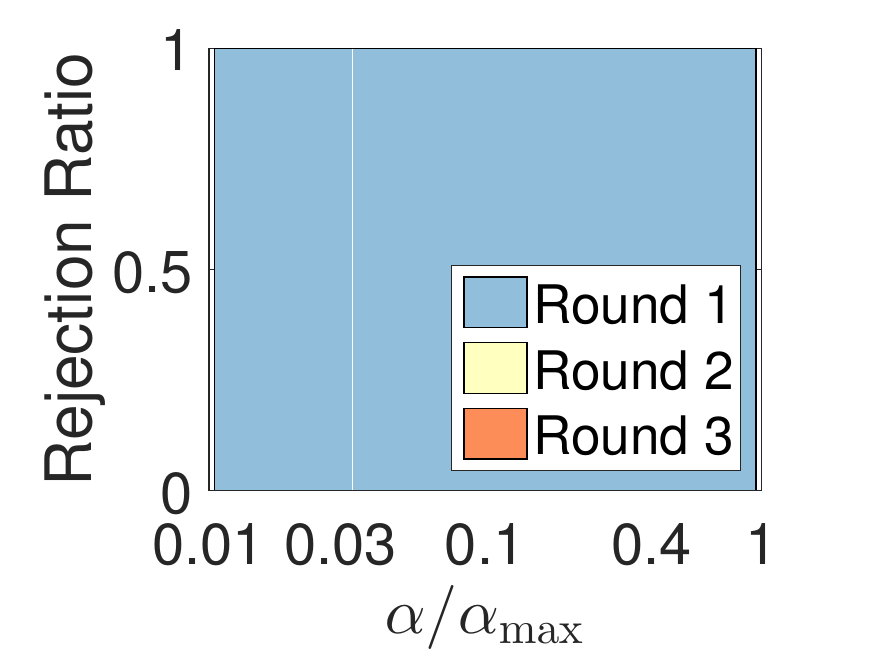}}
		\subfigure[ $\beta/\beta_{\rm{max}}$=0.05]{\includegraphics[scale=0.21]{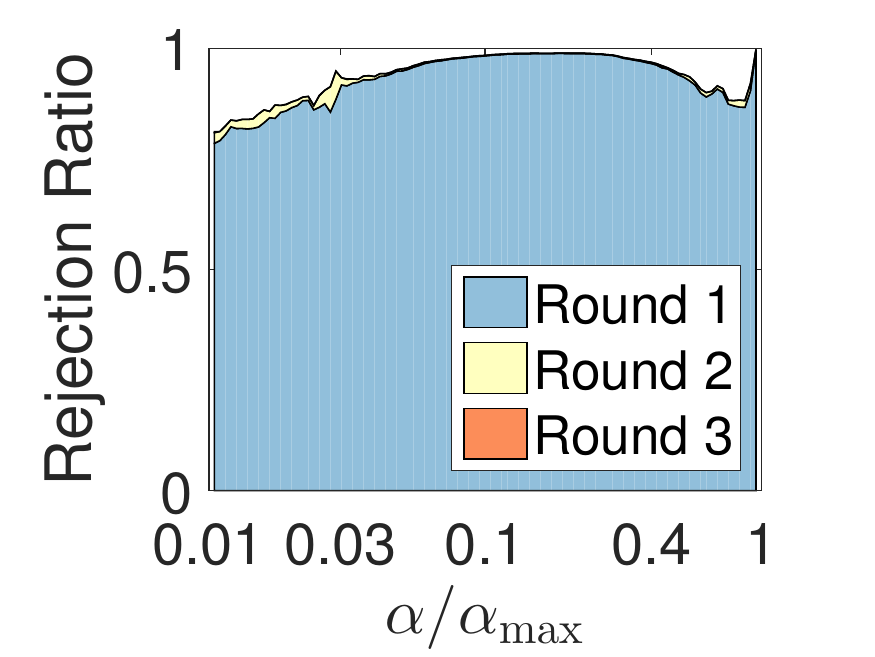}}
		\subfigure[ $\beta/\beta_{\rm{max}}$=0.1]{\includegraphics[scale=0.21]{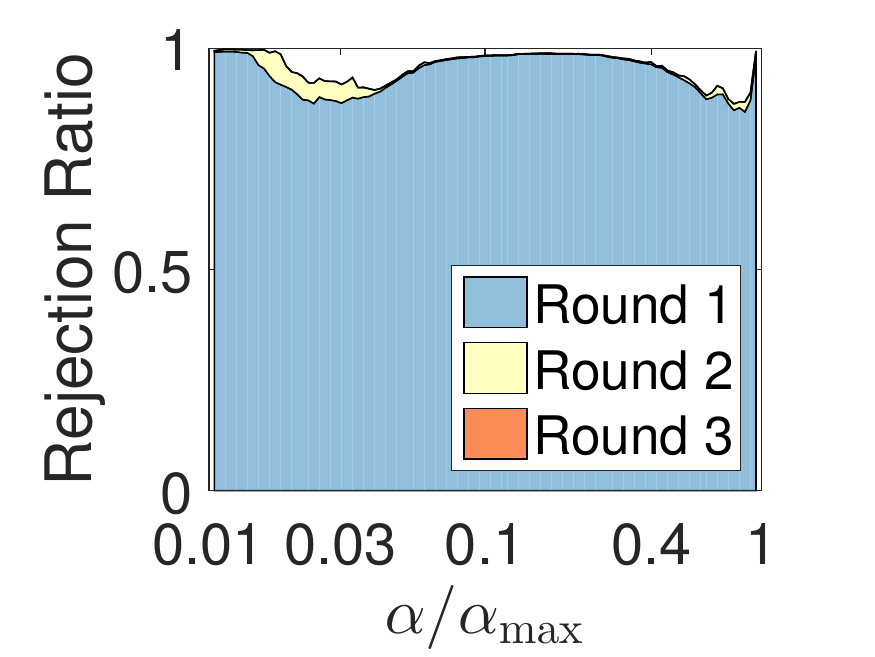}}
		\subfigure[ $\beta/\beta_{\rm{max}}$=0.5]{\includegraphics[scale=0.21]{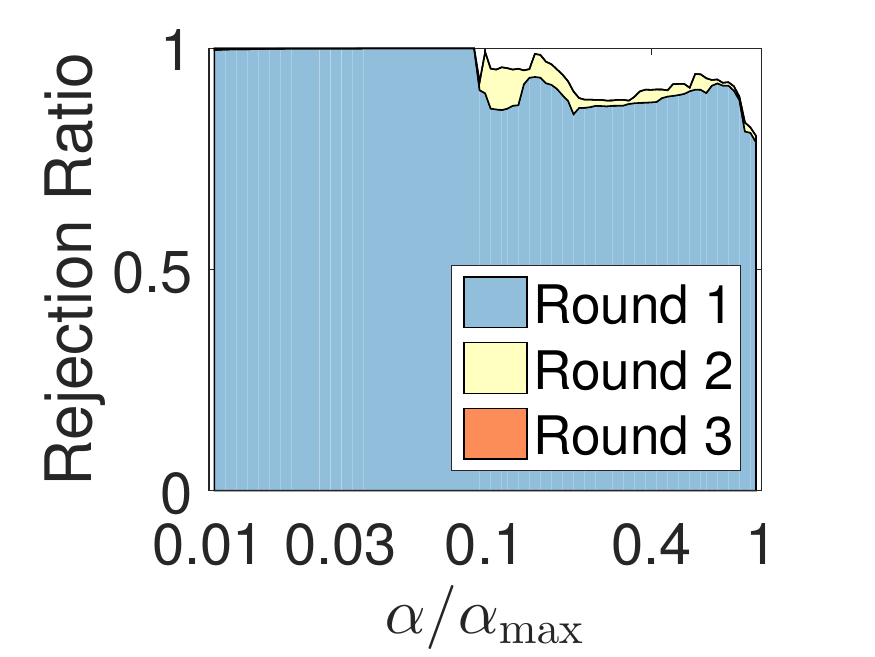}}
		\subfigure[ $\beta/\beta_{\rm{max}}$=0.9]{\includegraphics[scale=0.21]{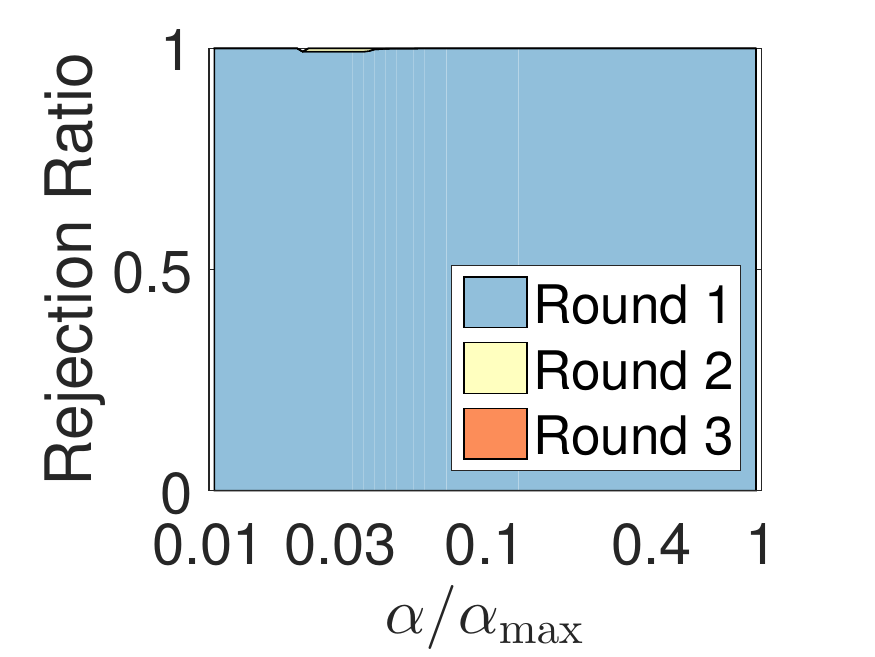}}
		\caption{ Rejection ratios of SIFS on url. }
		\label{fig:rejection-ratio-url}
	\end{center}
	\vspace*{-20pt}
\end{figure*}

\begin{figure*}[htb!]
	\begin{center}
		\subfigure[ $\beta/\beta_{\rm{max}}$=0.05]{\includegraphics[scale=0.21]{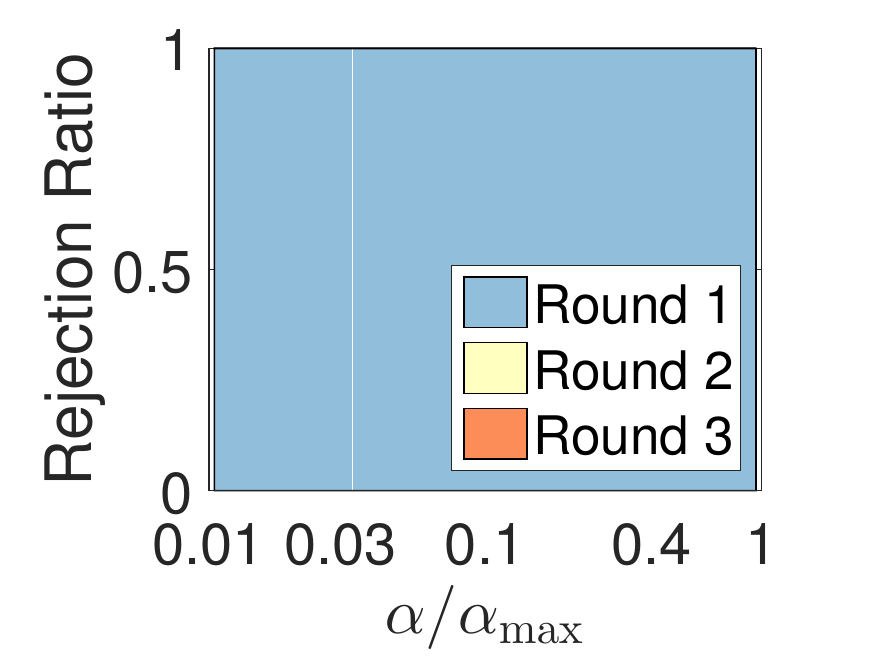}}
		\subfigure[ $\beta/\beta_{\rm{max}}$=0.1]{\includegraphics[scale=0.21]{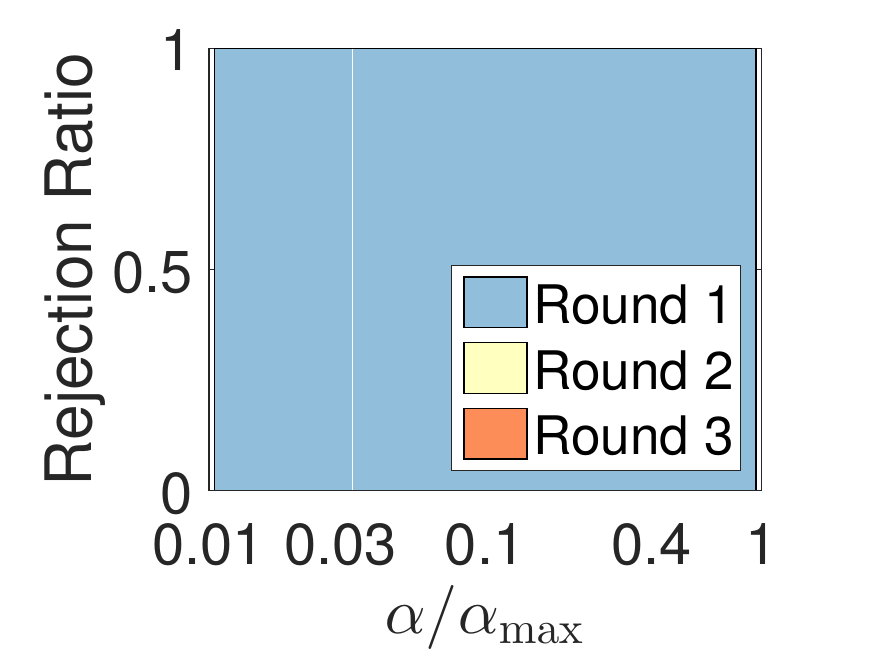}}
		\subfigure[ $\beta/\beta_{\rm{max}}$=0.5]{\includegraphics[scale=0.21]{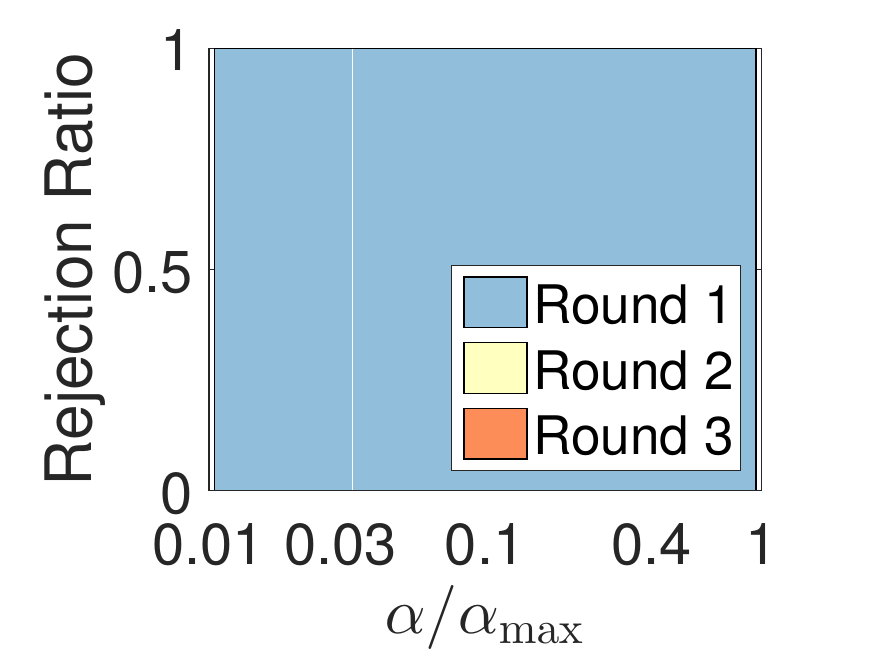}}
		\subfigure[ $\beta/\beta_{\rm{max}}$=0.9]{\includegraphics[scale=0.21]{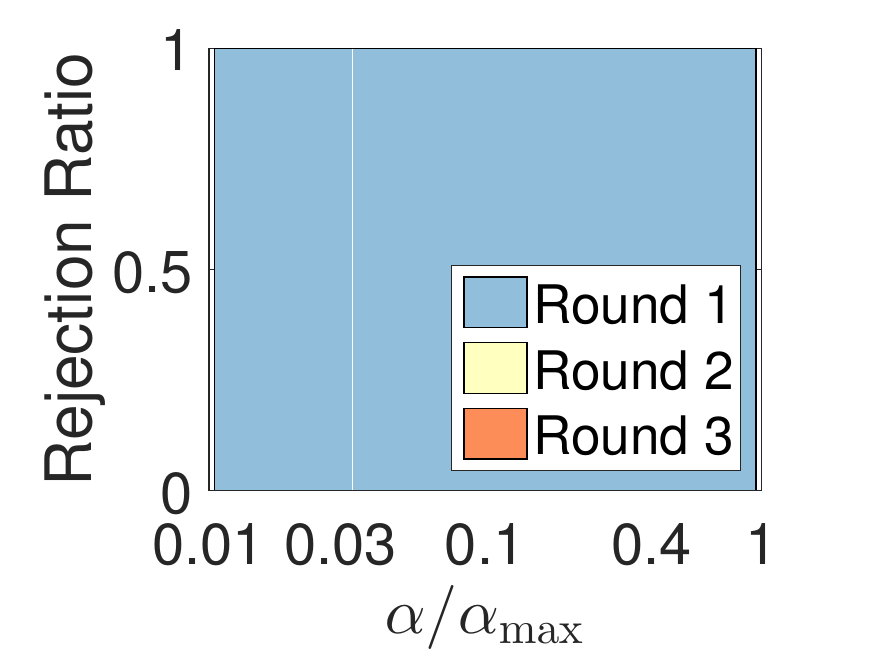}}
		\subfigure[ $\beta/\beta_{\rm{max}}$=0.05]{\includegraphics[scale=0.21]{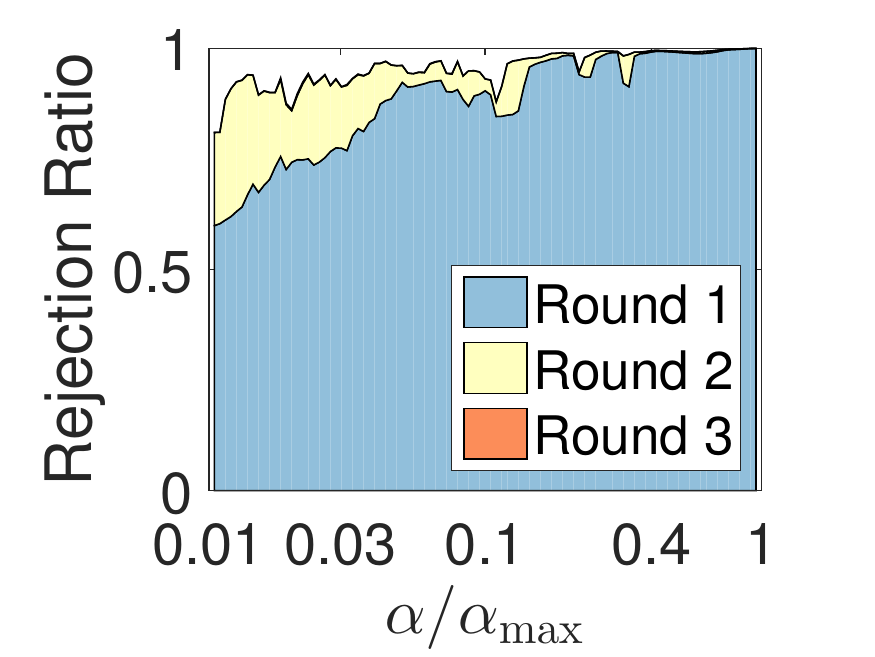}}
		\subfigure[ $\beta/\beta_{\rm{max}}$=0.1]{\includegraphics[scale=0.21]{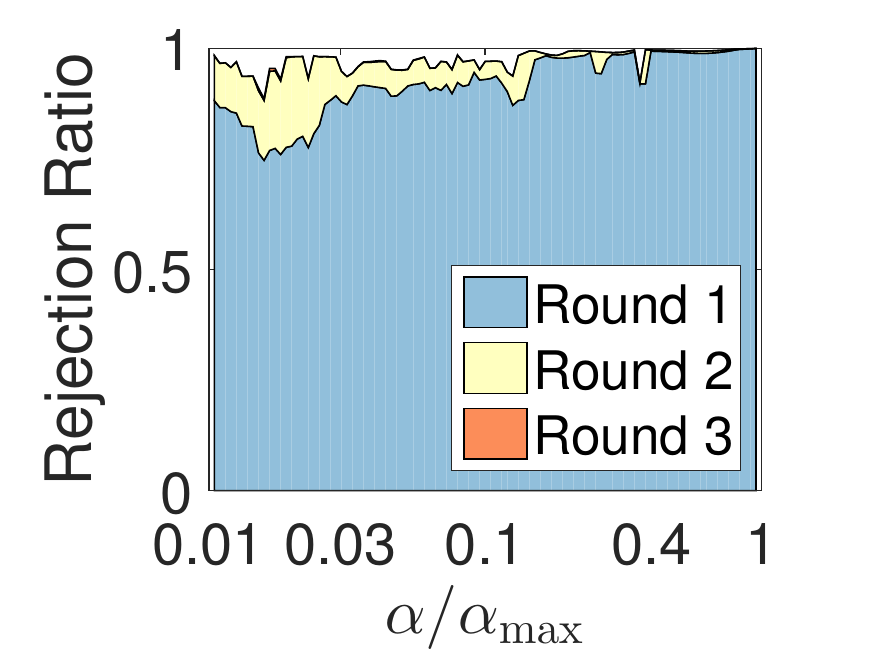}}
		\subfigure[ $\beta/\beta_{\rm{max}}$=0.5]{\includegraphics[scale=0.21]{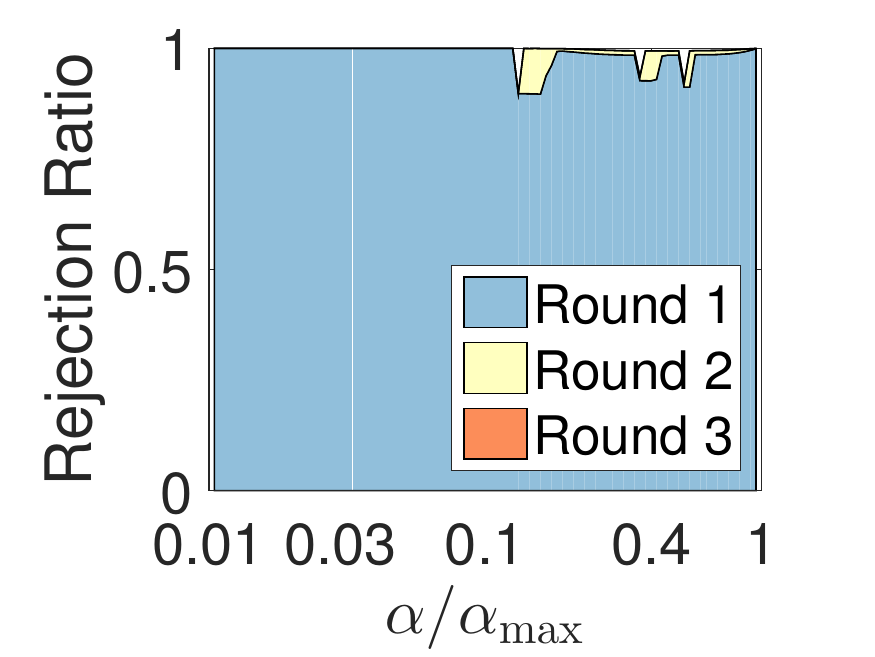}}
		\subfigure[ $\beta/\beta_{\rm{max}}$=0.9]{\includegraphics[scale=0.21]{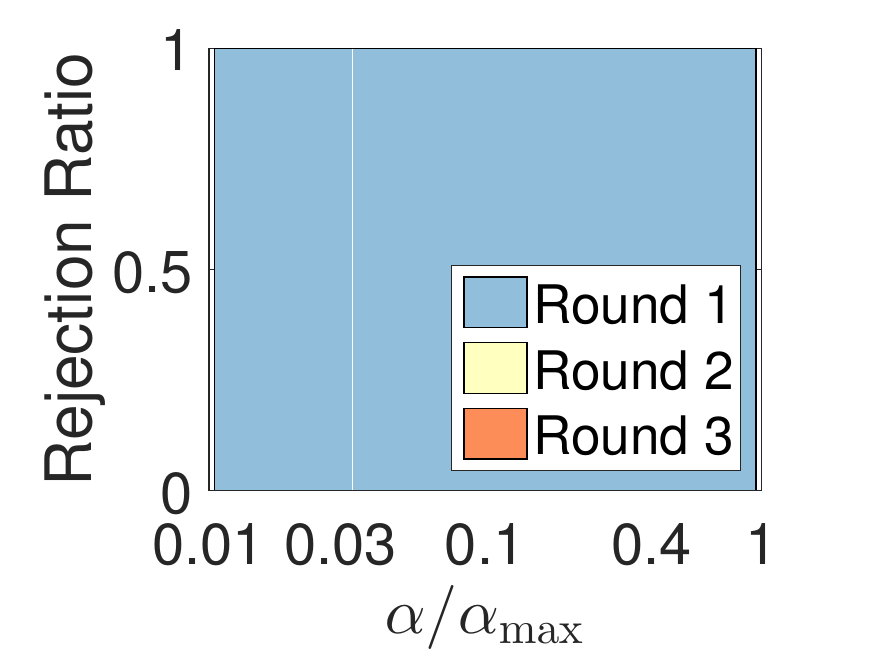}}
		\caption{Rejection ratios of SIFS on kddb. }
		\label{fig:rejection-ratio-kddb}
	\end{center}
	\vspace*{-20pt}
\end{figure*}

\begin{figure*}[htb!]
	\begin{center}
		\subfigure[ $\beta/\beta_{\rm{max}}$=0.05]{\includegraphics[scale=0.21]{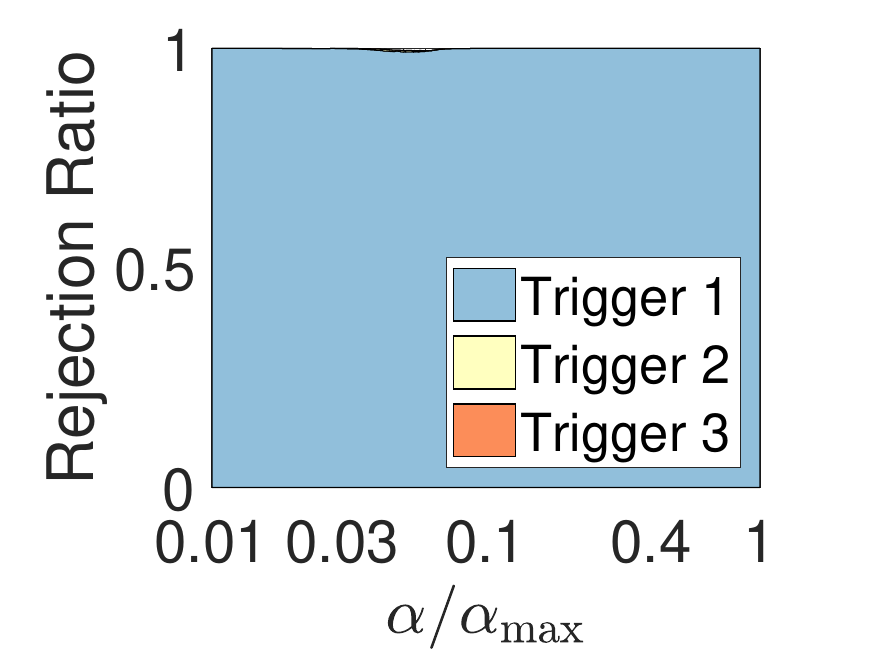}}
		\subfigure[ $\beta/\beta_{\rm{max}}$=0.1]{\includegraphics[scale=0.21]{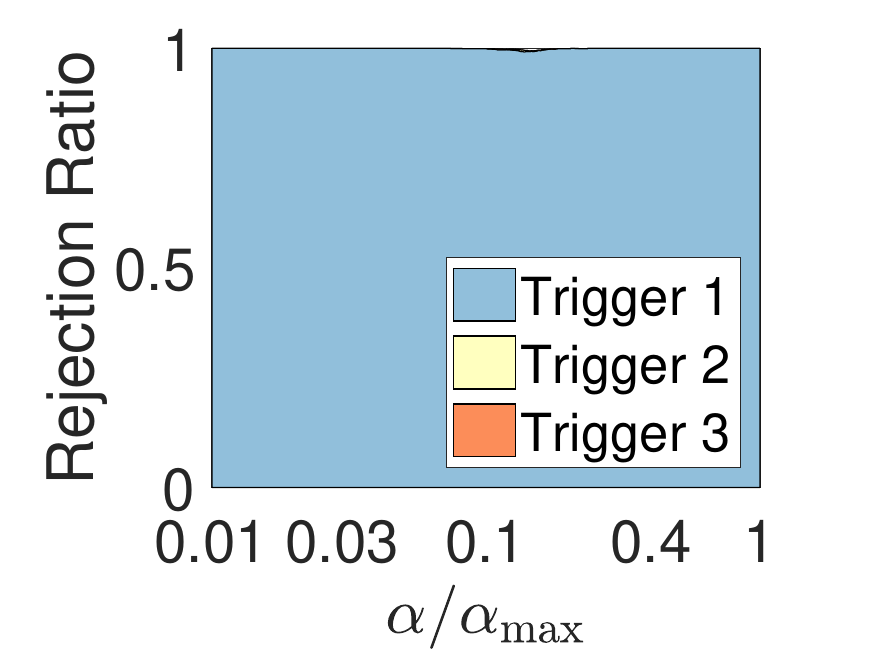}}
		\subfigure[ $\beta/\beta_{\rm{max}}$=0.5]{\includegraphics[scale=0.21]{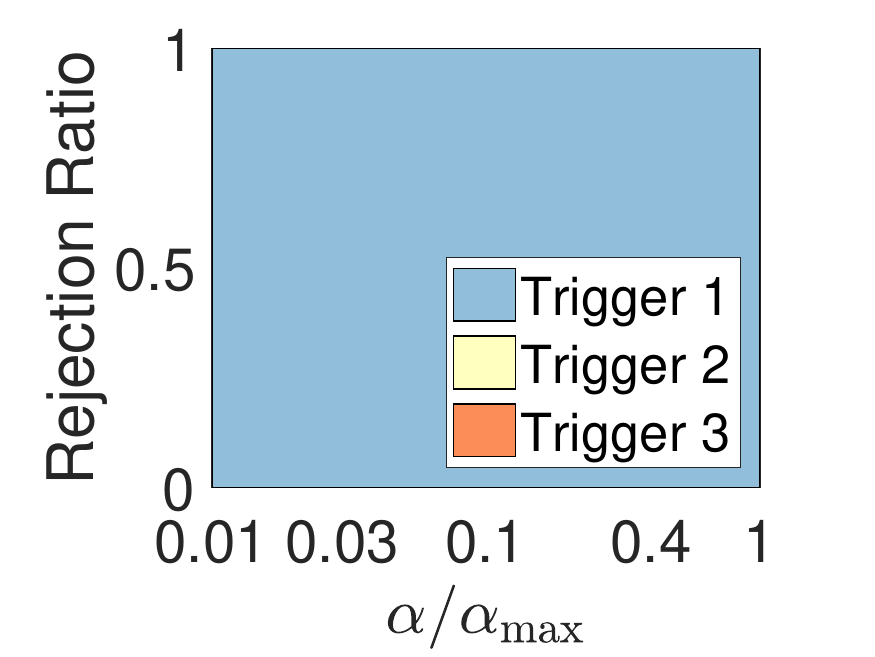}}
		\subfigure[ $\beta/\beta_{\rm{max}}$=0.9]{\includegraphics[scale=0.21]{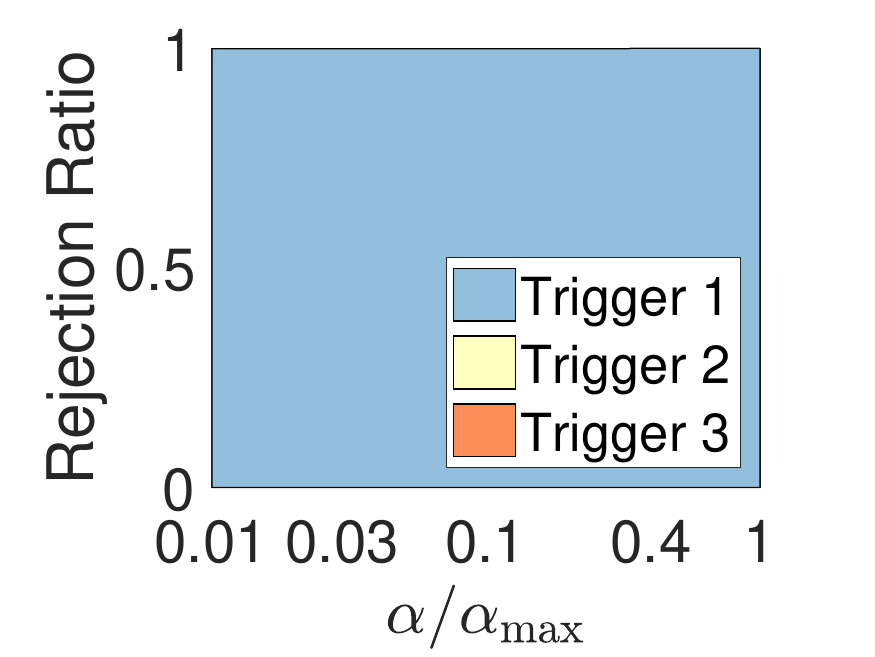}}
		\subfigure[ $\beta/\beta_{\rm{max}}$=0.05]{\includegraphics[scale=0.21]{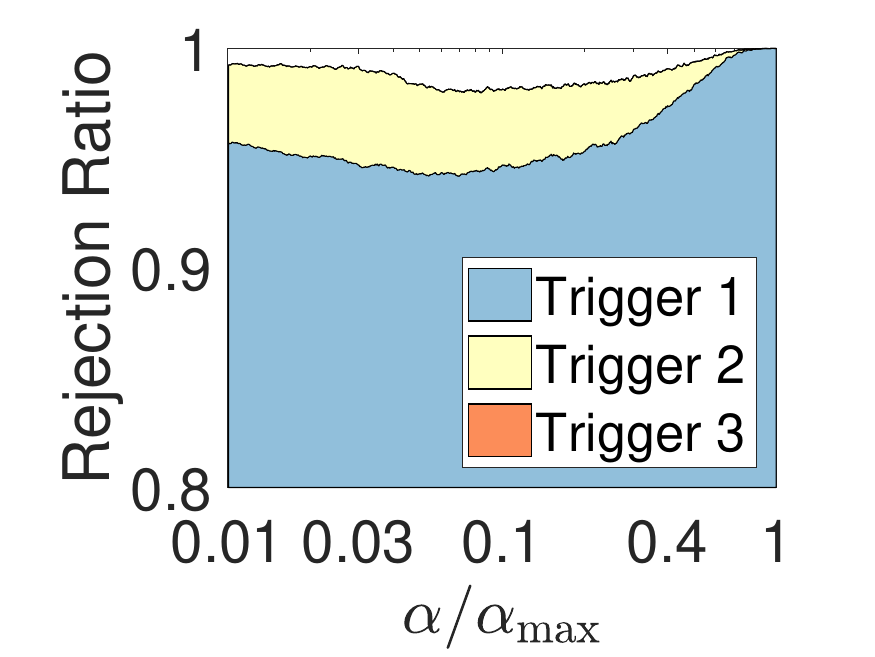}}
		\subfigure[ $\beta/\beta_{\rm{max}}$=0.1]{\includegraphics[scale=0.21]{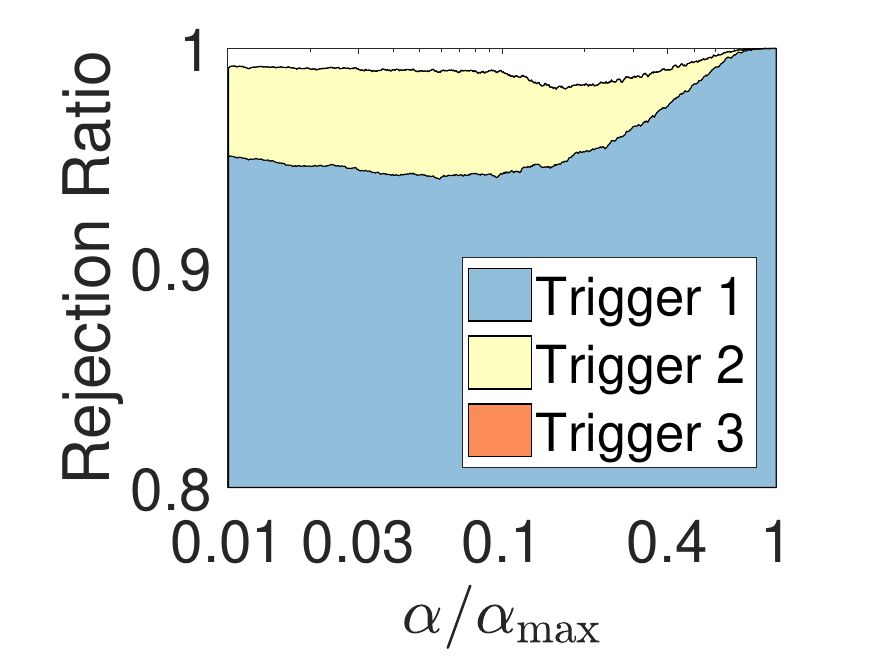}}
		\subfigure[ $\beta/\beta_{\rm{max}}$=0.5]{\includegraphics[scale=0.21]{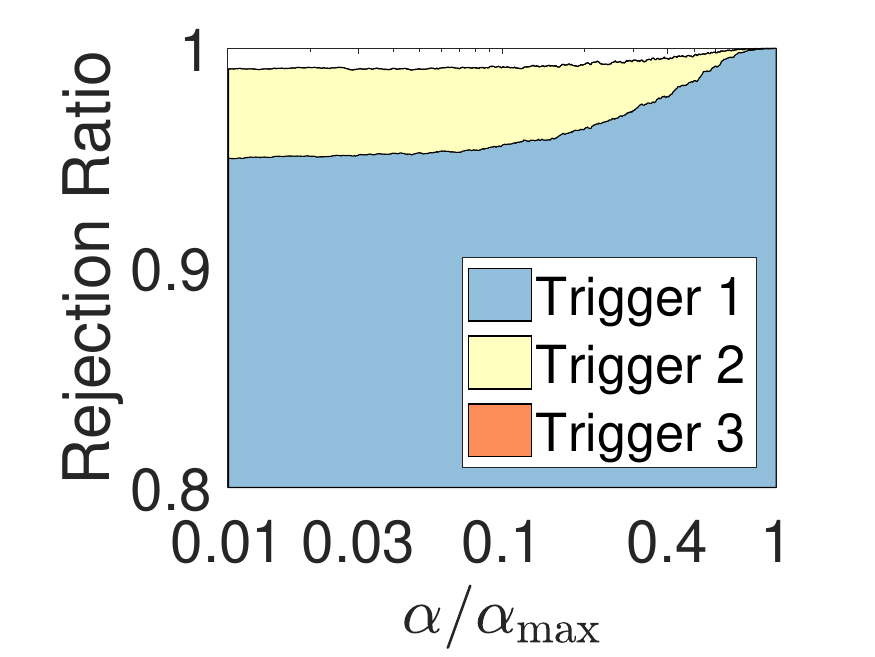}}
		\subfigure[ $\beta/\beta_{\rm{max}}$=0.9]{\includegraphics[scale=0.21]{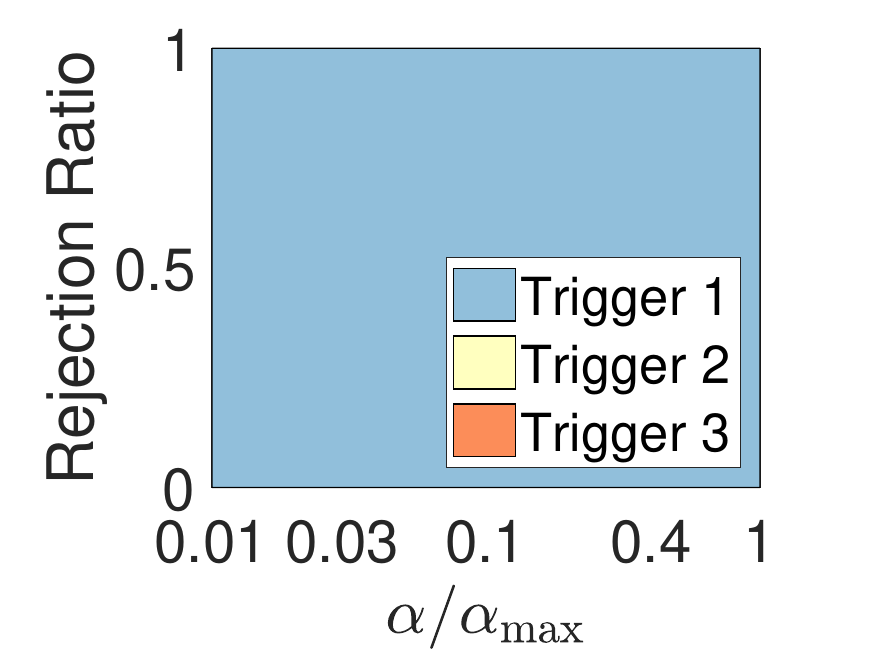}}
		\caption{Rejection ratios of SIFS on syn-multi1. }
		\label{fig:rejection-ratio-syn-multi1}
	\end{center}
	\vspace*{-20pt}
\end{figure*}

\begin{figure*}[htb!]
	\begin{center}
		\subfigure[ $\beta/\beta_{\rm{max}}$=0.05]{\includegraphics[scale=0.21]{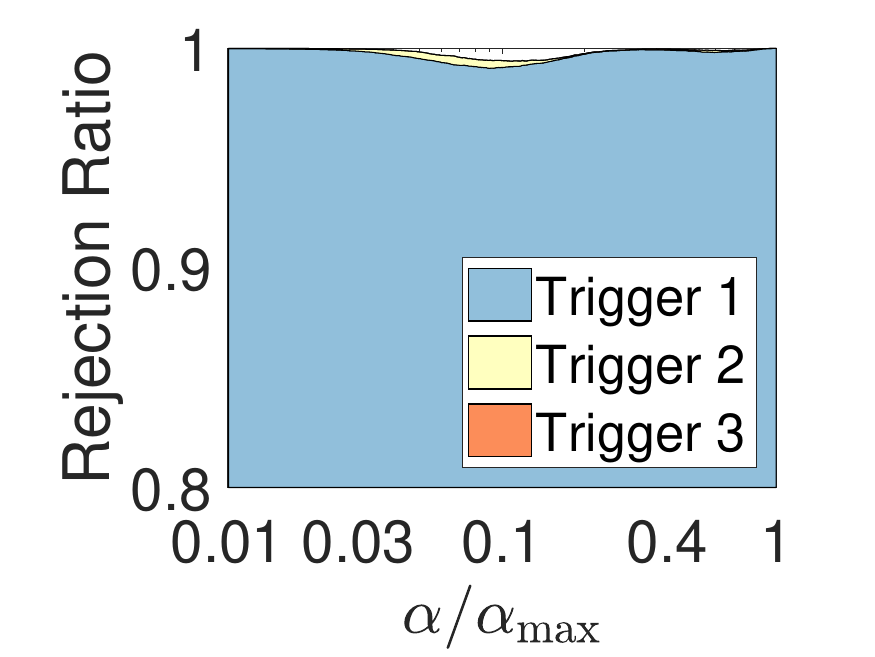}}
		\subfigure[ $\beta/\beta_{\rm{max}}$=0.1]{\includegraphics[scale=0.21]{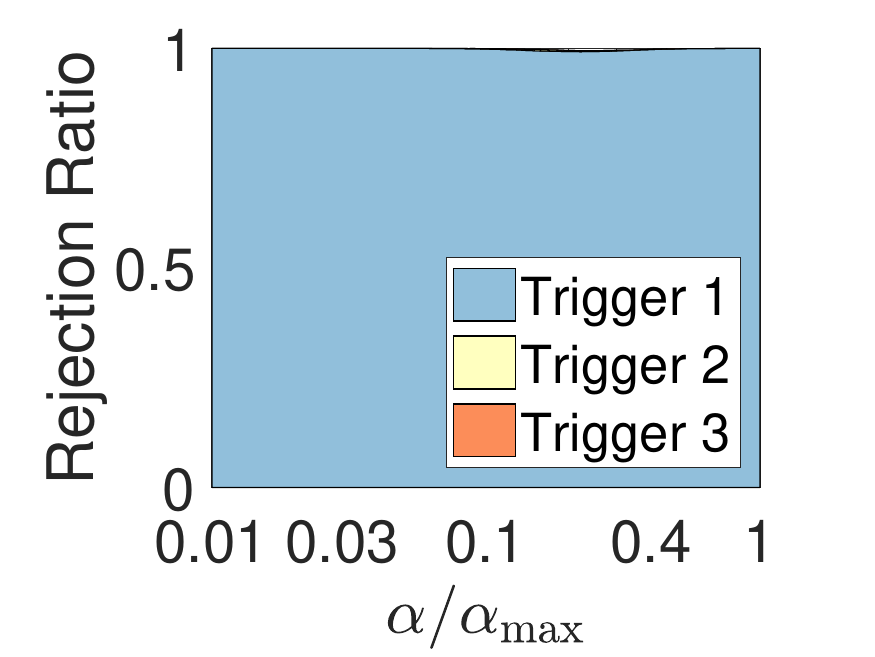}}
		\subfigure[ $\beta/\beta_{\rm{max}}$=0.5]{\includegraphics[scale=0.21]{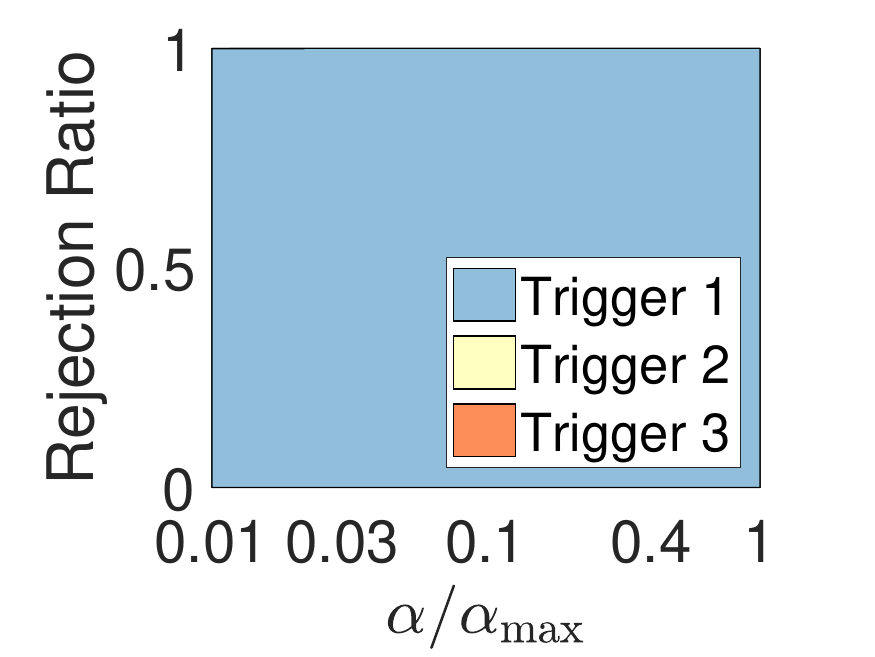}}
		\subfigure[ $\beta/\beta_{\rm{max}}$=0.9]{\includegraphics[scale=0.21]{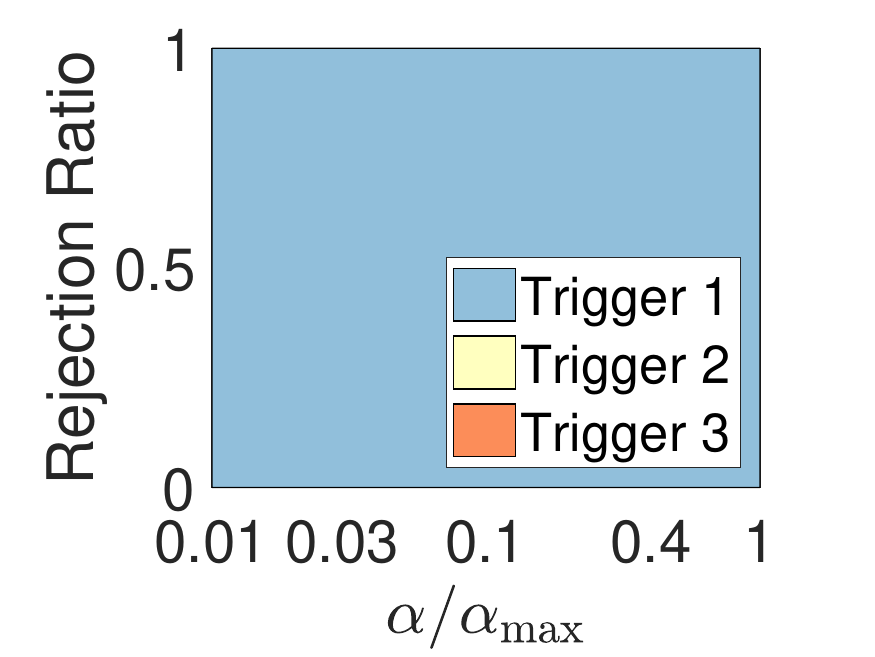}}
		\subfigure[ $\beta/\beta_{\rm{max}}$=0.05]{\includegraphics[scale=0.21]{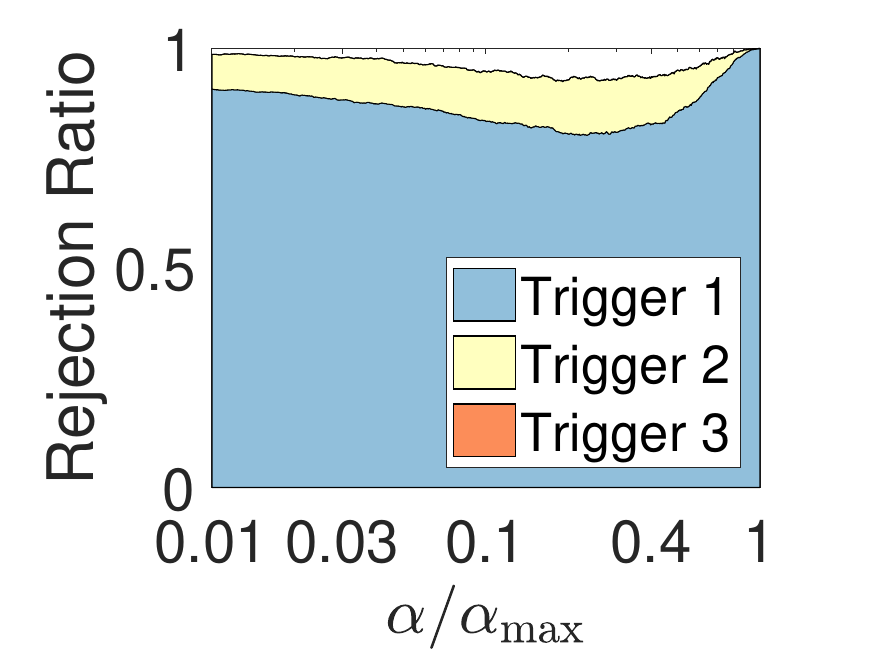}}
		\subfigure[ $\beta/\beta_{\rm{max}}$=0.1]{\includegraphics[scale=0.21]{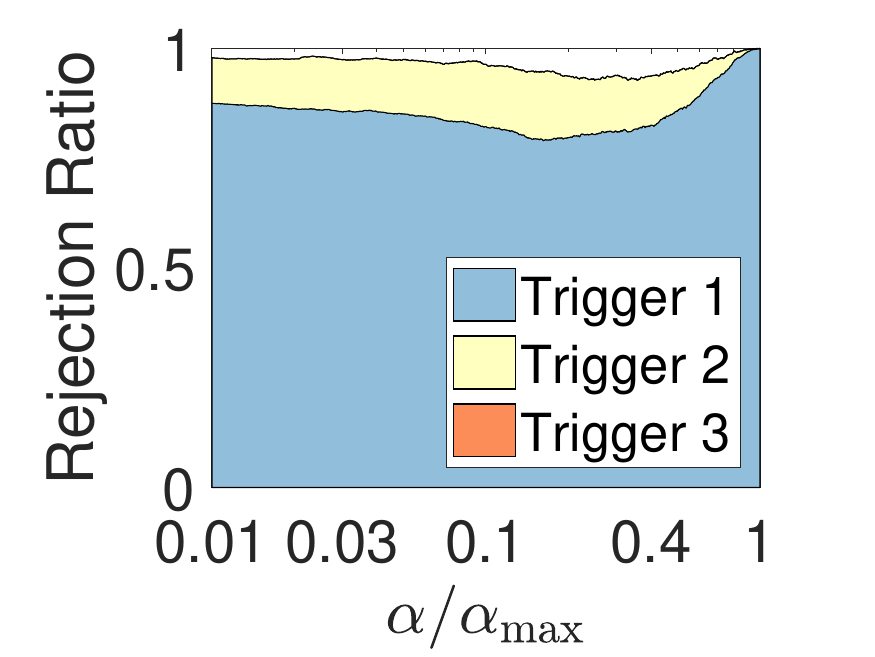}}
		\subfigure[ $\beta/\beta_{\rm{max}}$=0.5]{\includegraphics[scale=0.21]{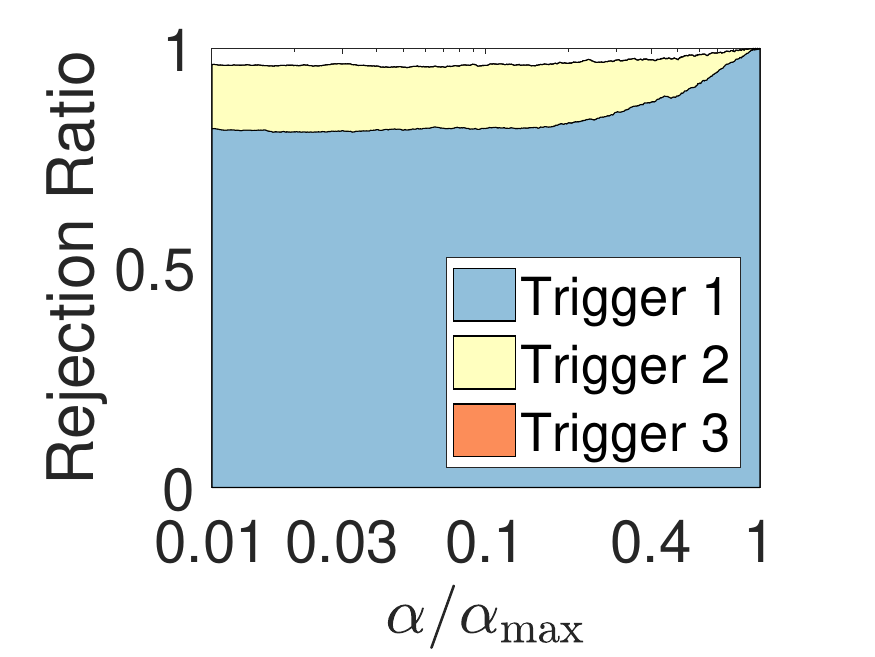}}
		\subfigure[ $\beta/\beta_{\rm{max}}$=0.9]{\includegraphics[scale=0.21]{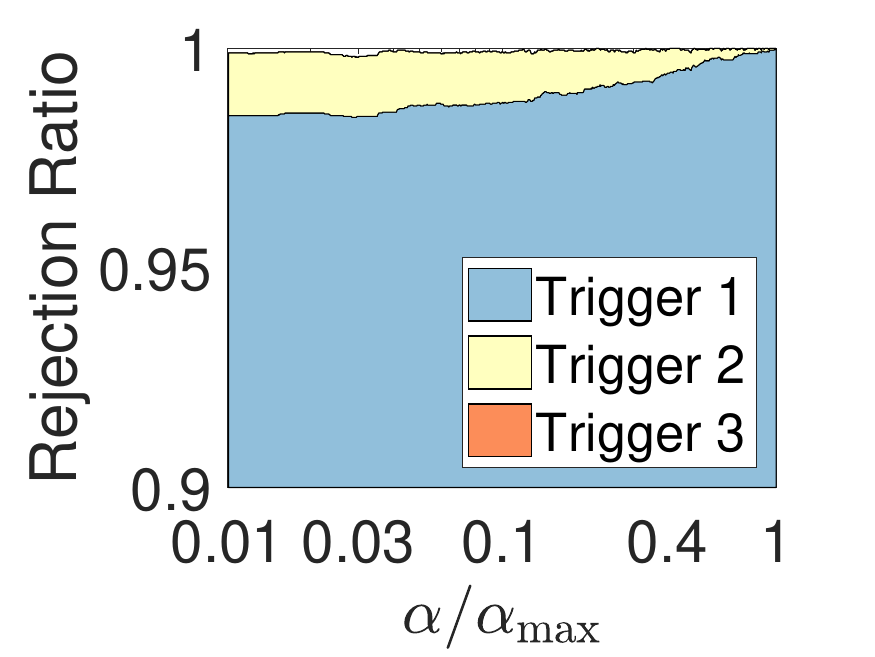}}
		\caption{Rejection ratios of SIFS on syn-multi3. }
		\label{fig:rejection-ratio-syn-multi3}
	\end{center}
	\vspace*{-20pt}
\end{figure*}

\begin{figure*}[htb!]
	\begin{center}
		\subfigure[ $\beta/\beta_{\rm{max}}$=0.05]{\includegraphics[scale=0.21]{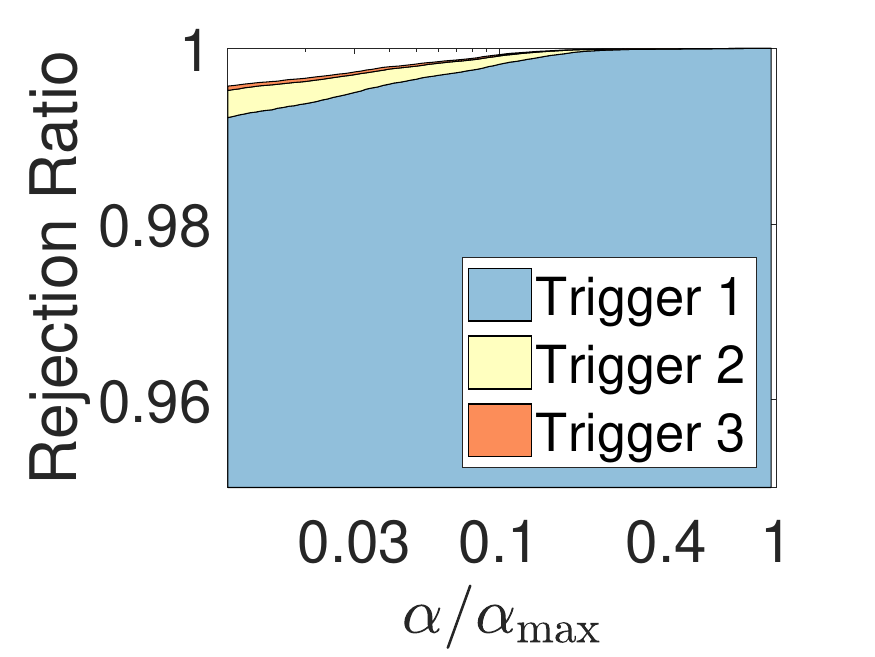}}
		\subfigure[ $\beta/\beta_{\rm{max}}$=0.1]{\includegraphics[scale=0.21]{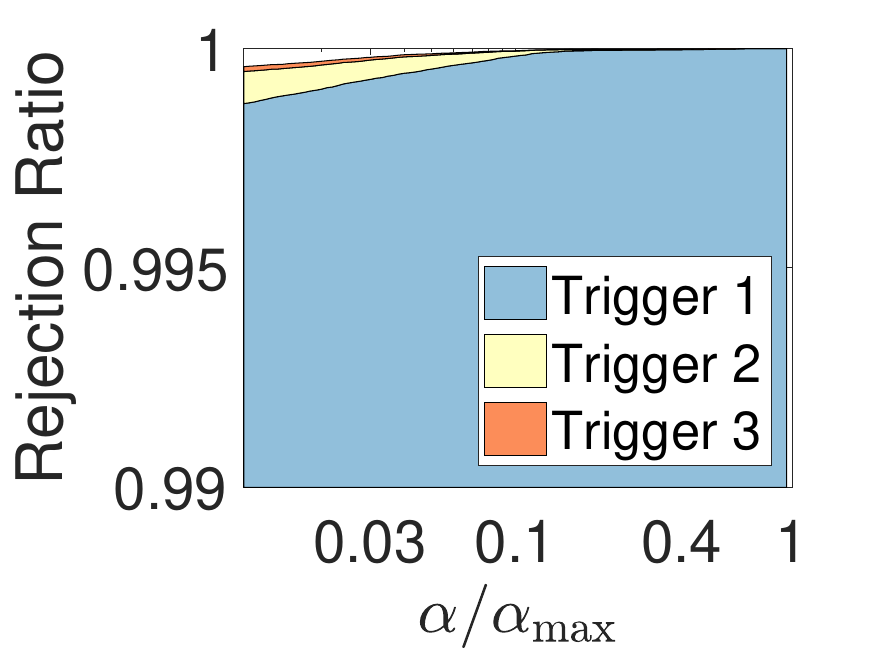}}
		\subfigure[ $\beta/\beta_{\rm{max}}$=0.5]{\includegraphics[scale=0.21]{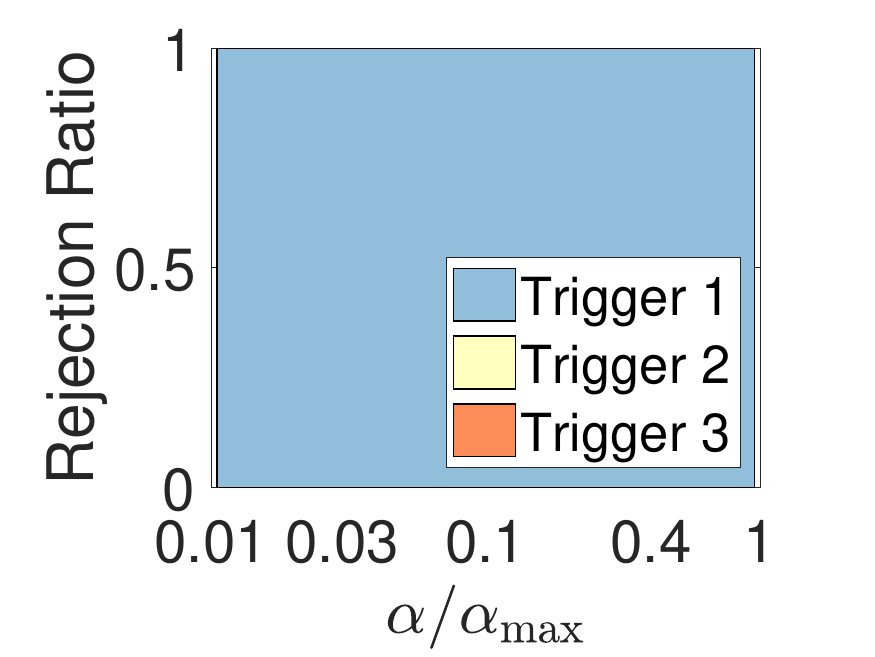}}
		\subfigure[ $\beta/\beta_{\rm{max}}$=0.9]{\includegraphics[scale=0.21]{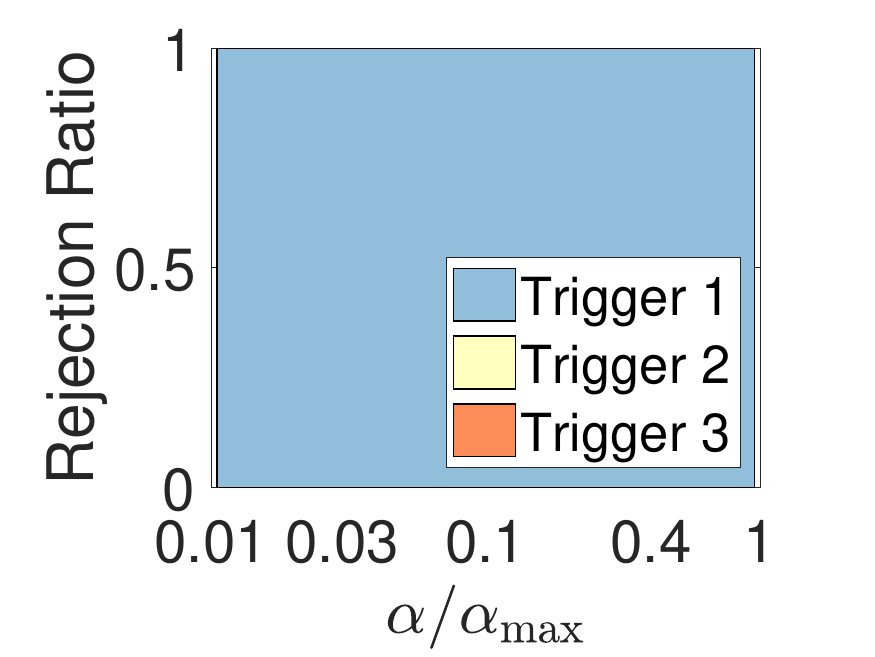}}
		\subfigure[ $\beta/\beta_{\rm{max}}$=0.05]{\includegraphics[scale=0.21]{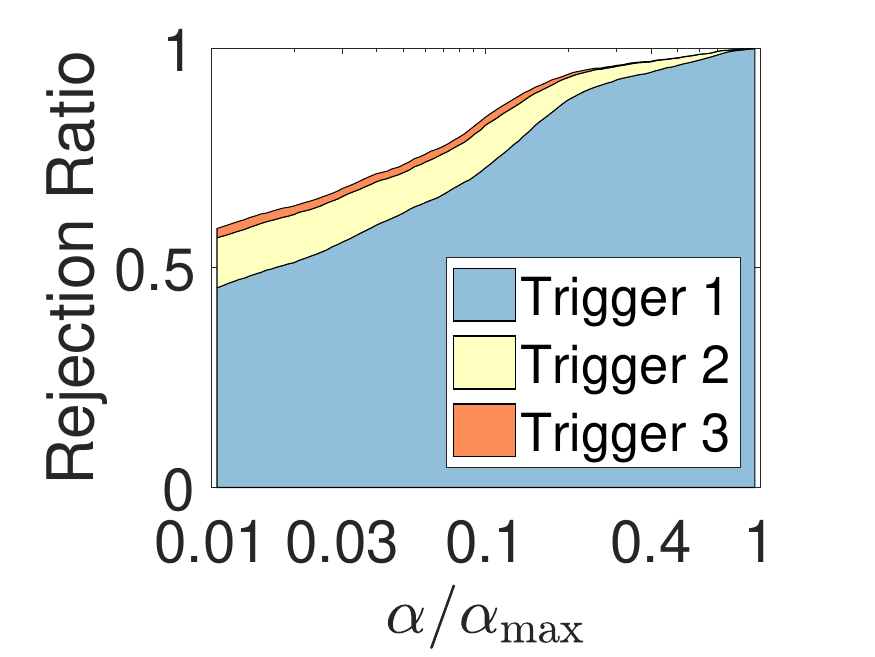}}
		\subfigure[ $\beta/\beta_{\rm{max}}$=0.1]{\includegraphics[scale=0.21]{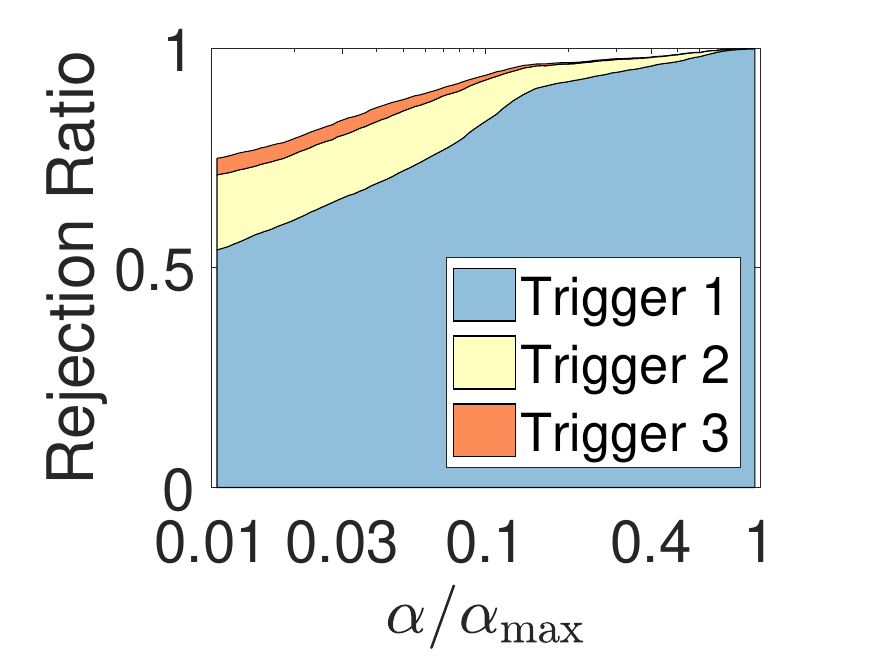}}
		\subfigure[ $\beta/\beta_{\rm{max}}$=0.5]{\includegraphics[scale=0.21]{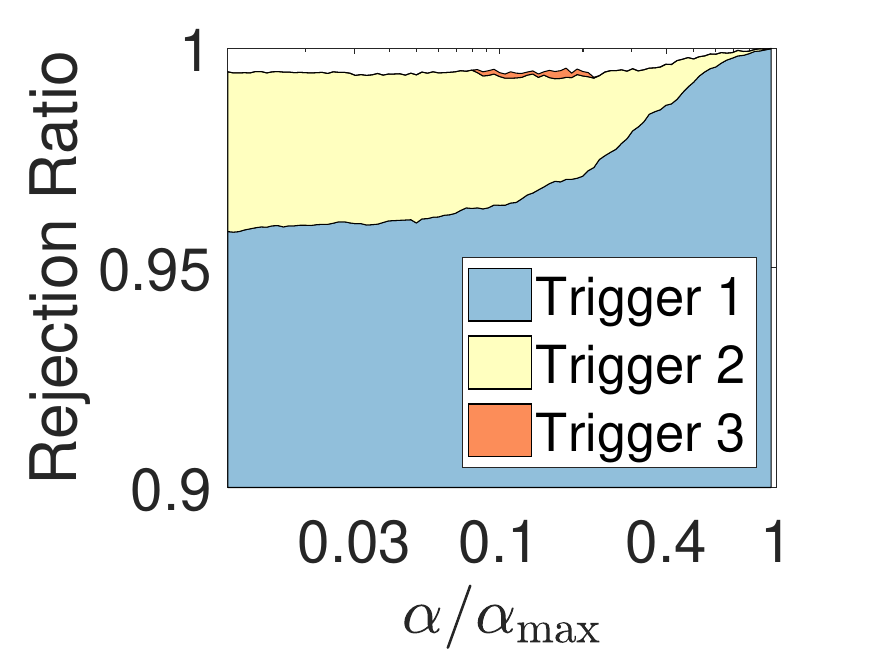}}
		\subfigure[ $\beta/\beta_{\rm{max}}$=0.9]{\includegraphics[scale=0.21]{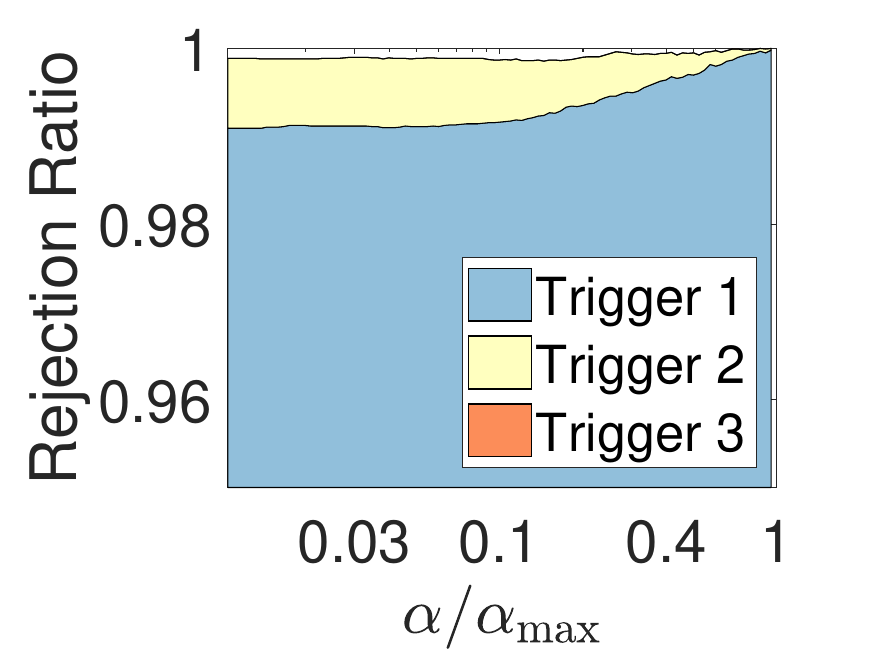}}
		\caption{Rejection ratios of SIFS on rcv1-multiclass. }
		\label{fig:rejection-ratio-rcv1-multiclass}
	\end{center}
	\vspace*{-20pt}
\end{figure*}

Below, we report the rejection ratios of SIFS on syn1 (\figref{fig:rejection-ratio-syn-1}), syn3 (\figref{fig:rejection-ratio-syn-3}), rcv1-train (\figref{fig:rejection-ratio-rcv1-train}), rcv1-test(\figref{fig:rejection-ratio-rcv1-test}), url (\figref{fig:rejection-ratio-url}), kddb (\figref{fig:rejection-ratio-kddb}), 	syn-multi1 (\figref{fig:rejection-ratio-syn-multi1}), syn-multi3 (\figref{fig:rejection-ratio-syn-multi3}), and rcv1-multiclass (\figref{fig:rejection-ratio-rcv1-multiclass}), which are omitted in the main paper. In all the figures, the first row presents the rejection ratios of feature screening and the second row presents those of sample screening.
\begin{figure*}[htb!]
	\begin{center}
		\subfigure[ $\beta/\beta_{\rm{max}}$=0.05]{\includegraphics[scale=0.21]{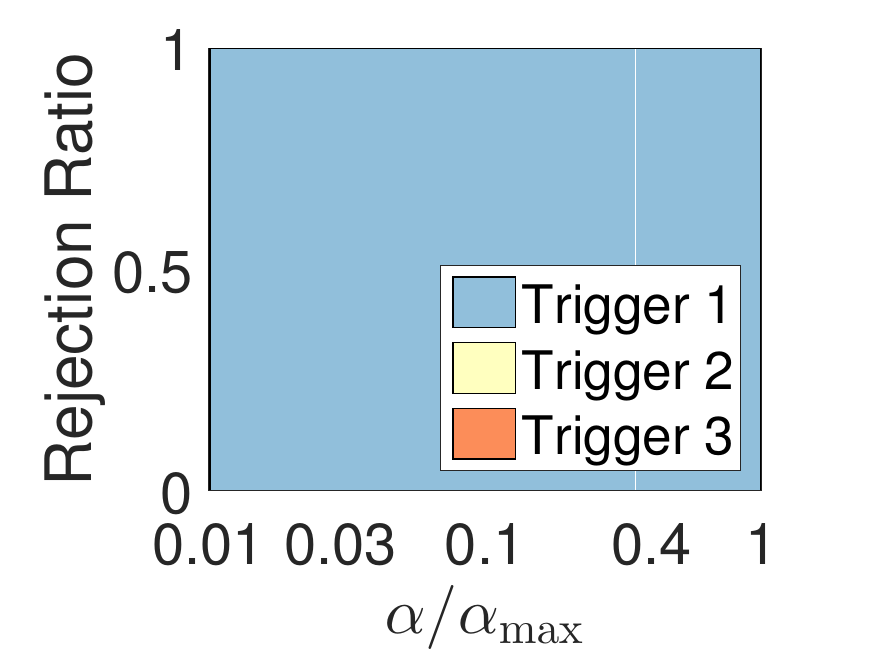}}
		\subfigure[ $\beta/\beta_{\rm{max}}$=0.1]{\includegraphics[scale=0.21]{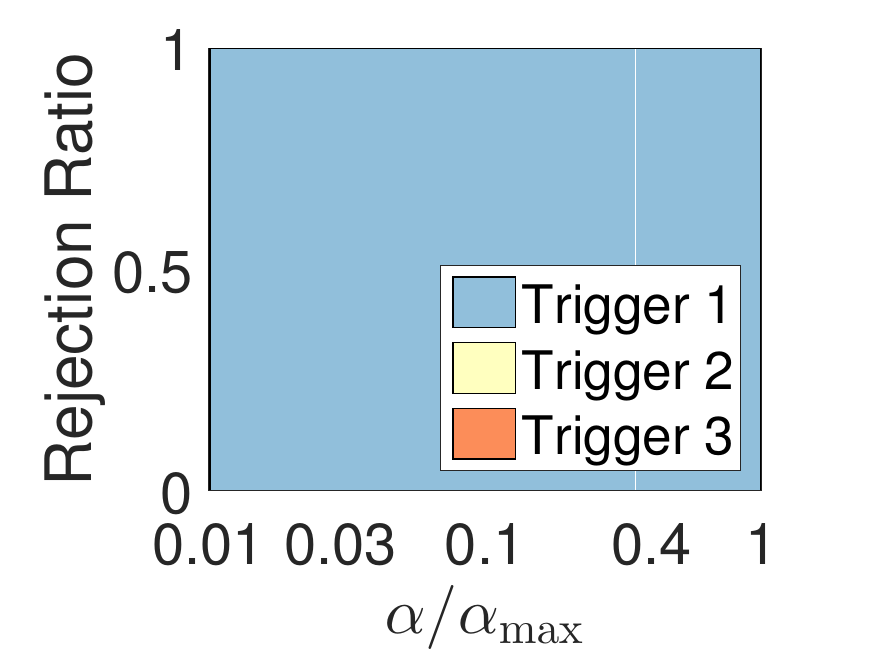}}
		\subfigure[ $\beta/\beta_{\rm{max}}$=0.5]{\includegraphics[scale=0.21]{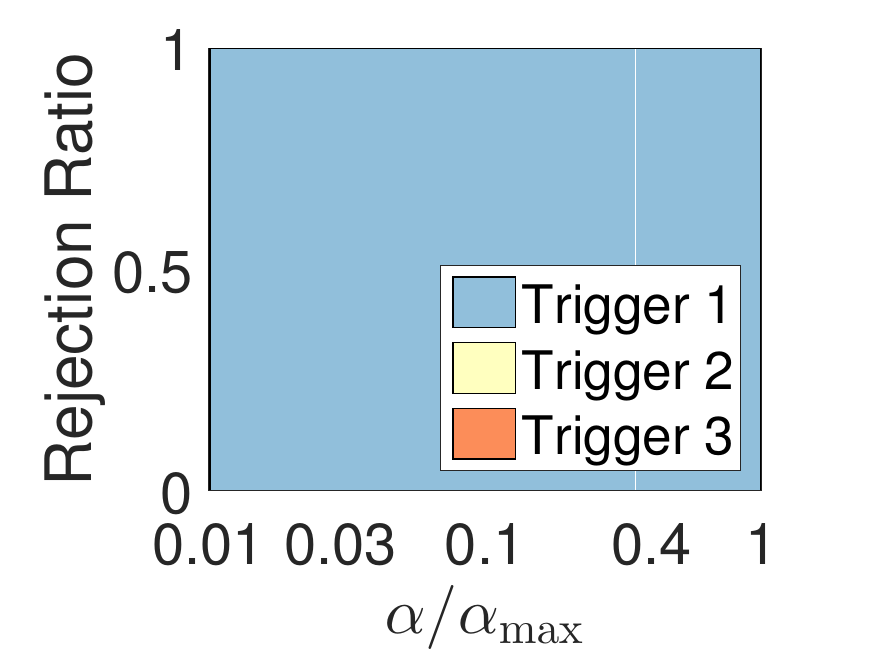}}
		\subfigure[ $\beta/\beta_{\rm{max}}$=0.9]{\includegraphics[scale=0.21]{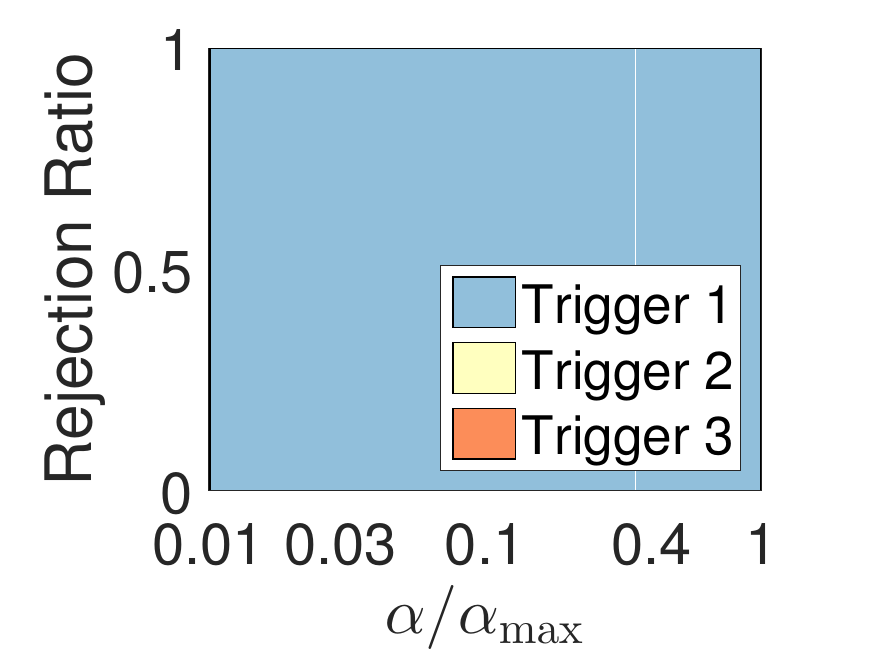}}
		\subfigure[ $\beta/\beta_{\rm{max}}$=0.05]{\includegraphics[scale=0.21]{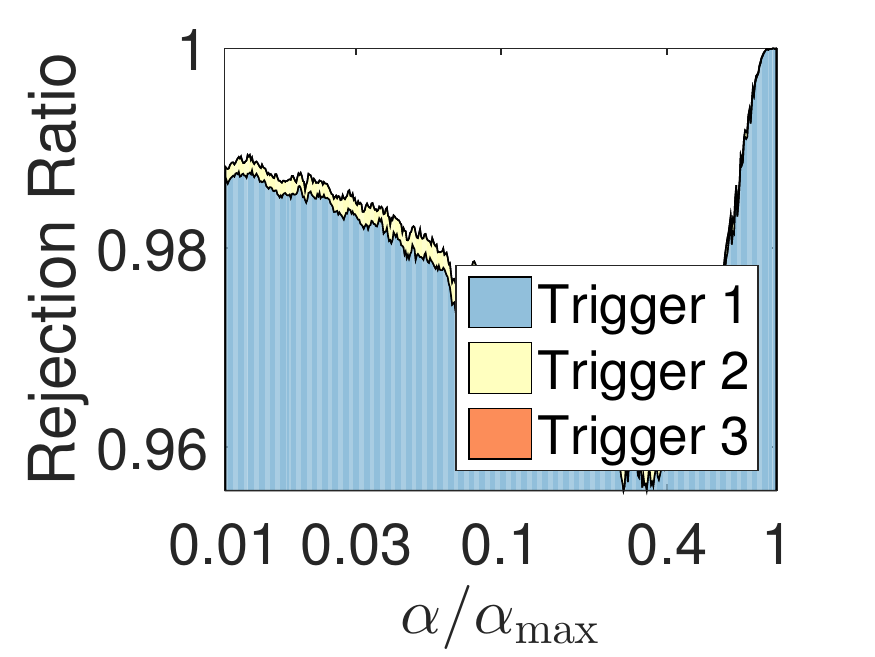}}
		\subfigure[ $\beta/\beta_{\rm{max}}$=0.1]{\includegraphics[scale=0.21]{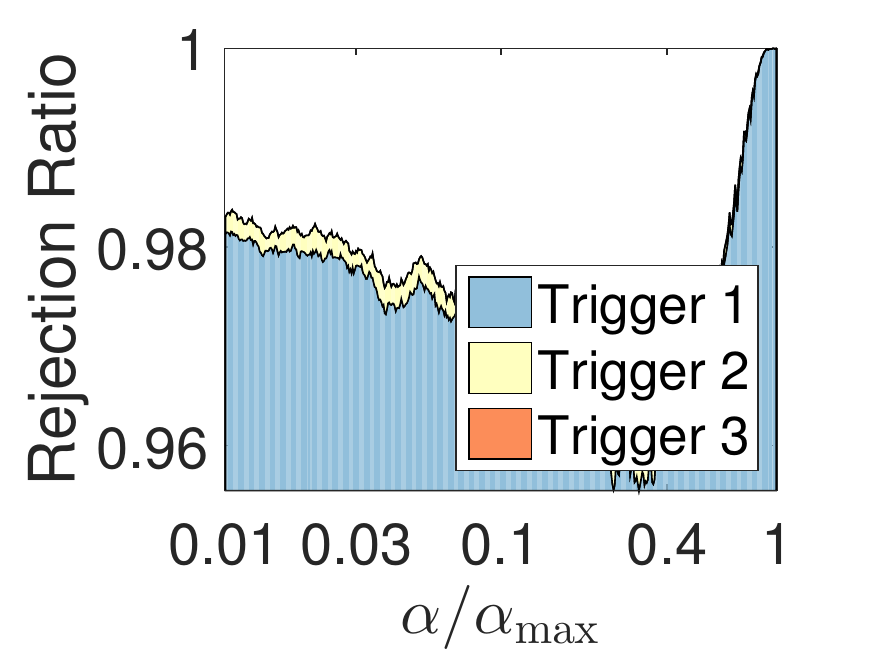}}
		\subfigure[ $\beta/\beta_{\rm{max}}$=0.5]{\includegraphics[scale=0.21]{images/toy_n_10000_d_10000_05_RL_rej_ratio}}
		\subfigure[ $\beta/\beta_{\rm{max}}$=0.9]{\includegraphics[scale=0.21]{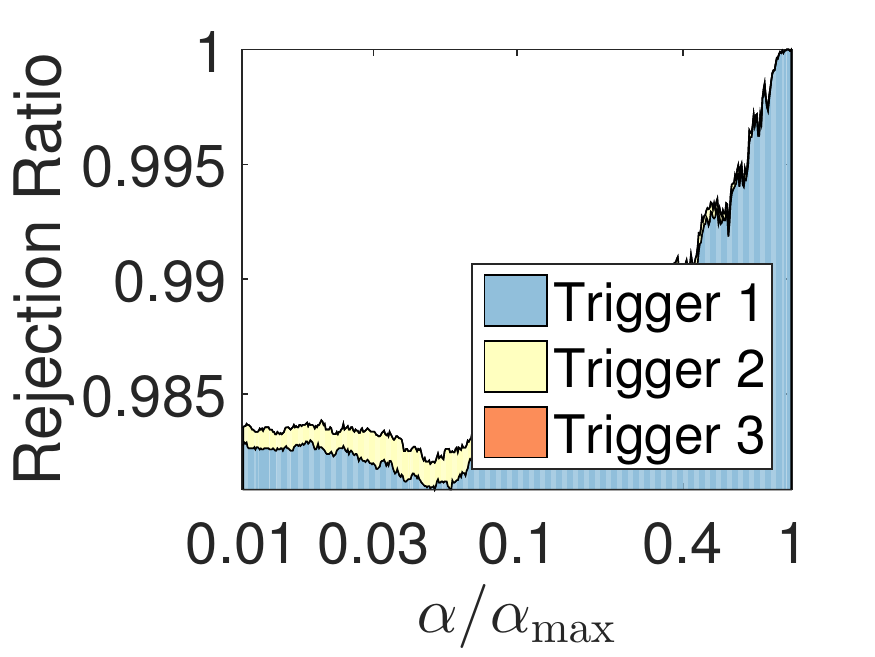}}
		\caption{ Rejection ratios of SIFS on syn1. }
		\label{fig:rejection-ratio-syn-1}
	\end{center}
	\vspace*{-20pt}
\end{figure*}

\clearpage
\newpage
\bibliography{egbib}

\end{document}